%% file: main.tex
\pdfoutput=1

\documentclass[11pt]{article}

\usepackage{booktabs}
\usepackage{multirow}
\usepackage{graphicx}
\usepackage[table,xcdraw]{xcolor}
\usepackage{ulem}

\usepackage{pifont}
\usepackage{soul}
\usepackage[preprint]{acl}

\usepackage{times}
\usepackage{latexsym}

\usepackage[T1]{fontenc}

\usepackage{kotex}
\usepackage{amsmath}

\usepackage{tcolorbox}
\tcbuselibrary{listings,breakable}
\usepackage{needspace}

\usepackage[utf8]{inputenc}

\usepackage{microtype}

\usepackage{inconsolata}

\usepackage{graphicx}

\usepackage{xspace}
\usepackage{makecell}
\usepackage{subcaption}

\title{Ko-MuSR: A Multistep Soft Reasoning Benchmark \\ for LLMs Capable of Understanding Korean}

\author{
  Chanwoo Park\thanks{Dept. of Computer Science, Seoul National University} \\
  \texttt{99chanwoo@snu.ac.kr} \\\And
  Suyoung Park\thanks{Graduate School of Data Science, Seoul National University} \\
  \texttt{itforspark@snu.ac.kr} \\\And
  JiA Kang\footnotemark[2] \\
  \texttt{jia6776@snu.ac.kr} \\\AND
  Jongyeon Park\footnotemark[2] \\
  \texttt{iscopark67@snu.ac.kr} \\\And
  Sangho Kim\footnotemark[2] \\
  \texttt{ksh4931@snu.ac.kr} \\\And
  Hyunji M. Park\footnotemark[2] \\
  \texttt{mhj233@snu.ac.kr} \\\AND
  Sumin Bae\footnotemark[1] \\
  \texttt{dt2085@snu.ac.kr} \\\And
  Mingyu Kang\footnotemark[1] \\
  \texttt{ahagyue@snu.ac.kr} \\\And
  Jaejin Lee\footnotemark[1]\footnotemark[2] \\
  \texttt{jaejin@snu.ac.kr}
  }

\newcommand{\tablecell}[1]{\begin{tabular}[t]{@{}p{\linewidth}@{}}#1\end{tabular}}

\newcommand{\FigureBoxCaption}{}
\newcommand{\FigureBoxLabel}{}

\usepackage{tcolorbox}
\tcbuselibrary{skins, breakable}
\usepackage{caption} 
\newenvironment{hintbox}[3][]{
  \renewcommand{\FigureBoxCaption}{#2}
  \renewcommand{\FigureBoxLabel}{#3}
  \begin{tcolorbox}[
    enhanced, 
    breakable,
    colback=white,
    colframe=black,
    boxrule=0.8pt,
    width=\linewidth,
    before skip=5pt,
    after skip=0pt, 
    boxsep=5pt, 
    #1
  ]
}{
  \end{tcolorbox}
  \par\vspace{5pt} 
  \captionof{figure}{\FigureBoxCaption}\label{\FigureBoxLabel}
}

\begin{document}
\maketitle

\input{sections/00-Abstract}
\input{sections/01-Introduction}
\input{sections/02-RelatedWork}

\input{sections/03-MuSR}

\input{sections/04-KoMuSR}

\input{sections/05-Evaluation}
\input{sections/06-Discussion}

\input{sections/91-Conclusion}

\clearpage
\input{sections/92-Limitations}
\input{sections/93-Ethics}
\input{sections/94-Acknowledgements}


\bibliography{anthology,custom}

\clearpage
\appendix
\input{sections/A02-Ablation-Prompt-Variants}
\input{sections/A04-SLM-Prompting}
\input{sections/A05-Benchmark-Eval-Method}

\input{sections/A10-Computing}
\input{sections/A11-Annotation}

\input{sections/A12-IRB}

\input{sections/A91-ARR-Checklist}
\input{sections/A92-Prompts}
\input{sections/A98-Dataset-Examples}
\input{sections/A99-Large-Tables}

\end{document}

%% file: sections/00-Abstract.tex
\begin{abstract}
We present \textit{Ko-MuSR}, the first benchmark to comprehensively evaluate multistep soft reasoning in long Korean narratives while minimizing data contamination. Built following MuSR~\cite{musr}, Ko-MuSR features fully Korean narratives, reasoning chains, and multiple-choice questions verified by human annotators for logical consistency and answerability.
Evaluations of four large language models---two multilingual and two Korean-specialized---show that multilingual models outperform Korean-focused ones even in Korean reasoning tasks, indicating cross-lingual generalization of reasoning ability. Carefully designed prompting strategies, which combine few-shot examples, reasoning traces, and task-specific hints, further boost accuracy, approaching human-level performance. Ko-MuSR offers a solid foundation for advancing Korean NLP by enabling systematic evaluation of long-context reasoning and prompting strategies. \footnote{Our code can be found at https://github.com/mcrl/Ko-MuSR}

\end{abstract}

%% file: sections/01-Introduction.tex
\section{Introduction}
\label{sec:introduction}

The development of multilingual artificial intelligence (AI) is becoming increasingly important as large language models (LLMs) expand beyond applications centered on English. Since most of the world’s population lives outside the English-speaking realm, there is a significant need for models that cater to diverse linguistic communities. This need has led to the emergence of multilingual LLMs~\cite{lin-etal-2022-shot, gpt4, qwen3}, which demonstrate strong performance across both high- and low-resource languages. Simultaneously, countries like South Korea and France are investing in models~\cite{hypercloavax, exaone4, magistral} tailored to specific languages to enhance their competitiveness in the AI ecosystem.

To evaluate these models, multilingual benchmarks, such as XNLI~\cite{conneau-etal-2018-xnli}, XCOPA~\cite{ponti-etal-2020-xcopa}, and MMMLU~\cite{mmmlu}, have been created. There is a need for benchmarks specifically assessing Korean, and efforts to develop them are currently underway. However, in comparison to English, the variety of Korean benchmarks is still limited, and the ability to generalize task performance across different languages is not yet fully understood.

Current Korean benchmarks inadequately evaluate multistep reasoning outside of coding, STEM fields, or mathematics. There is a notable absence of tasks that require long-context reading, information extraction, and multi-hop reasoning. While CSAT-QA can address this gap, it is vulnerable to data contamination during the critical training stages of LLMs because it is based on the Korean College Scholastic Ability Test (CSAT), which is publicly accessible and widely used in Korea. Additionally, although HRMCR (HAE-RAE Multi-Step Commonsense Reasoning)~\cite{son-etal-2025-multi}  focuses on soft reasoning, it does not fully cover long-context comprehension.

Motivated by the limitations of existing Korean benchmarks, we propose \textit{Ko-MuSR}, a Korean benchmark for multistep soft reasoning. Drawing inspiration from the MuSR benchmark~\cite{musr}, Ko-MuSR constructs logically coherent passages and questions without contamination. Human annotators carefully reviewed and evaluated the tasks to ensure that the answers could be reliably derived from the provided passages, thereby validating the design of the benchmark.

We evaluated Ko-MuSR using two multilingual LLMs and two LLMs specialized in Korean. Surprisingly, our experiments show that the specialized models underperformed compared to the multilingual ones in Korean reasoning tasks. This suggests that the capability for multistep reasoning can transfer across languages, especially between English and Korean. This finding highlights the necessity of improving specialized models to reach the same level of performance as their multilingual counterparts. Additionally, we demonstrate that carefully designed in-context learning strategies can significantly enhance model accuracy, thereby reducing the performance gap. We also find that in-context learning strategies produce inconsistent effects among smaller language models (SLMs), as demonstrated in our evaluation of five additional SLMs.


The main contributions of this paper are summarized as follows:
\begin{itemize}
    \item We introduce Ko-MuSR, the first benchmark specifically designed for long-context multistep soft reasoning in Korean, ensuring it is free from contamination. All benchmark items have been reviewed and solved by human participants, which confirms their answerability and logical consistency.
    \item  Our evaluations of multilingual and Korean-specialized models reveal unexpected gaps in reasoning performance for Korean. We also provide evidence of reasoning transfer between English and Korean.
    \item We discover that models can significantly benefit from carefully crafted in-context learning examples, with some even surpassing human performance in a particular subtask.
\end{itemize}

%% file: sections/02-RelatedWork.tex
\section{Related Work}
\label{sec:related-work}

\paragraph{Multilingual LLMs.}
A variety of multilingual LLMs have been developed to address the needs of non-English speakers around the globe. Early models, such as Multilingual BERT (mBERT)~\cite{bert} and XLM~\cite{xlm}, have evolved into more advanced models like LLaMA4~\cite{llama4}  and Qwen3~\cite{qwen3}, which explicitly incorporate multilingual capabilities into their architectures to better serve a global audience. These models are trained on trillions of tokens gathered from the internet, representing a wide range of linguistic sources. With over 30 trillion tokens in their training datasets, recent multilingual LLMs show impressive performance on multilingual benchmarks, such as Multilingual Massive Multitask Language Understanding (MMMLU)~\cite{mmmlu} and Multilingual Grade School Math (MGSM)~\cite{mgsm}, often matching or even exceeding the capabilities of specialized models that focus on individual languages.

\paragraph{Language-specialized LLMs.}
There has been a growing focus on developing bilingual or language-specialized models to enhance regional AI competitiveness. Examples of these models include CroissantLLM~\cite{croissant}, Magistral~\cite{magistral}, Exaone 4.0~\cite{exaone4}, and HyperCLOVA X SEED Think~\cite{hypercloavax}. Despite being created with relatively limited computational resources, these models achieve competitive, and in some cases superior, performance in their target languages through careful data curation and optimization for specific domains.

\paragraph{Generalization across different languages.}
Multilingual LLMs are becoming increasingly popular because their performance on multilingual benchmarks often exceeds that of language-specific models. For instance, GPT-4~\cite{gpt4} demonstrates significant improvements over GPT-3.5~\cite{gpt3}  on the MMMLU benchmark across 25 languages, showcasing its strong ability to generalize across different languages. Similarly, state-of-the-art open-parameter models like LLaMA 4 and Qwen 3 achieve high scores (above 80) on the same benchmark. Even in Korean soft reasoning tasks, such as HRMCR, a multistep reasoning benchmark with short contexts, these multilingual LLMs often outperform counterparts that are specialized for the Korean language.

\paragraph{Discrepancy across different languages.}
Despite advancements in multilingual model training, reasoning capabilities across different languages continue to be a significant challenge. GPT-4 indicates that performance on the MMMLU benchmark varies greatly by language, often declining in low-resource settings. Similarly, evaluations of PaLM~\cite{palm} on the Mathematical Generalization and Solution Method (MGSM) tasks show considerable degradation when problems are presented in low-resource languages.



\begin{figure*}[t!]
	\centering
  	\includegraphics[width=\linewidth]{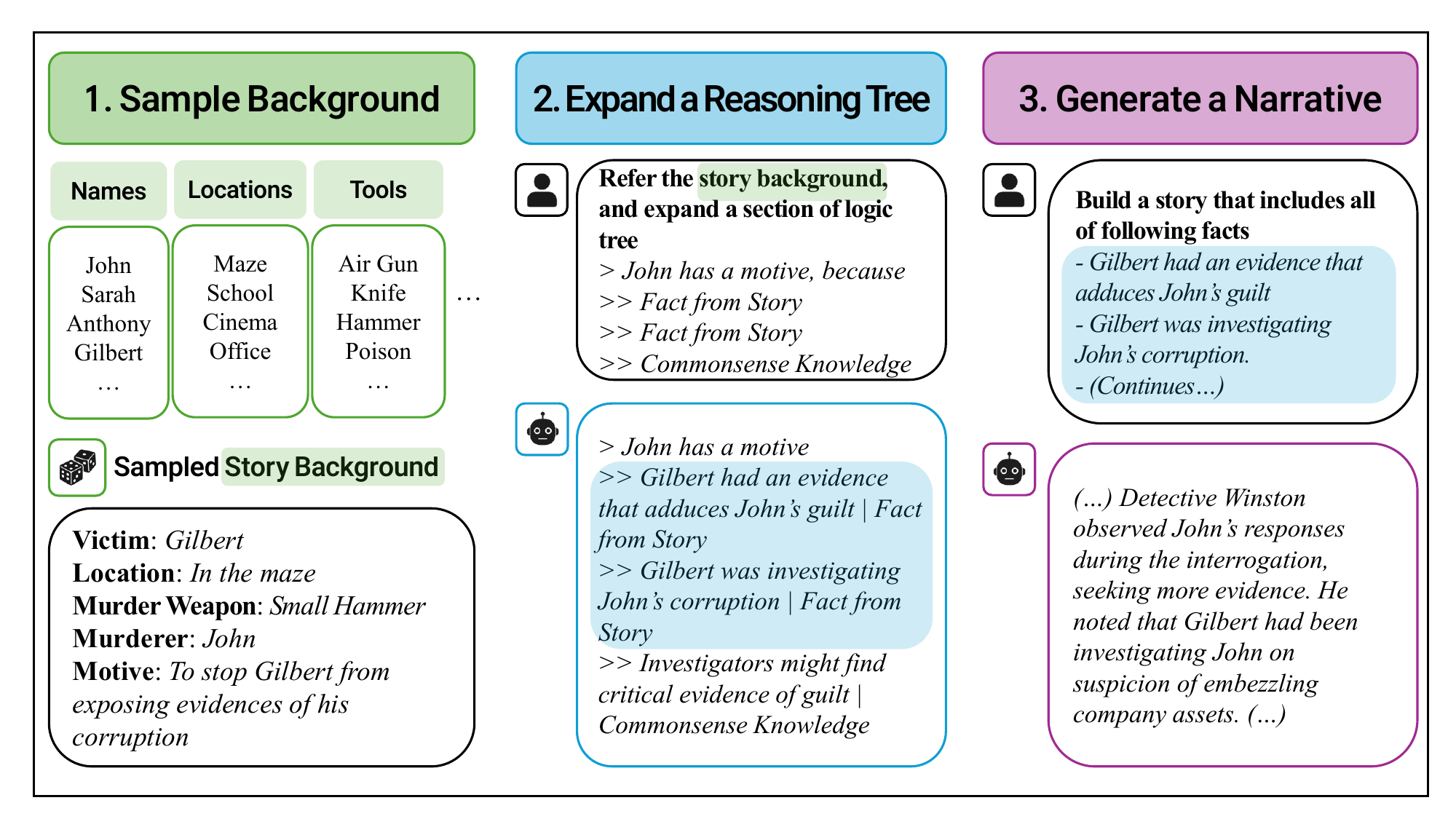}
	\caption{Illustration of the MuSR synthesis process.
    \label{fig:overview}}
\end{figure*}

\paragraph{Benchmarks for Korean LLMs.}
Although various benchmarks have been developed to evaluate Korean-capable LLMs, there is still a significant gap in the diversity and maturity of these benchmarks compared to those available for English models. 

KoBEST~\cite{jang-etal-2022-kobest}, KMMLU~\cite{son-etal-2025-kmmlu}, and Ko-H5~\cite{park-etal-2024-open} have played pioneering roles in evaluating Korean LLMs; however, they do not evaluate reasoning in long-context situations. Benchmarks that focus on generative capabilities, such as LogicKor~\cite{logickor} and KoMT-Bench~\cite{KoMT-Bench}, depend on LLM-based judges, which means their evaluations are influenced by the choice of evaluator models. 

For domain-specific reasoning tasks, HRM8k ~\cite{hrm8k} and KR-Humaneval ~\cite{krhumaneval} provide benchmarks for mathematics and coding, respectively, but they do not assess natural language reasoning skills. Additionally, multiple-choice benchmarks like HAE-RAE~\cite{son-etal-2024-hae} include questions that require local information but fail to evaluate long-context understanding.

Recent efforts in creating benchmarks for multistep soft reasoning in Korean include HRMCR~\cite{son-etal-2025-multi}, which offers tasks with contexts limited to fewer than 500 characters and lacks a natural narrative structure. CSAT-QA~\cite{csatqa}  provides long-context reasoning questions based on the Korean national university entrance exam. However, its public availability and the use of exam materials as training data raise concerns about potential data contamination. Similarly, CLIcK~\cite{kim-etal-2024-click}, which contains a subset of CSAT-QA, faces the same issue. 

Despite the increasing demand for evaluating Korean models on multistep reasoning involving long narratives, there are few suitable benchmarks available, and the existing ones are susceptible to contamination. To address this gap, we have developed a Korean benchmark for multistep soft reasoning that features long, natural narratives and minimizes the risk of overlap with training data. It is designed to test models’ culturally grounded interpretation, rather than simple linguistic accuracy.

%% file: sections/03-MuSR.tex
\section{MuSR}
\label{sec:musr}
MuSR~\cite{musr} is a benchmark created to assess the capability of language models to perform multi-step reasoning within English narratives. Each instance includes a narrative, a question, and multiple-choice answers, with narrative lengths varying from 500 to 1,000 words. Models are required to extract relevant facts, apply logical reasoning, and deduce the correct answer.

The benchmark consists of three subtasks: \textit{murder mysteries}, \textit{object placements}, and \textit{team allocations}.

In the \textit{murder mysteries} subtask, two suspects are presented, one of whom is the murderer. The narrative provides information about the "means," "motive," and "opportunity" of the guilty suspect, while the innocent suspect lacks at least one of these elements. The question posed is to determine which of the two characters is most likely to be guilty.

In the \textit{object placements} subtask, the narrative outlines the locations of objects and the beliefs of characters regarding these locations. The question asks where a specific character believes an object is situated. To answer this question correctly, one must consider which characters witnessed each movement of the objects. Each narrative includes two objects and describes three movements, along with information about whether each of the three characters observed these movements. If a character did not witness a movement, they will continue to believe that the object remains in its original position.

In the \textit{team allocations} subtask, the narrative describes the individual skills of three characters and their relationships within the team. The objective is to assign one character to a solo task while the other two work collaboratively on a different task. This involves determining the best team composition based on both individual abilities and the compatibility between team members.

\paragraph{Synthesis.} 
As shown in Figure~\ref{fig:overview}, the synthesis of MuSR occurs in three stages: \textit{sampling}, \textit{reasoning tree expansion}, and \textit{narrative generation}.  
\begin{enumerate}
\item Sampling stage:  In this initial stage, story backgrounds are sampled from various domain seeds, such as crime scenes, character names, or object types.
\item Reasoning tree expansion stage: In this stage, LLMs enhance the structure by adding fact and deduction nodes.
\item Narrative generation stage: Finally, LLMs generate coherent stories that are aligned with the reasoning tree, relying solely on the fact nodes.
\end{enumerate}

\paragraph{Evaluation.}
For evaluation purposes, the benchmark recommends the \textit{1-Shot CoT+}  configuration. In this setting, a model is given one example question, along with its reasoning process and the final answer. Each question also includes a \textit{hint}, which provides a detailed description of the reasoning strategy specific to that subtask. 

Human participants achieve average accuracies between 90\% and 95\% under this 1-Shot CoT+ setting, indicating that the benchmark is easily solvable by human individuals with multi-step reasoning abilities.

%% file: sections/04-KoMuSR.tex
\section{Constructing Ko-MuSR}
\label{sec:komusr}
Similar to the MuSR synthesis process, our synthesis pipeline consists of three stages: (i) sampling a problem instance, (ii) constructing a reasoning tree, and (iii) generating a narrative based on the reasoning tree. We use two LLMs from OpenAI (GPT-4o and o1)\footnote{https://openai.com} for the synthesis. Human annotators are then tasked with solving each question to verify that the correct answer can indeed be inferred from the corresponding narrative. Finally, we recruit independent participants with no prior exposure to the dataset to ensure that each instance is solvable by humans. These steps ensure that all instances are valid and free from trivial or unsolvable cases.

\subsection{Modifications of the MuSR Synthesis}
We have adapted the MuSR pipeline to create Ko-MuSR, focusing on high-quality narratives in Korean. All prompts have been translated into Korean. For domain seeds related to object placement and team allocation subtasks, we generate Korean-specific seeds using GPT-4o. For murder mysteries, we create culturally appropriate character names in Korean with GPT-4o. Additionally, other domain seeds, such as crime scenes, are translated and adjusted as needed to ensure they fit naturally within the Korean context and culture. This process ensures that domain seeds remain faithful to MuSR while producing culturally coherent settings.

We also translate all prompts for the reasoning tree and narrative generation into Korean to ensure that the logical structure is expressed in Korean. This prevents the inclusion of keywords that could trivialize deduction in the reasoning tree and, consequently, in the narrative.

For reasoning tree expansion and narrative synthesis, we use GPT-4o for the murder mystery and team allocation subtasks. However, for object placement, the outputs from GPT-4o did not meet our quality standards, so we opted to use o1, which produced higher-quality instances.

\subsection{Reviewing the Intermediate Dataset}
All synthesized problems were initially reviewed by the authors, who acted as annotators. Each annotator received one example and a translated hint from MuSR to ensure consistency with the original benchmark design. The review process consisted of two phases. 

In the first phase, each annotator independently solved every problem, submitted their answer, and categorized the instance as either \textit{pass}, \textit{minor issue}, or \textit{fail}.  Instances labeled as \textit{pass} were accepted only if the annotator’s answer matched the correct solution. Those marked as \textit{fail} were discarded.

In the second phase, any instance labeled as \textit{minor issue} or incorrectly marked as \textit{pass}  was re-evaluated by two independent annotators. These secondary annotators were given the supposed correct answer and asked to verify whether it could be logically derived from the provided narrative. An instance was included in the final dataset only if both annotators deemed it as \textit{pass}; otherwise, it was excluded.

\input{sections/tables/table-human-eval}

\subsection{Human Evaluation}
To further validate the quality of our dataset, we sampled a subset of questions and recruited five external participants, ages 20 to 30, who were not affiliated with our research group. They were provided with one example and a hint to assist them in answering the questions. The questionnaire was administered on a laptop with internet access disabled to prevent participants from using AI tools, thus eliminating any potential influence of such tools on human benchmark performance. This process allowed us to measure both human performance and the effectiveness of our validation procedure. Table~\ref{tab:human-eval} presents the human performance results on Ko-MuSR. This ensures the validity of our review process and provides a measure of human performance on Ko-MuSR.

\input{sections/tables/table-benchmark-statistics}

\subsection{Dataset Statistics}

Ko-MuSR consists of three subtasks, each containing 250 examples. Dataset statistics are summarized in Table~\ref{tab:benchmark-stats}.

%% file: sections/tables/table-human-eval.tex
\begin{table}[t]
\centering
\small
\setlength{\tabcolsep}{6pt}
\renewcommand{\arraystretch}{1.05}
\begin{tabular}{c|ccc}
\toprule
\textbf{Statistics} &
\makecell{Murder\\Mysteries} &
\makecell{Object\\Placements} &
\makecell{Team\\Allocations} \\
\midrule \midrule
\# Questions & 31 & 40 & 38 \\  \midrule
Max & 93.5 & 97.5 & 92.1 \\
Average & 85.5 & 93.8 & 87.5 \\
Median & 83.9 & 97.5 & 86.8 \\
Min & 77.4 & 85.0 & 81.6 \\
\bottomrule
\end{tabular}
\caption{Human evaluation result of Ko-MuSR.}
\label{tab:human-eval}
\end{table}

%% file: sections/tables/table-benchmark-statistics.tex
\begin{table}[t]
\centering
\small
\setlength{\tabcolsep}{5pt}
\renewcommand{\arraystretch}{1.05}
\resizebox{\columnwidth}{!}{%
\begin{tabular}{cc|ccc}
\toprule
\textbf{Lang.} & \textbf{Statistics} &
\makecell{Murder\\Mysteries} &
\makecell{Object\\Placements} &
\makecell{Team\\Allocations} \\
\midrule \midrule
\multirow{3}{*}{En} 
& \# Questions & 250 & 256 & 250 \\
& Avg. Length (chars) & 5471 & 4919 & 3104 \\
& Avg. Length (words) & 905 & 787 & 494 \\
\midrule
\multirow{3}{*}{Ko} 
& \# Questions & 250 & 250 & 250 \\
& Avg. Length (chars) & 2373 & 1049 & 1091 \\
& Avg. Length (words) & 561 & 262 & 264 \\
\bottomrule
\end{tabular}}
\caption{Statistics of MuSR and Ko-MuSR narratives across subtasks.}
\label{tab:benchmark-stats}
\end{table}

%% file: sections/05-Evaluation.tex
\input{sections/tables/table-zeroshot-direct-nohint}
\input{sections/tables/table-slms-v3}

\section{Evaluation}
\label{sec:evaluation}
\subsection{Evaluation Setup}

\paragraph{Models.}
We evaluate four LLMs with MuSR and Ko-MuSR: two multilingual models, Qwen3 32B~\cite{qwen3} and LLaMA 4 Scout Instruct~\cite{llama4}, and two models specialized in Korean, Exaone 4.0 32B~\cite{exaone4} and HyperCLOVA X SEED Think 14B~\cite{hypercloavax}. We then evaluate 5 SLMs: three multilingual models, QWEN3 0.6B, 1.7B, 4B and two models specialized in Korean: Exaone 4.0 1.2B and Exaone 3.5 2.4B Instruct~\cite{exaone3.5}.

\paragraph{Model training.}
Qwen3 32B is trained on over 36 trillion tokens in total, including 5 trillion high-quality, reasoning-oriented tokens during its pretraining stage. Qwen 3 0.6B, 1.7B, and 4B models are distilled from Qwen3 32B. LLaMA 4 Scout Instruct is pretrained on over 30 trillion multimodal tokens, which encompass text, images, and video, with at least one billion tokens available for each language across more than 100 languages. Exaone 4.0 32B is trained on approximately 14 trillion tokens, covering English, Korean, and Spanish, with an emphasis on balancing multilingual reasoning and linguistic precision. Exaone 4.0 1.2B is trained on approximately 12 trillion tokens with the same method as Exaone 4.0 32B. HyperCLOVA X SEED Think 14B is trained on around 6 trillion tokens in English and Korean, with a strong focus on Korean domain adaptation. 
Exaone 3.5 2.4B has been trained on approximately 6.5 trillion tokens in English and Korean, incorporating various data sources for improved performance in general domains. For evaluation, we use the default generation configurations provided by the models. All evaluation experiments are conducted twice, and we report the average accuracy from the two runs.

\paragraph{Prompts.}
In line with the MuSR framework~\cite{musr}, we developed multiple prompt variants to assess model performance under various prompting conditions. First, we create a \textit{prompt split} that is independent of the test set. For English, we use one example from the original MuSR artifact and generate two additional examples using the same methodology, resulting in three examples for each subtask. For Korean, we independently construct three new prompt examples for each subtask as well.

We use the \textit{Chain-of-Thought (CoT)} prompting strategy~\cite{cot} in our evaluation. For English, we use one original MuSR CoT example that includes a question, a step-by-step reasoning process, and a final answer. We synthesize two additional examples using the same methodology. For Korean, we create three new CoT examples, each containing complete reasoning traces. Additionally, we adapt the task-specific \textit{hints} from MuSR and translate them into Korean for Ko-MuSR. 
Hints provide detailed descriptions of how to approach the task, typically consisting of four to five sentences. 
Prompt templates and detailed examples can be found in Appendix~\ref{sec:prompt-details}.

A specific name identifies each configuration. For example, the name \textit{3-Shot/CoT/Hint} indicates that the LLM is provided with three in-context learning examples that include both reasoning traces and hints. We clearly specify the number of examples, whether Chain-of-Thought (CoT) reasoning is used by referring to it as \textit{CoT}  or \textit{Direct} for straightforward responses, and we indicate the presence or absence of hints by using \textit{Hint} and \textit{No Hint}. 


\subsection{Model Performance}

Table~\ref{tab:zeroshot-direct-nohint} presents the zero-shot performance of models on both MuSR and Ko-MuSR. Some MuSR instances for \textit{object placement} contain five answer choices, which leads to a random baseline accuracy of 24.6. For Ko-MuSR, we ensure that all questions have a consistent number of answer choices across subtasks. We measure model accuracy for each subtask, \textit{Murder Mysteries (MM)}, \textit{Object Placements (OP)}, and \textit{Team Allocations (TA)}.

We begin by evaluating model accuracy without employing advanced prompting techniques (such as Chain of Thought or hints) to establish a baseline for direct reasoning capabilities. Among all LLMs, Qwen3 32B achieves the highest accuracy across both English and Korean benchmarks, closely followed by LLaMA 4 Scout Instruct. This consistency indicates that strong multilingual reasoning abilities can effectively transfer between languages, particularly between English and Korean. These results suggest that multistep reasoning in long narratives is primarily influenced by the reasoning structure rather than by linguistic specialization.

We will next evaluate the performance of our LLMs under improved prompting conditions (see Table~\ref{tab:3shot-cot-hint}). By incorporating three few-shot examples along with CoT and hint prompts, we observe a significant increase in accuracy across all models. 

Table \ref{tab:prompt-ablation-simple} illustrates a case involving the Naver HyperClova X SEED Think 14B model with various prompting configurations. In Appendix~\ref{sec:ablation-prompt}, we analyze the impact of the number of few-shot examples and the prompting strategies, demonstrating that both chain-of-thought examples and hints contribute to performance enhancements. The configuration with CoT and hints achieves nearly optimal performance in the team allocation task within the Korean context, where three models exceed human accuracy, and one matches it. As model accuracy approaches human performance, the differences among the models diminish. However, Korean-specialized models still underperform in the object placement task within this context, indicating there is still room for improvement in Korean-specific reasoning.

We finally evaluate the effects of the same prompting configuration—\textit{3-Shot/CoT/Hint}—on small language models (SLMs), which is shown in Table~\ref{tab:slm-main-v3}. We notice that the impact of the prompting is inconsistent across different SLMs. In an extreme case, the accuracy of Qwen3 1.7B on the Object Placements task in English decreases by 12.7 points when 3-shot CoT Hint prompts are used. This finding indicates that prompting methods should be carefully tailored to each specific task to enhance the multistep reasoning abilities of SLMs.


%% file: sections/tables/table-zeroshot-direct-nohint.tex

\begin{table*}[!t]
\centering
\begin{minipage}{\linewidth}
\centering
\resizebox{0.8\textwidth}{!}{ 
\begin{tabular}{l|ccc|ccc}\toprule
Model &En-MM &En-OP &En-TA &Ko-MM &Ko-OP &Ko-TA \\\midrule \midrule
Random &50.0 &24.6 &33.3 &50.0 &25.0 &33.3 \\ \midrule
QWEN3 32B &\textbf{66.4} &52.7 &\textbf{73.4} &\textbf{67.6} &\textbf{80.0} &\textbf{79.6} \\
LLaMA 4 Scout Instruct &\ul{65.4} &\ul{54.1} &\ul{60.4} &\ul{56.0} &\ul{67.8} &\ul{64.2} \\
Exaone 4.0 32B &61.2 &\textbf{55.7} &58.8 &52.8 &37.4 &63.2 \\
HyperClova X SEED Think 14B &51.6 &38.7 &44.0 &44.6 &41.2 &58.6 \\ \midrule
Human (Avg.) &92.1 &90.0 &95.1 &85.5 &93.8 &87.5 \\
\bottomrule
\end{tabular}}
\vspace{-0.5\baselineskip}
\caption{Evaluation result with zero shot, direct, and no hint. Boldface indicates the highest numbers, and underlined numbers indicate the second highest.}\label{tab:zeroshot-direct-nohint}
\end{minipage}\\
\vspace{\baselineskip}

\begin{minipage}{\linewidth}
\centering
\resizebox{0.8\textwidth}{!}{ 
\begin{tabular}{l|ccc|ccc}\toprule
Model &En-MM &En-OP &En-TA &Ko-MM &Ko-OP &Ko-TA \\\midrule \midrule
Random &50.0 &24.6 &33.3 &50.0 &25.0 &33.3 \\ \midrule
QWEN3 32B &\textbf{82.4} &\textbf{74.2} &\textbf{88.8} &\textbf{76.0} &\ul{81.8} &\textbf{89.0} \\
LLaMA 4 Scout Instruct &\ul{75.6} &68.6 &84.2 &\ul{72.8} &\textbf{83.0} &\ul{88.0} \\
Exaone 4.0 32B &75.2 &\ul{72.1} &\ul{85.8} &71.2 &71.6 &85.8 \\
HyperClova X SEED Think 14B &68.6 &66.2 &81.0 &70.6 &50.0 &87.8 \\ \midrule
Human (Avg.) &92.1 &90.0 &95.1 &85.5 &93.8 &87.5 \\
\bottomrule
\end{tabular}}
\vspace{-0.5\baselineskip}
\caption{Evaluation result with three shots, CoT, and hints. Boldface indicates the highest numbers, and underlined numbers indicate the second highest.}\label{tab:3shot-cot-hint}
\end{minipage}\\
\vspace{\baselineskip}

\begin{minipage}{\linewidth}
\centering
\resizebox{0.8\textwidth}{!}{ 
\begin{tabular}{l|ccc|ccc}\toprule
Model &En-MM &En-OP &En-TA &Ko-MM &Ko-OP &Ko-TA \\\midrule\midrule
HyperClova X SEED Think 14B &51.6 &38.7 &44.0 &44.6 &41.2 &58.6 \\
+ 1 Shot / Direct / No Hint &63.8 &42.0 &53.6 &50.0 &39.0 &51.8 \\
+ 3 Shots / Direct / No Hint &\ul{67.8} &63.7 &76.0 &\ul{65.4} &\textbf{52.4} &\ul{86.2} \\
+ 3 Shots / Direct / Hint &67.0 &\ul{64.8} &\textbf{83.8} &63.8 &49.6 &83.8 \\
+ 3 Shots / CoT / Hint &\textbf{68.6} &\textbf{66.2} &\ul{81.0} &\textbf{70.6} &\ul{50.0} &\textbf{87.8} \\
\bottomrule
\end{tabular}}
\vspace{-0.5\baselineskip}
\caption{Case study of the HyperClova SEEDX Think 14B model with various prompting configurations. The first row denotes the zero-shot, direct, and no-hint configuration. Boldface indicates the highest numbers, and underlined numbers indicate the second highest.}
\label{tab:prompt-ablation-simple}
\end{minipage}

\end{table*}

%% file: sections/tables/table-slms-v3.tex
\begin{table*}[!htbp]\centering
\resizebox{0.72\textwidth}{!}{ 
\begin{tabular}{l|ccc|ccc}\toprule
&En-MM &En-OP &En-TA &Ko-MM &Ko-OP &Ko-TA \\\midrule \midrule
QWEN3 0.6B &\textbf{59.2} &\textbf{52.0} &\textbf{38.8} &\textbf{53.6} &28.6 &\textbf{38.0} \\
+ 3 Shots / CoT / Hint &55.4 &46.7 &33.0 &49.0 &\textbf{30.2} &33.4 \\ \midrule
Exaone 4.0 1.2B &55.6 &\textbf{44.7} &49.8 &48.0 &29.0 &\textbf{44.8} \\
+ 3 Shots / CoT / Hint &\textbf{64.8} &\textbf{44.7} &\textbf{54.8} &\textbf{58.6} &\textbf{31.6} &35.8 \\ \midrule
QWEN3 1.7B &60.6 &\textbf{52.3} &\textbf{59.0} &50.0 &\textbf{34.0} &\textbf{63.2} \\
+ 3 Shots / CoT / Hint &\textbf{66.0} &39.6 &55.4 &\textbf{63.0} &33.0 &54.2 \\ \midrule
Exaone 3.5 2.4B Instruct &51.4 &41.6 &\textbf{44.0} &51.0 &30.0 &42.4 \\
+ 3 Shots / CoT / Hint &\textbf{63.4} &\textbf{48.6} &37.2 &\textbf{56.0} &\textbf{41.2} &\textbf{42.8} \\ \midrule
QWEN3 4B &62.8 &55.5 &66.0 &59.0 &51.0 &75.4 \\
+ 3 Shots / CoT / Hint &\textbf{74.0} &\textbf{70.7} &\textbf{85.4} &\textbf{69.8} &\textbf{55.2} &\textbf{79.6} \\
\bottomrule
\end{tabular}}
\caption{Evaluation results with SLMs. The 3-Shot/CoT/Hint configuration may reduce model accuracy in SLMs. Additional numbers can be found in Appendix~\ref{sec:slm-prompting}.}
\label{tab:slm-main-v3}
\end{table*}

%% file: sections/06-Discussion.tex

\section{Discussions}

\paragraph{LLMs' reasoning generalizes across languages.}

Previous evaluations of HRCRM~\cite{son-etal-2025-multi} have demonstrated that multilingual models can effectively reason using Korean-specific knowledge. HRCRM is divided into two subsets: the \textit{date} subset, which involves reasoning about Korean holidays, date expressions, and conversions between the solar and lunar calendars, and the \textit{zodiac} subset, which tests knowledge of Korea’s traditional age system and zodiac relationships. 

Despite requiring reasoning grounded in local cultural contexts, the benchmark shows that multilingual models such as o1~\cite{oaio1} and GPT-4o~\cite{oaigpt4o} outperform a Korean-specialized model, Exaone 3.5-32B~\cite{exaone3.5}. This result suggests that the reasoning abilities of multilingual LLMs generalize beyond the primary languages on which they were trained, encompassing even culturally specific reasoning.

Our findings on Ko-MuSR further support this observation. As demonstrated in Table~\ref{tab:zeroshot-direct-nohint}, multilingual models exhibit strong reasoning performance, even on Korean benchmarks that require understanding long, narrative-style contexts. 

Interestingly, the ranking of zero-shot accuracy among the LLMs aligns with the estimated number of text tokens used during their pretraining. Qwen3 32B, which was trained on over 36 trillion text tokens, achieves the highest accuracy. This is followed by LLaMA 4 Scout Instruct, which was trained on approximately 30 trillion multimodal tokens, including text, images, and video. 
Among LLMs specialized for Korean, Exaone 4.0 32B, trained on 14 trillion text tokens and featuring a denser parameterization, outperforms HyperCLOVA X SEED Think 14B, which was trained on 6 trillion tokens.

Both the HRCRM and Ko-MuSR results demonstrate that large multilingual models are capable of multistep reasoning across languages, cultural boundaries, and even in long-context narrative settings. However, their performance still falls short of human accuracy, indicating a need for improvement.




\paragraph{Detailed prompts and reasoning examples are beneficial.}
Although state-of-the-art multilingual models maintain strong performance, language-specific models can bridge the gap through effective prompting strategies. Our results demonstrate that using \textit{3-shot/CoT} prompting, along with descriptive hints, significantly improves performance across the four evaluated LLMs. We examine the impact of each component on model performance in Appendix~\ref{sec:ablation-prompt} and find that among the three components, the \textit{hint}---a detailed task description--- consistently provides the most significant performance gain across models. This approach allows several models to achieve or even exceed human-level accuracy, thereby reducing performance discrepancies among different model families. These findings emphasize that specialized models, despite having smaller training budgets, can attain competitive reasoning performance in their target language through structured prompting.

\paragraph{SLMs benefit less from detailed prompts and reasoning examples.}

\citet{lu2025smalllanguagemodelssurvey} have shown that smaller models benefit less from in-context learning compared to larger ones by evaluating 24 small language models (SLMs). They compared the average accuracy of 5-shot and 0-shot settings across eight English benchmarks using models ranging from 100 M to 4 B parameters. The results indicate that the performance gains from in-context examples increase with model size; in fact, some smaller models even exhibit performance drops. This suggests that including examples in the prompt can hinder accuracy when the model size is insufficiently small.

Our evaluation of Ko-MuSR reinforces this finding. The smallest SLM (Qwen3 0.6B) shows improvements with the 3-shot/CoT/Hint prompt in only one subtask, while larger SLMs benefit across multiple subtasks. Additionally, we observe significant performance degradation---up to 12.7 points in the most extreme case---when a 3-shot/CoT/Hint prompt is provided to smaller models. In contrast, LLMs generally improve under the same prompting conditions and do not show these performance drops.

Overall, both the results from the English benchmarks and our Ko-MuSR findings consistently demonstrate that LLMs gain more from in-context learning, while smaller models may be adversely affected. This trend appears to persist across languages, indicating a general limitation of in-context learning for SLMs. Further analysis can be found in Appendix~\ref{sec:slm-prompting}.

%% file: sections/91-Conclusion.tex
\section{Conclusions}
\label{sec:conclusion}

In this paper, we introduce \textit{Ko-MuSR}, a Korean benchmark for evaluating multistep soft reasoning in long narrative contexts. It follows MuSR's synthesis process and minimizes the risk of contamination. It consists of 250 questions across three subtasks assessing different reasoning aspects in LLMs. The dataset is generated through a controlled pipeline and refined via human review to ensure factual and logical validity. Our empirical results show that state-of-the-art multilingual models outperform Korean-specialized ones on Ko-MuSR. Providing three in-context examples with explicit reasoning and hints markedly improves performance, nearing human-level accuracy. These results suggest that structured prompting and long-context conditioning can improve the reasoning ability of Korean LLMs, enabling specialized models to narrow the gap with their multilingual counterparts. Despite its limited task scope, Ko-MuSR offers a solid foundation for advancing Korean NLP by enabling systematic evaluation of long-context reasoning and prompting strategies.

%% file: sections/92-Limitations.tex
\section*{Limitations 
\label{sec:limitations}}

We reconstruct MuSR for only for Korean, limiting our scope of research in model capability in two languages: English and Korean.

While we present representative state-of-the-art open-parameter LLMs for the evaluation, we do not evaluate all the models that are capable of Korean and English.

While repeated trials on the benchmark could further reduce variance caused by randomness, we conducted only two evaluation runs due to computational resource constraints.

%% file: sections/93-Ethics.tex
\section*{Ethics Statement \label{sec:ethics}}

We construct the Ko-MuSR benchmark using only LLMs, primarily OpenAI\footnote{https://openai.com} GPT-4o and o1 models. Since the content is fictional and synthesized, the dataset does not contain personal information or real-world references that may raise ethical concerns.

The authors, as members of the research group, participated in reviewing the Korean benchmark. Their work was supported by the grants listed in the Acknowledgements section. Compensation for their contribution was acknowledged at KRW 10,030 per hour, corresponding to the 2025 minimum wage in Korea.

We evaluated the human performance on the benchmark. All evaluation procedures were conducted under the supervision of the Institutional Review Board of Seoul National University (SNUIRB, No. 2506/001-001). Participants’ contributions were recognized as one full day (8 hours) of work and compensated with KRW 100,000, exceeding the 2025 minimum daily wage of KRW 80,240 in Korea.

%% file: sections/94-Acknowledgements.tex
\section*{Acknowledgements \label{sec:ack}}

This work was partially supported by the National Research Foundation of Korea (NRF) under Grant No. RS-2023-00222663 (Center for Optimizing Hyperscale AI Models and Platforms), and by the Institute for Information and Communications Technology Promotion (IITP) under Grant No. 2018-0-00581 (CUDA Programming Environment for FPGA Clusters) and No. RS-2025-02304554 (Efficient and Scalable Framework for AI Heterogeneous Cluster Systems), all funded by the Ministry of Science and ICT (MSIT) of Korea. It was also partially supported by the Korea Health Industry Development Institute (KHIDI) under Grant No. RS-2025-25454559 (Frailty Risk Assessment and Intervention Leveraging Multimodal Intelligence for Networked Deployment in Community Care), funded by the Ministry of Health and Welfare (MOHW) of Korea. Additional support was provided by the BK21 Plus Program for Innovative Data Science Talent Education (Department of Data Science, Seoul National University, No. 5199990914569) and the BK21 FOUR Program for Intelligent Computing (Department of Computer Science and Engineering, Seoul National University, No. 4199990214639), both funded by the Ministry of Education (MOE) of Korea. This work was also partially supported by the Artificial Intelligence Industrial Convergence Cluster Development Project, funded by the MSIT and Gwangju Metropolitan City. Research facilities were provided by the Institute of Computer Technology (ICT) at Seoul National University.

%% file: sections/A02-Ablation-Prompt-Variants.tex
\section{Ablations on Prompt Variants}
\label{sec:ablation-prompt}

\input{sections/tables/table-ablation-num-fewshots}
\input{sections/tables/table-ablation-hints}
\input{sections/tables/table-ablation-cot}

\paragraph{Effects of In-Context Learning}
Table~\ref{tab:ablation-fewshots} shows the effect of the number of in-context examples on model performance. We find that providing few-shot examples substantially improves model accuracy on Korean tasks, suggesting that models can analogically reason about similar problems when given task-specific examples. However, the performance gain from adding more examples diminishes under the CoT and Hint settings, where the models already approach their upper performance bound. In contrast, for the HyperCLOVA X SEED Think model, incorporating three examples yields a significant boost, indicating that three-shot prompting is a particularly effective configuration for this model.

\paragraph{Effects of Hints}
\label{sec:ablation-prompt-hint}
Table~\ref{tab:ablation-hint} reports the impact of providing task-specific hints. We find that hints generally enhance model accuracy, underscoring the importance of precise task descriptions.

\paragraph{Effects of Chain-of-Thought Reasoning}
Table~\ref{tab:ablation-cot} presents the effect of including chain-of-thought (CoT) examples. We observe that CoT examples are particularly beneficial for Korean-specialized models, while multilingual models occasionally experience performance degradation. Nevertheless, since CoT prompting provides substantial improvements when effective and rarely leads to significant drops, we adopt CoT examples as part of our main evaluation configuration.

%% file: sections/tables/table-ablation-num-fewshots.tex
\begin{table*}[!htp]\centering
\resizebox{\textwidth}{!}{ 
\begin{tabular}{l|ccc|ccc}\toprule 
Model &En-MM &En-OP &En-TA &Ko-MM &Ko-OP &Ko-TA \\\midrule \midrule
QWEN3 32B &66.4 &52.7 &73.4 &67.6 &80.0 &79.6 \\
+ 1 Shot / Direct / No Hint &73.6 &62.1 &81.0 &69.2 &83.0 &80.6 \\
+ 3 Shots / Direct / No Hint &76.2 &64.6 &76.0 &71.8 &85.4 &81.8 \\
+ 1 Shot / CoT / Hint &83.6 &75.8 &89.6 &77.8 &79.2 &87.8 \\
+ 3 Shots / CoT / Hint &82.4 &74.2 &88.8 &76.0 &81.8 &89.0 \\\midrule
LLaMA 4 Scout &65.4 &54.1 &60.4 &56.0 &67.8 &64.2 \\
+ 1 Shot / Direct / No Hint &69.6 &56.4 &70.0 &72.8 &73.0 &70.8 \\
+ 3 Shots / Direct / No Hint &72.6 &61.5 &78.0 &70.4 &75.6 &82.6 \\
+ 1 Shot / CoT / Hint &77.2 &69.7 &84.4 &75.0 &80.8 &85.6 \\
+ 3 Shots / CoT / Hint &75.6 &68.6 &84.2 &72.8 &83.0 &88.0 \\\midrule
Exaone 4.0 32B &61.2 &55.7 &58.8 &52.8 &37.4 &63.2 \\
+ 1 Shot / Direct / No Hint &74.8 &57.8 &63.4 &62.4 &48.8 &65.4 \\
+ 3 Shots / Direct / No Hint &76.0 &62.5 &69.4 &68.6 &60.2 &73.6 \\
+ 1 Shot / CoT / Hint &80.8 &74.0 &86.6 &67.6 &71.0 &79.6 \\
+ 3 Shots / CoT / Hint &75.2 &72.1 &85.8 &71.2 &71.6 &85.8 \\\midrule
HyperClova X SEED Think 14B &51.6 &38.7 &44.0 &44.6 &41.2 &58.6 \\
+ 1 Shot / Direct / No Hint &63.8 &42.0 &53.6 &50.0 &39.0 &51.8 \\
+ 3 Shots / Direct / No Hint &67.8 &63.7 &76.0 &65.4 &52.4 &86.2 \\
+ 1 Shot / CoT / Hint &70.2 &51.4 &75.2 &63.8 &37.6 &84.6 \\
+ 3 Shots / CoT / Hint &68.6 &66.2 &81.0 &70.6 &50.0 &87.8 \\
\bottomrule
\end{tabular}}
\caption{Ablation - Effects of Number of Few Shot Examples. The first row denotes the zero-shot, direct, and no-hint configuration.}\label{tab:ablation-fewshots}
\end{table*}

%% file: sections/tables/table-ablation-hints.tex
\begin{table*}[!htp]\centering
\resizebox{\textwidth}{!}{ 
\begin{tabular}{l|ccc|ccc}\toprule
Model &En-MM &En-OP &En-TA &Ko-MM &Ko-OP &Ko-TA \\\midrule \midrule
QWEN3 32B &66.4 &52.7 &73.4 &67.6 &80.0 &79.6 \\
+ 1 Shot / Direct / No Hint &73.6 &62.1 &81.0 &69.2 &83.0 &80.6 \\
+ 1 Shot / Direct / Hint &84.4 &74.8 &89.2 &78.6 &87.0 &87.8 \\
+ 3 Shots / CoT / No Hint &74.4 &69.2 &80.2 &73.4 &75.4 &83.0 \\
+ 3 Shots / CoT / Hint &82.4 &74.2 &88.8 &76.0 &81.8 &89.0 \\ \midrule
LLaMA 4 Scout &65.4 &54.1 &60.4 &56.0 &67.8 &64.2 \\
+ 1 Shot / Direct / No Hint &69.6 &56.4 &70.0 &72.8 &73.0 &70.8 \\
+ 1 Shot / Direct / Hint &74.0 &61.1 &86.0 &76.8 &79.8 &81.2 \\
+ 3 Shots / CoT / No Hint &71.6 &62.9 &78.2 &71.4 &75.8 &83.8 \\
+ 3 Shots / CoT / Hint &75.6 &68.6 &84.2 &72.8 &83.0 &88.0 \\ \midrule
Exaone 4.0 32B &61.2 &55.7 &58.8 &52.8 &37.4 &63.2 \\
+ 1 Shot / Direct / No Hint &74.8 &57.8 &63.4 &62.4 &48.8 &65.4 \\
+ 1 Shot / Direct / Hint &76.6 &73.5 &85.8 &69.6 &61.6 &80.0 \\
+ 3 Shots / CoT / No Hint &74.6 &60.2 &70.0 &70.4 &62.8 &77.2 \\
+ 3 Shots / CoT / Hint &75.2 &72.1 &85.8 &71.2 &71.6 &85.8 \\ \midrule
HyperClova X SEED Think 14B &51.6 &38.7 &44.0 &44.6 &41.2 &58.6 \\
+ 1 Shot / Direct / No Hint &63.8 &42.0 &53.6 &50.0 &39.0 &51.8 \\
+ 1 Shot / Direct / Hint &69.0 &43.0 &67.0 &59.4 &42.0 &58.2 \\
+ 3 Shots / CoT / No Hint &67.0 &63.7 &74.0 &64.6 &47.0 &84.6 \\
+ 3 Shots / CoT / Hint &68.6 &66.2 &81.0 &70.6 &50.0 &87.8 \\
\bottomrule
\end{tabular}}
\caption{Ablation - Effects of Hint. The first row denotes the zero-shot, direct, and no-hint configuration.}\label{tab:ablation-hint}
\end{table*}

%% file: sections/tables/table-ablation-cot.tex
\begin{table*}[!htp]\centering
\resizebox{\textwidth}{!}{ 
\begin{tabular}{l|ccc|ccc}\toprule
Model &En-MM &En-OP &En-TA &Ko-MM &Ko-OP &Ko-TA \\\midrule \midrule
QWEN3 32B &66.4 &52.7 &73.4 &67.6 &80.0 &79.6 \\
+ 1 Shot / Direct / Hint &84.4 &74.8 &89.2 &78.6 &87.0 &87.8 \\
+ 1 Shot / CoT / Hint &83.6 &75.8 &89.6 &77.8 &79.2 &87.8 \\
+ 3 Shots / Direct / Hint &82.4 &76.8 &90.2 &77.8 &86.8 &88.8 \\
+ 3 Shots / CoT / Hint &82.4 &74.2 &88.8 &76.0 &81.8 &89.0 \\ \midrule
LLaMA 4 Scout &65.4 &54.1 &60.4 &56.0 &67.8 &64.2 \\
+ 1 Shot / Direct / Hint &74.0 &61.1 &86.0 &76.8 &79.8 &81.2 \\
+ 1 Shot / CoT / Hint &77.2 &69.7 &84.4 &75.0 &80.8 &85.6 \\
+ 3 Shots / Direct / Hint &76.0 &68.2 &86.6 &73.6 &81.2 &87.2 \\
+ 3 Shots / CoT / Hint &75.6 &68.6 &84.2 &72.8 &83.0 &88.0 \\ \midrule
Exaone 4.0 32B &61.2 &55.7 &58.8 &52.8 &37.4 &63.2 \\
+ 1 Shot / Direct / Hint &76.6 &73.4 &85.8 &69.6 &61.6 &80.0 \\
+ 1 Shot / CoT / Hint &80.8 &74.0 &86.6 &67.6 &71.0 &79.6 \\
+ 3 Shots / Direct / Hint &73.6 &72.7 &83.6 &70.0 &64.8 &84.2 \\
+ 3 Shots / CoT / Hint &75.2 &72.1 &85.8 &71.2 &71.6 &85.8 \\ \midrule
HyperClova X SEED Think 14B &51.6 &38.7 &44.0 &44.6 &41.2 &58.6 \\
+ 1 Shot / Direct / Hint &69.0 &43.0 &67.0 &59.4 &42.0 &58.2 \\
+ 1 Shot / CoT / Hint &70.2 &51.4 &75.2 &63.8 &37.6 &84.6 \\
+ 3 Shots / Direct / Hint &67.0 &64.8 &83.8 &63.8 &49.6 &83.8 \\
+ 3 Shots / CoT / Hint &68.6 &66.2 &81.0 &70.6 &50.0 &87.8 \\

\bottomrule

\end{tabular}}
\caption{Ablation - Effects of CoT Examples. The first row denotes the zero-shot, direct, and no-hint configuration.}\label{tab:ablation-cot}
\end{table*}

%% file: sections/A04-SLM-Prompting.tex
\section{SLM Prompting}
\label{sec:slm-prompting}

\input{sections/tables/table-ablation-slm}

This section examines whether detailed prompting improves the reasoning capability of small language models (SLMs), which are increasingly adopted for their computational efficiency. Table~\ref{tab:ablation-slm} reports the performance of SLMs under various prompting configurations. For the smallest model, Qwen3 0.6B, we find that prompting the model with no examples yields the best results, while even simple in-context examples lead to an accuracy drop. For larger SLMs, the benefits of few-shot prompting are mixed. The Qwen3 1.7B model exhibits different optimal configurations across subtasks, whereas Exaone 4.0 1.2B benefits from chain-of-thought and hint-based prompting. Our largest SLMs, Exaone 3.5B Instruct and Qwen3 4B, generally achieve the highest accuracy when given few-shot chain-of-thought examples with hints, suggesting that larger models gain more from detailed in-context learning examples.

%% file: sections/tables/table-ablation-slm.tex
\begin{table*}[!htbp]\centering
\resizebox{\textwidth}{!}{ 
\begin{tabular}{l|ccc|ccc}\toprule
Models &En-MM &En-OP &En-TA &Ko-MM &Ko-OP &Ko-TA \\\midrule \midrule
QWEN3 0.6B &\textbf{59.2} &\textbf{52.0} &\textbf{38.8} &\ul{53.6} &\ul{28.6} &\textbf{38.0} \\
+ 1 Shot / Direct / No Hint &55.4 &\ul{50.8} &31.4 &\ul{53.6} &23.8 &\ul{34.0} \\
+ 3 Shots / Direct / No Hint &54.4 &48.8 &36.4 &\textbf{57.0} &\ul{28.6} &32.4 \\
+ 1 Shot / CoT / Hint &\ul{57.2} &46.7 &\ul{36.6} &52.0 &27.4 &32.4 \\
+ 3 Shots / CoT / Hint &55.4 &46.7 &33.0 &49.0 &\textbf{30.2} &33.4 \\ \midrule
Exaone 4.0 1.2B &55.6 &\ul{44.7} &49.8 &48.0 &29.0 &\textbf{44.8} \\
+ 1 Shot / Direct / No Hint &60.4 &\textbf{50.0} &47.8 &55.4 &27.2 &\ul{42.2} \\
+ 3 Shots / Direct / No Hint &61.0 &42.8 &45.8 &49.4 &27.2 &40.2 \\
+ 1 Shot / CoT / Hint &\textbf{66.8} &44.1 &\textbf{57.8} &\ul{58.0} &\ul{29.8} &41.4 \\
+ 3 Shots / CoT / Hint &\ul{64.8} &\ul{44.7} &\ul{54.8} &\textbf{58.6} &\textbf{31.6} &35.8 \\ \midrule
QWEN3 1.7B &60.6 &\ul{52.3} &\textbf{59.0} &50.0 &\ul{34.0} &\textbf{63.2} \\
+ 1 Shot / Direct / No Hint &64.0 &50.2 &54.4 &55.2 &33.4 &\ul{56.0} \\
+ 3 Shots / Direct / No Hint &59.6 &\textbf{53.1} &53.4 &\ul{58.0} &\textbf{34.8} &51.0 \\
+ 1 Shot / CoT / Hint &\textbf{70.8} &42.2 &\ul{58.2} &56.0 &30.8 &54.6 \\
+ 3 Shots / CoT / Hint &\ul{66.0} &39.6 &55.4 &\textbf{63.0} &33.0 &54.2 \\ \midrule
Exaone 3.5 2.4B Instruct &51.4 &41.6 &\textbf{44.0} &51.0 &30.0 &\ul{42.4} \\ 
+ 1 Shot / Direct / No Hint &56.2 &45.3 &\ul{42.6} &55.8 &40.0 &37.6 \\
+ 3 Shots / Direct / No Hint &56.8 &45.7 &38.4 &\textbf{58.0} &40.2 &39.4 \\
+ 1 Shot / CoT / Hint &\ul{61.4} &\ul{47.7} &39.6 &\ul{56.4} &\ul{40.4} &37.8 \\ 
+ 3 Shots / CoT / Hint &\textbf{63.4} &\textbf{48.6} &37.2 &56.0 &\textbf{41.2} &\textbf{42.8} \\ \midrule
QWEN3 4B &62.8 &55.5 &66.0 &59.0 &51.0 &75.4 \\
+ 1 Shot / Direct / No Hint &67.0 &53.3 &69.4 &63.4 &51.8 &74.0 \\
+ 3 Shots / Direct / No Hint &68.2 &49.0 &69.6 &\ul{66.4} &53.6 &69.8 \\
+ 1 Shot / CoT / Hint &\textbf{76.2} &\ul{69.3} &\textbf{86.2} &\textbf{69.8} &\textbf{56.0} &\ul{76.4} \\
+ 3 Shots / CoT / Hint &\ul{74.0} &\textbf{70.7} &\ul{85.4} &\textbf{69.8} &\ul{55.2} &\textbf{79.6} \\
\bottomrule
\end{tabular}}
\caption{Ablation - SLMs. The first row denotes the zero-shot, direct, and no-hint configuration. Boldface indicates the highest numbers, and underlined
numbers indicate the second highest.}\label{tab:ablation-slm}
\end{table*}

%% file: sections/A05-Benchmark-Eval-Method.tex
\section{Evaluation Configuration}
\label{sec:benchmark-eval-method}

We use \texttt{transformers}~\cite{wolf2020transformers}, \texttt{lm-evaluation-harness}~\cite{eval-harness} and \texttt{vLLM}~\cite{kwon2023vllm} frameworks for model evaluation. All models are evaluated in both English and Korean settings to ensure consistent evaluation configurations across languages. As described in Section~\ref{sec:evaluation}, we adopt the \textit{3-Shot CoT Hint} configuration for all evaluations. The test splits of both MuSR and Ko-MuSR benchmarks are used for assessment, while the \textit{prompt split}—a separate subset excluded from the test set—is used for few-shot examples. Since our models support chat templates, we apply the corresponding templates when providing in-context learning examples. For Ko-MuSR, we release the prompt split consisting of three examples and their corresponding solutions.

We provide evaluation prompts for 3-Shot, CoT, Hint configuration as we find this configuration beneficial for models to increase accuracy. See Appendix~\ref{sec:prompt-details} for prompt details.

%% file: sections/A10-Computing.tex
\section{Computation}
\label{sec:eval-computation}
\input{sections/tables/table-appendix-computing}

We conduct all experiments on an on-premise GPU cluster consisting of eight nodes, each equipped with four NVIDIA RTX 3090 GPUs. The GPUs within each node are interconnected via a PCIe Gen4, and the nodes are connected through an NVIDIA Mellanox InfiniBand EDR (100 Gbps) network. We employ \texttt{vLLM} to build a multi-GPU, multi-node inference system for model evaluation. Within a node, model parameters are distributed with tensor parallelism while we utilize pipeline parallelism across nodes. Table~\ref{tab:computing} summarizes the configuration of the inference system and reports the elapsed time for evaluating the \textit{3-Shot CoT Hint} configuration. With repetitions of experiments, we report 3{,}849 GPU Hours(RTX 3090) in total.

Upon release on the Hugging Face Hub\footnote{\url{https://huggingface.co}}, HyperClovaX SEED Think 14B model did not include support for tensor parallelism. We modified its configuration file to enable tensor parallel inference. To validate our implementation, we evaluated the model on the Hellaswag~\cite{zellers-etal-2019-hellaswag} benchmark. While the original authors reported an accuracy of 61.25, our reproduced result of 64.04 confirms the correctness of our implementation.

%% file: sections/tables/table-appendix-computing.tex
\begin{table*}[!htbp]\centering
\resizebox{\textwidth}{!}{ 
\begin{tabular}{l|cccc|ccc|ccc}\toprule
Models &Num nodes & Num GPUs &TP &PP &En-MM &En-OP &En-TA &Ko-MM &Ko-OP &Ko-TA \\\midrule
QWEN3 32B &2 &8 &4 &2 &205 &127 &329 &224 &174 &307 \\
LLaMA 4 Scout Instruct &4 &16 &4 &4 &105 &55 &136 &109 &82 &180 \\
Exaone 4.0 32B &2 &8 &4 &2 &92 &76 &120 &106 &71 &118 \\
HyperClova X SEED Think 14B &1 &4 &4 &1 &112 &108 &108 &112 &102 &109 \\ \midrule
QWEN3 0.6B &1 &1 &1 &1 &38 &15 &44 &49 &17 &50 \\
Exaone 4.0 1.2B &1 &1 &1 &1 &12 &12 &20 &15 &14 &20 \\
QWEN3 1.7B &1 &1 &1 &1 &121 &40 &120 &99 &56 &163 \\
Exaone 3.5 2.4B Instruct &1 &1 &1 &1 &106 &100 &104 &96 &93 &97 \\
QWEN 3 4B &1 &2 &2 &1 &145 &74 &158 &142 &126 &224 \\
\bottomrule
\end{tabular}}
\caption{Number of GPus and Time(in minutes) Required for Evaluation with 3-Shot, CoT, Hint Configuration}
\label{tab:computing}
\end{table*}

%% file: sections/A11-Annotation.tex
\section{Data Review Process}
\label{sec:annotation}

As described in Section ~\ref{sec:komusr}, we review all data instances to ensure the validity of the question and the quality of data instances. 
Although the synthesized texts are unlikely to contain personally identifiable information or unsafe content, annotators carefully confirmed that such content was excluded from all instances; otherwise, they left notes for senior reviewers to examine. 

Specific criteria for marking \textit{pass}, \textit{minor issue} and \textit{fail} are as follows:

\begin{itemize}
    \item \textbf{Pass}: The answer can be derived from the narrative and has no issue to be used as a benchmark instance.
    \item \textbf{Minor Issue}: The instance contains small issues (e.g., typos, unclear wording) but it seems to be usable with minor edits.
    \item \textbf{Fail}: The instance has critical problems (e.g., logical errors, ambiguous answers) and cannot be used without significant edits.  
\end{itemize}

See Table~\ref{tab:annotation-instruction} for the annotation guidelines given to the annotators.

%% file: sections/A12-IRB.tex
\section{Measuring Human Performance}
\label{sec:irb}

\subsection{Instructions}

We evaluate human performance using the 1-shot CoT Hint configuration. Each participant is provided with one example per subtask, where the example includes both a question and a sample answer with step-by-step reasoning. Participants are then asked to solve 40 questions per section.  
We exclude 9 questions from the \textit{murder mysteries} section and 2 questions from the \textit{object placements} section, as these instances were later removed from the benchmark after we tightened the data quality standards.  
 Appendix~\ref{sec:irb-examples} provides examples and example solutions that were provided to the participants. For detailed instructions, see Appendix~\ref{sec:misc}.

\subsection{IRB Approval}

We obtained approval from the Institutional Review Board (IRB) to conduct the human evaluation for Ko-MuSR, as our study involves human subjects. This process ensures full compliance with research ethics and participant protection standards.  
Following the IRB-approved protocol, we recruited five voluntary participants. All participants were fully informed that their responses would be anonymized and their performance scores would be used for statistical analysis and publication.  
Table~\ref{tab:irb-consent} presents key disclaimers provided to participants.

%% file: sections/A91-ARR-Checklist.tex
\section{Checklist for ARR Submission}

\label{sec:arr-checklist}

\paragraph{(A1) Limitations of the work}

Please refer to the Limitations section of the main text.

\paragraph{(A2) Potential risks of the work}

Our benchmark is synthesized with OpenAI APIs and therefore is unlikely to contain any personally identifiable information, unsafe content, or materials infringing copyrights. Our human annotators reviewed all data instances to ensure the above-stated content is excluded from the benchmark.


\paragraph{(B1) Citations of the used artifacts}
We tried our best to cite all papers, code repositories, and resources we used in the main content of the paper.


\paragraph{(B2) License or terms of use of the artifacts}

\textbf{Benchmarks.}
The MuSR dataset is licensed under MIT. We synthesize our data with the artifact that MuSR releases. We disclose our code and data under the MIT license as well.

\paragraph{(B3) Proper use of existing artifacts and Intended use of created artifacts} 

The distributed MuSR artifact and the benchmark are utilized properly. We use the code for creating custom data as authors intend to open the code. We use their test set for model evaluation. As mentioned earlier, use of artifacts complies with the authors' intended use of artifacts.
Following MuSR, we will release the code and the data to enable future researchers to create data instances and evaluate Korean-capable language models.

\paragraph{(B4) Description of steps for removing personal identifiable information(PII) and offensive contents from data}

Our benchmark is synthesized with OpenAI APIs and therefore is unlikely to contain any personally identifiable information, unsafe content, or materials infringing copyrights. Our human annotators reviewed all data instances to ensure the above-stated content is excluded from the benchmark.

\paragraph{(B5) Documentation of the artifacts}
Our data for the benchmark is available at https://huggingface.co/datasets/thunder-research-group/Ko-MuSR, along with the dataset card. Our code is available at https://github.com/mcrl/Ko-MuSR.

\paragraph{(B6) Statistics for the data}

See Table \ref{tab:benchmark-stats}.

\paragraph{(C1) Descriptions of the number of parameters in the model, the total computational budget, and computing infrastructure}

We utilize open-parameter models and cite them for describing the details.
We describe our computing infrastructure and total computational budget in Appendix \ref{sec:eval-computation}.

\paragraph{(C2) Details of experimental setup}
For experiment prompts, see Section \ref{sec:evaluation} for the overview and Appendix \ref{sec:prompt-details}.
For model and the generation configuration, see Section \ref{sec:evaluation}. For system setup for model inference, see Appendix \ref{sec:eval-computation}.

\paragraph{(C3) Descriptive statistics about results}

We repeat all experiments twice and report the mean of the runs. We report all results from repeated runs in Appendix \ref{sec:misc}.


\paragraph{(C4) Packages/Software Versions}

\paragraph{Data Synthesis} We use the original MuSR codebase (latest version as of 2025.03.01, commit \texttt{b1f4d4})~\cite{musr} for dataset synthesis, and OpenAI Python SDK (v1.78.0)~\cite{openai2025sdk} for API-based data generation.

\paragraph{Evaluation} We use \texttt{torch} (v2.7.1)~\cite{paszke2019pytorch} and \texttt{transformers} (v4.55.2)~\cite{wolf2020transformers} for model implementation, 
\texttt{vllm} (v0.10.0)~\cite{kwon2023vllm} for multi-gpu, multi-node inference,
\texttt{flash-attn} (v2.7.4)~\cite{dao2022flashattention} for inference-time attention optimization, 
and \texttt{lm-eval-harness} (v0.4.8)~\cite{eval-harness} for benchmark evaluation.
Our modification for enabling model inference is described in Appendix ~\ref{sec:eval-computation}.


\paragraph{(D1) Full text of instructions or disclaimers of any risks}
\paragraph{Annotation} See Appendix~\ref{sec:annotation} for the annotation(data review) process. See Table~\ref{tab:annotation-instruction} for annotation instructions.

\paragraph{Human Evaluation} See Appendix~\ref{sec:irb} for the human evaluation process. See Table~\ref{tab:irb-consent} for the consent form we have provided. See Appendix~\ref{sec:irb-examples} for the full text of instructions and the question examples they were given.

\paragraph{(D2) Recruitment process and payment of paid participants}
The authors, as members of the research group, participated in reviewing the Korean benchmark. Their work was supported by the grants listed in the Acknowledgements section. Compensation for their contribution was acknowledged at KRW 10,030 per hour, corresponding to the 2025 minimum wage in Korea.

We evaluated the human performance on the benchmark. All evaluation procedures were conducted under the supervision of the Institutional Review Board of Seoul National University (SNUIRB, No. 2506/001-001). Participants’ contributions were recognized as one full day (8 hours) of work and compensated with KRW 100,000, exceeding the 2025 minimum daily wage of KRW 80,240 in Korea.

\paragraph{(D3) Consent from the used/curated data}
The annotators, including the authors, are the members of the research group. All participants were fully notified that their annotations would be used to construct a Korean Benchmark. The participants for human evaluations were fully notified that overall statistics of their scores would be published. See Appendix \ref{sec:irb} for the details.

\paragraph{(D4) Review of data collection protocol by an ethics review board}
This study has been reviewed and approved by the Institutional Review Board (SNUIRB, No. 2506/001-001), in compliance with the national regulations on research ethics. Therefore, the data collection protocol fully satisfies the ethical standards for involving human participants. 
The participants in this study were recruited as healthy adults, and informed consent was obtained through an official IRB-approved consent form. All participants were explicitly informed of the study’s purpose, procedures, risks, benefits, and their rights to withdraw at any time without penalty. Their responses were collected anonymously, and no personally identifiable information was included in the dataset.

\paragraph{(D5) Basic demographic and geographic characteristics of the annotator population}

All annotators are authors or members of the research group. 7 annotators are involved in creating the Korean benchmark. All annotators are Korean. All annotators are Asian, native to Korean, aged from 20 to 30 (adults).

5 human participants involving human evaluation are students in the organization, whose ages range from 20 to 30.


\paragraph{(E1) Use of AI assistants}

We do not use AI assistants for this work.


%% file: sections/A92-Prompts.tex
\section{Prompt Description}
\label{sec:prompt-details}

\subsection{System Prompt}

All models were given the following system prompts per task. For English test instances, we use the English system prompt, while we use Korean system prompts otherwise. See Figures ~\ref{fig:system-ko} and ~\ref{fig:system-en}.

\subsection{Evaluation Prompt}

\subsubsection{Hints}

Listings are hints provided for the evaluation. Note that hints for Korean settings are the translated version of the English counterpart. See Figures from ~\ref{fig:hint-komm} to ~\ref{fig:hint-enta}.

\subsubsection{Evaluation Prompt Structure}

Listings are prompt structures utilized for the evaluation. Note that the prompt structure for Korean settings is a translated version of the English counterpart. See Figures from ~\ref{fig:struct-komm-direct} to ~\ref{fig:struct-enta-cot}.

\subsection{Data Synthesis Prompt}

See Figures from ~\ref{fig:prompt-mm-expand-ko} to ~\ref{fig:prompt-ta-gen-en}.

\onecolumn
\input{sections/tables/prompt_textboxes/system-enko}

\input{sections/tables/prompt_textboxes/hint-komm}
\input{sections/tables/prompt_textboxes/hint-enmm}
\input{sections/tables/prompt_textboxes/hint-koop}
\input{sections/tables/prompt_textboxes/hint-enop}
\input{sections/tables/prompt_textboxes/hint-kota}
\input{sections/tables/prompt_textboxes/hint-enta}

\input{sections/tables/prompt_textboxes/structure-komm-direct}
\input{sections/tables/prompt_textboxes/structure-enmm-direct}
\input{sections/tables/prompt_textboxes/structure-komm-cot}
\input{sections/tables/prompt_textboxes/structure-enmm-cot}

\input{sections/tables/prompt_textboxes/structure-koop-direct}
\input{sections/tables/prompt_textboxes/structure-enop-direct}
\input{sections/tables/prompt_textboxes/structure-koop-cot}
\input{sections/tables/prompt_textboxes/structure-enop-cot}

\input{sections/tables/prompt_textboxes/structure-kota-direct}
\input{sections/tables/prompt_textboxes/structure-enta-direct}
\input{sections/tables/prompt_textboxes/structure-kota-cot}
\input{sections/tables/prompt_textboxes/structure-enta-cot}

\input{sections/tables/synthesis-prompts/komm-expansion}
\input{sections/tables/synthesis-prompts/komm-expansion-translated}
\input{sections/tables/synthesis-prompts/komm-generation}
\input{sections/tables/synthesis-prompts/komm-generation-translated}
\input{sections/tables/synthesis-prompts/koop-expansion}
\input{sections/tables/synthesis-prompts/koop-expansion-translated}
\input{sections/tables/synthesis-prompts/koop-generation}
\input{sections/tables/synthesis-prompts/koop-generation-translated}
\input{sections/tables/synthesis-prompts/kota-expansion}
\input{sections/tables/synthesis-prompts/kota-expansion-translated}
\input{sections/tables/synthesis-prompts/kota-generation}
\input{sections/tables/synthesis-prompts/kota-generation-translated}

\twocolumn

%% file: sections/tables/prompt_textboxes/system-enko.tex
\begin{hintbox}{System Prompt(Korean)}{fig:system-ko}
당신은 사용자가 제시한 질문에 답변하는 친절한 도우미입니다.\\

다음은 이 태스크를 해결하는 예시입니다:
\end{hintbox}

\begin{hintbox}{System Prompt(English)}{fig:system-en}

You are a helpful assistant that will answer the questions given by the user.\\

Here is an example of solving the task:
\end{hintbox}

%% file: sections/tables/prompt_textboxes/hint-komm.tex
\begin{hintbox}{Hint- Murder Mysteries (Korean)}{fig:hint-komm}
살인자는 반드시 세 가지 조건을 충족해야 합니다:

동기(피해자를 죽일 만한 이유가 있는가), 수단(범행 도구에 접근할 수 있는가), 기회(범행 현장에 접근할 수 있었는가)

무고한 용의자는 이 세 가지 중 두 가지까지만 입증될 수 있으며, 의심스러운 행동을 보일 수는 있어도 세 가지 조건을 모두 갖추지는 않습니다. 만약 두 용의자 모두 수단, 동기, 기회를 가진 것처럼 보인다면, 그 중에서도 가장 명확하게 이 세 가지가 드러난 인물을 선택해야 합니다. 반대로, 어느 누구도 세 가지를 모두 갖추지 않은 경우라면, 세 가지 조건 중 가장 뚜렷하게 드러난 인물을 선택해야 합니다.
\end{hintbox}

%% file: sections/tables/prompt_textboxes/hint-enmm.tex
\begin{hintbox}{Hint - Murder Mysteries (English)}{fig:hint-enmm}
The murderer needs to have a means (access to weapon), motive (reason to kill the victim), and opportunity (access to crime scene) in order to have killed the victim. Innocent suspects may have two of these proven, but not all three. An innocent suspect may be suspicious for some other reason, but they will not have all of motive, means, and opportunity established.\\

If you believe that both suspects have motive, means, and opportunity, you should make an educated guess pick the one for whom these are best established. If you believe that neither suspect has all three established, then choose the suspect where these are most clearly established.
\end{hintbox}

%% file: sections/tables/prompt_textboxes/hint-koop.tex
\begin{hintbox}{Hint - Object Placements (Korean)}{fig:hint-koop}
이야기를 바탕으로, 우리는 특정 인물이 어떤 물건이 어디에 있다고 생각하고 있는지를 파악하려고 합니다.

이를 위해서는 이야기 읽으며 각 인물이 그 물건의 위치를 시점별로 어떻게 인식하고 있는지를 추적해야 합니다.

물건이 옮겨질 때, 인물이 그 장면을 직접 보았거나, 옮겨진 위치를 인지할 수 있는 상태였다면 그들은 물건이 옮겨진 위치를 알고 있다고 봅니다.

하지만 물건이 움직일 때 인물이 그 장면을 보지 못했거나 다른 일에 집중하고 있었다면, 여전히 마지막으로 본 위치에 있다고 믿게 됩니다.
\end{hintbox}

%% file: sections/tables/prompt_textboxes/hint-enop.tex
\begin{hintbox}{Hint - Object Placements (English)}{fig:hint-enop}
Based on this story, we want to identify where someone believes that a certain object is at the end of the story. In order to do that, you need to read the story and keep track of where they think the object is at each point. When an object is moved, the person may observe its new location if they saw it move.\\

To see where an object ends up, they must be able to see the location that it moves to and not be too distracted by what they are doing. If they do not observe the object moving, then they will still believe it to be in the last location where they observed it.
\end{hintbox}

%% file: sections/tables/prompt_textboxes/hint-kota.tex
\begin{hintbox}{Hint - Team Allocations (Korean)}{fig:hint-kota}
스토리를 통해 각 사람이 한 가지 업무에 얼마나 능숙한지를 파악할 수 있습니다. 일반적으로 각 사람은 어떤 작업에 대해 뛰어나거나, 보통이거나, 부족한 실력을 가지고 있습니다. 우리는 가능한 한 각자의 강점을 최대한 발휘할 수 있도록 사람을 적절한 업무에 배치하려고 합니다.\\

또한, 두 사람이 함께 맡아야 하는 업무가 하나 있으며, 이 경우 두 사람의 팀워크 수준(훌륭함, 보통, 나쁨)도 전체 업무 성과에 중요한 영향을 미칩니다.\\

단, 두 사람이 함께 일해야 하는 작업에서 한 사람이 실력이 부족하고, 두 사람의 팀워크도 좋지 않다면, 다른 한 사람이 아무리 뛰어나도 전체 결과에는 도움이 되지 않을 수 있습니다.\\

각기 다른 강점, 약점, 그리고 두 사람 간의 상호작용을 고려하여, 전체 과제가 가장 효율적으로 수행될 수 있도록 팀원들을 적절히 배치해야 합니다.
\end{hintbox}

%% file: sections/tables/prompt_textboxes/hint-enta.tex
\begin{hintbox}{Hint - Team Allocations (English)}{fig:hint-enta}
The story should allow you to determine how good each person is at a skill. Roughly, each person is either great, acceptable, or bad at a task. We want to find an optimal assignment of people to tasks that uses their skills as well as possible. In addition, one task will have to have two people assigned to it. The effectiveness of their teamwork (great team, acceptable team, or bad team) also impacts the overall quality of the assignment.\\

When two people need to work on a task and one is bad at it, they don’t necessarily benefit from the other person being good, unless they work well together.\\

With different strengths, weaknesses, and interpersonal dynamics at play, you should allocate your team to find the single assignment to ensure that the tasks overall are completed as effectively as possible.
\end{hintbox}

%% file: sections/tables/prompt_textboxes/structure-komm-direct.tex
\begin{hintbox}{Direct Prompt Structure for Murder Mysteries (Korean)}{fig:struct-komm-direct}
\textit{\{\{narrative\}\}}\\

다음 중 살인자일 가능성이 가장 높은 사람은 누구인가?\\

다음 선택지 중 하나를 고르세요:\\
1 - Choice 1\\
2 - Choice 2\\

반드시 하나의 선택지만 고르셔야 합니다. \\

\textit{\{\{If hint is given, hint will be provided here.\}\}}\\

마지막에 아래와 같은 형식으로 정답을 작성하세요: '정답: X'
\end{hintbox}

%% file: sections/tables/prompt_textboxes/structure-enmm-direct.tex
\begin{hintbox}{Direct Prompt Structure for Murder Mysteries (English)}{fig:struct-enmm-direct}

\textit{\{\{narrative\}\}}\\

Who is the most likely murderer?\\

Choose one of the following options:\\
1 - Choice 1\\
2 - Choice 2\\

You must pick one option.\\

\textit{\{\{If hint is given, hint will be provided here.\}\}}\\

Finally, the last thing you generate should be "ANSWER: (your answer here, including the choice number)"
\end{hintbox}

%% file: sections/tables/prompt_textboxes/structure-komm-cot.tex
\begin{hintbox}{CoT Prompt Structure for Murder Mysteries (Korean)}{fig:struct-komm-cot}
\textit{\{\{narrative\}\}}\\

다음 중 살인자일 가능성이 가장 높은 사람은 누구인가?\\

다음 선택지 중 하나를 고르세요:\\
1 - Choice 1\\
2 - Choice 2\\

반드시 하나의 선택지만 고르셔야 합니다. 선택하기 전에, 단계별로 당신의 추론 과정을 설명하세요. \\

\textit{\{\{If hint is given, hint will be provided here.\}\}}\\

단계별로 논리적인 추론을 마친 후, 마지막에 아래와 같은 형식으로 정답을 작성하세요: '정답: X'
\end{hintbox}

%% file: sections/tables/prompt_textboxes/structure-enmm-cot.tex
\begin{hintbox}{CoT Prompt Structure for Murder Mysteries (English)}{fig:struct-enmm-cot}
\textit{\{\{narrative\}\}}\\

Who is the most likely murderer?\\

Choose one of the following options:\\
1 - Choice 1\\
2 - Choice 2\\

You must pick one option. Before selecting a choice, explain your reasoning step by step.\\

\textit{\{\{If hint is given, hint will be provided here.\}\}}\\

Explain your reasoning step by step before you answer. Finally, the last thing you generate should be "ANSWER: (your answer here, including the choice number)"
\end{hintbox}

%% file: sections/tables/prompt_textboxes/structure-koop-direct.tex
\begin{hintbox}{Direct Prompt Structure for Object Placements (Korean)}{fig:struct-koop-direct}
\textit{\{\{narrative\}\}}\\

주어진 이야기를 고려할 때, \{character\}는 \{object\}을 찾기 위해 어디를 가장 먼저 확인하겠는가?\\

\textit{\{\{If hint is given, hint will be provided here.\}\}}\\

다음 선택지 중 하나를 고르세요:\\
1 - Choice 1\\
2 - Choice 2\\
3 - Choice 3\\
4 - Choice 4\\

반드시 하나의 보기를 선택해야 합니다. 마지막에 다음 형식으로 정답을 출력하세요: '정답: X'
\end{hintbox}

%% file: sections/tables/prompt_textboxes/structure-enop-direct.tex
\begin{hintbox}{Direct Prompt Structure for Object Placements (English)}{fig:struct-enop-direct}
\textit{\{\{narrative\}\}}\\

Which location is the most likely place \{character\} would look to find the \{object\} given the story?\\

\textit{\{\{If hint is given, hint will be provided here.\}\}}\\

Choose one of the following options:\\
1 - Choice 1\\
2 - Choice 2\\
3 - Choice 3\\
4 - Choice 4\\

You must pick one option. Finally, the last thing you generate should be "ANSWER: (your answer here, include the choice number)"
\end{hintbox}

%% file: sections/tables/prompt_textboxes/structure-koop-cot.tex
\begin{hintbox}{CoT Prompt Structure for Object Placements (Korean)}{fig:struct-koop-cot}
\textit{\{\{narrative\}\}}\\

주어진 이야기를 고려할 때, \{character\}는 \{object\}을 찾기 위해 어디를 가장 먼저 확인하겠는가?\\

\textit{\{\{If hint is given, hint will be provided here.\}\}}\\

다음 선택지 중 하나를 고르세요:\\
1 - Choice 1\\
2 - Choice 2\\
3 - Choice 3\\
4 - Choice 4\\

반드시 하나의 보기를 선택해야 합니다. 추론 과정을 단계별로 설명한 후, 마지막에 다음 형식으로 정답을 출력하세요: '정답: X'
\end{hintbox}

%% file: sections/tables/prompt_textboxes/structure-enop-cot.tex
\begin{hintbox}{CoT Prompt Structure for Object Placements (English)}{fig:struct-enop-cot}
\textit{\{\{narrative\}\}}\\

Which location is the most likely place \{character\} would look to find the \{object\} given the story?\\

\textit{\{\{If hint is given, hint will be provided here.\}\}}\\

Choose one of the following options:\\
1 - Choice 1\\
2 - Choice 2\\
3 - Choice 3\\
4 - Choice 4\\

You must pick one option. Explain your reasoning step by step, then output the answer in the form 'ANSWER: X'
\end{hintbox}

%% file: sections/tables/prompt_textboxes/structure-kota-direct.tex
\begin{hintbox}{Direct Prompt Structure for Team Allocations (Korean)}{fig:struct-kota-direct}
\textit{\{\{narrative\}\}}\\

이 이야기를 바탕으로, 두 가지 업무가 모두 효율적으로 수행되도록 사람을 어떻게 배정하시겠습니까?\\

다음 선택지 중 하나를 고르세요:\\
1 - Choice 1\\
2 - Choice 2\\
3 - Choice 3\\

당신은 세 가지 중 하나의 선택지를 골라야 합니다.\\

\textit{\{\{If hint is given, hint will be provided here.\}\}}\\

마지막에는 아래 형식으로 답변을 마무리해 주세요:  "정답: (당신의 선택, 번호 포함)"
\end{hintbox}

%% file: sections/tables/prompt_textboxes/structure-enta-direct.tex
\begin{hintbox}{Direct Prompt Structure for Team Allocations (English)}{fig:struct-enta-direct}
\textit{\{\{narrative\}\}}\\

Given the story, how would you uniquely allocate each person to make sure both tasks are accomplished efficiently?\\

Choose one of the following options:\\
1 - Choice 1\\
2 - Choice 2\\
3 - Choice 3\\

You must pick one option.\\

\textit{\{\{If hint is given, hint will be provided here.\}\}}\\

Finally, the last thing you generate should be "ANSWER: (your answer here, including the choice number)"
\end{hintbox}

%% file: sections/tables/prompt_textboxes/structure-kota-cot.tex
\begin{hintbox}{CoT Prompt Structure for Team Allocations (Korean)}{fig:struct-kota-cot}
\textit{\{\{narrative\}\}}\\

이 이야기를 바탕으로, 두 가지 업무가 모두 효율적으로 수행되도록 사람을 어떻게 배정하시겠습니까?\\

다음 선택지 중 하나를 고르세요:\\
1 - Choice 1\\
2 - Choice 2\\
3 - Choice 3\\

당신은 세 가지 중 하나의 선택지를 골라야 합니다.\\

\textit{\{\{If hint is given, hint will be provided here.\}\}}\\

당신의 풀이 과정을 단계별로 설명한 뒤, 마지막에는 아래 형식으로 답변을 마무리해 주세요:  "정답: (당신의 선택, 번호 포함)"
\end{hintbox}

%% file: sections/tables/prompt_textboxes/structure-enta-cot.tex
\begin{hintbox}{CoT Prompt Structure for Team Allocations (English)}{fig:struct-enta-cot}
\textit{\{\{narrative\}\}}\\

Given the story, how would you uniquely allocate each person to make sure both tasks are accomplished efficiently?\\

Choose one of the following options:\\
1 - Choice 1\\
2 - Choice 2\\
3 - Choice 3\\

You must pick one option.\\

\textit{\{\{If hint is given, hint will be provided here.\}\}}\\

Explain your reasoning step by step before you answer. Finally, the last thing you generate should be "ANSWER: (your answer here, including the choice number)"
\end{hintbox}

%% file: sections/tables/synthesis-prompts/komm-expansion.tex
\begin{hintbox}{Prompts for Narrative Generation - Murder Mysteries}{fig:prompt-mm-expand-ko}

당신의 임무는 예시에 제시된 방식으로 이야기를 위한 논리 트리를 생성하는 것입니다. \\
이 트리에서는 각 사실이 자신의 자식 노드로부터 논리적으로 도출되어야 합니다. \\
이미 이름이 있는 사실(루트 노드 등)이 있을 경우, 새롭게 덮어쓰지 말아야 합니다.\\

이야기의 종류:\\

우리는 살인 미스터리를 만들고 있습니다. 살인 미스터리에는 용의자가 살인을 저지를 수 있는 '범행 수단, 범행 동기, 범행 기회'가 뒤얽힌 복잡한 증거망이 필요합니다. \\
이야기는 형사의 시점에서 쓰여야 합니다. 증거를 모으는 과정은 보통의 수사 방식(심문, 대화 엿듣기, 범죄 기록 조회, 우편물·쓰레기 조사 등)을 통해 이뤄집니다.\\

1. 트리에 등장하는 각 사실은 자식 노드로부터 논리적으로 도출되어야 합니다.\\
2. 모든 "스토리에서 추출된 사실" 노드와 "상식" 노드는, 결과적으로 만들어지는 추론에 밀접하게 관련되어야 합니다.\\
3. 각 루트 사실(root fact)에는 출처가 표시됩니다("스토리에서 추출된 사실" 혹은 "상식").\\
4. "스토리에서 추출된 사실"은 이야기 속 캐릭터, 장소, 물건에 대한 구체적 진술이어야 합니다.\\
5. "상식"은 대부분의 사람이 알고 동의할 만한 사실이어야 합니다. 이야기 속 특정 인물이나 사건을 직접적으로 언급하면 안 됩니다.\\
6. "상식"은 형식상 '규칙(rule)' 역할을 하며, 형제 노드의 사실을 결합해 상위 사실을 도출하는 용도로 사용됩니다.\\
7. 생성되는 트리는 '완성해야 할 함의 추론'의 트리 구조와 일치해야 합니다.\\
8. '완성해야 할 함의 추론' 트리의 "스토리에서 추출된 사실" 요청과 "상식"에 대해 한 개씩의 노드를 생성해야 합니다.\\
9. '완성해야 할 함의 추론' 트리에서 요청되지 않은 "스토리에서 추출된 사실"이나 "상식" 노드를 생성하지 마세요.\\

용의자가 '범행 수단'을 가졌다는 것은 살해 도구에 접근할 수 있음을 의미합니다.\\
용의자가 '범행 동기'를 가졌다는 것은 그(또는 그녀)가 피해자를 죽일 이유가 있음을 의미합니다.\\
용의자가 '범행 기회'를 가졌다는 것은 범행 현장에 있었음을 의미합니다.\\

설령 황당하거나 비현실적인 설정의 부모 노드 사실이라도, 그 사실을 유지하고 논리 트리를 구성해야 합니다.\\

예제를 확인하세요.\\

시나리오:\\

피해자: 유리\\
범행 장소: 집\\
살해 도구: 총\\
용의자: 민수\\
용의자의 역할: 형제\\
범행 동기: 금전적 이익\\

현재 트리: \\
민수는 살인자다. | 추론된 결론 \\
> 민수는 범행 수단이 있다. | 추론된 사실 \\
> > 민수는 총 쏘는 연습을 했다. | 스토리에서 추출된 사실 \\
> > 민수는 총을 소유하고 있다. | 스토리에서 추출된 사실 \\
> > 총을 소유하고 있으며 사용법을 연습했다면, 살인을 저지를 능력이 있다. | 상식 \\
> 민수는 범행 동기가 있다. | 추론된 사실 \\
> > 민수는 극도로 돈에 절박했다. | 스토리에서 추출된 사실 \\
> > 민수는 유리의 돈에 극도로 절박했다. | 스토리에서 추출된 사실 \\
> > 누군가가 극도로 절박할 때, 목표를 이루기 위해 극단적인 방법을 사용할 수 있으며, 여기에는 살인도 포함될 수 있다. | 상식 \\
> 민수는 범행 기회가 있다. | 스토리에서 추출된 사실 \\

완성해야 할 함의 추론:\\
민수는 범행 기회가 있다.\\
> 스토리에서 추출된 사실\\
> 상식\\

출력:\\
민수는 범행 기회가 있다.\\
> 민수는 유리의 집에 출입할 수 있다. | 스토리에서 추출된 사실\\
> 누군가의 집에 출입할 수 있다는 것은 그들을 살해할 범행 기회를 제공한다. | 상식\\
또 다른 예제입니다.\\

시나리오:\\

피해자: 철수\\
범행 장소: 경마장\\
살해 도구: 삽\\
용의자: 지민\\
용의자의 역할: 러닝 파트너\\
범행 동기: 다른 사람에게 해가 가는 것을 막기 위해\\

현재 트리: \\
지민이는 살인자다. | 추론된 결론 \\
> 지민이는 범행 수단이 있다. | 추론된 사실 \\
> > 지민이는 농부다 | 스토리에서 추출된 사실 \\
> > 농부들은 일반적으로 삽과 같은 원예 도구를 사용한다. | 상식 \\
> 지민이는 범행 동기가 있다. | 스토리에서 추출된 사실 \\
> 지민이는 범행 기회가 있다. | 스토리에서 추출된 사실 \\

완성해야 할 함의 추론: \\
지민이는 범행 동기가 있다. \\
> 스토리에서 추출된 사실 \\
> 스토리에서 추출된 사실 \\
> 상식 \\

출력:\\
지민이는 범행 동기가 있다.\\
> 지민이는 민수를 깊이 사랑한다. | 스토리에서 추출된 사실\\
> 철수는 민수를 위협했다. | 스토리에서 추출된 사실\\
> 깊고 강렬한 사랑은 사랑하는 사람이 위협받을 때 극단적인 행동, 심지어 살인까지 저지르게 할 수 있다. | 상식\\
또 다른 예제입니다.\\

시나리오:\\

피해자: 준호\\
범행 장소: 공원 벤치\\
살해 도구: 헤로인 과다복용\\
용의자: 민철\\
용의자의 역할: 마약 사용자\\
범행 동기: 공개적인 굴욕\\

현재 트리:\\
민철이는 살인자다. | 추론된 결론\\
> 민철이는 범행 수단이 있다. | 스토리에서 추출된 사실\\
> 민철이는 범행 동기가 있다. | 스토리에서 추출된 사실\\
> 민철이는 범행 기회가 있다. | 스토리에서 추출된 사실\\

완성해야 할 함의 추론:\\
민철이는 범행 수단이 있다.\\
> 스토리에서 추출된 사실\\
> 스토리에서 추출된 사실\\
> 상식\\

출력:\\
민철이는 범행 수단이 있다.\\
> 민철이는 헤로인에 접근할 수 있다. | 스토리에서 추출된 사실\\
> 민철이는 헤로인 과다복용에 필요한 양을 알고 있다. | 스토리에서 추출된 사실\\
> 헤로인에 접근할 수 있고, 치명적인 양을 알고 있다면, 피해자에게 의도적으로 치명적인 용량을 투여하여 살인을 저지를 수 있다. | 상식\\

이제 당신의 차례입니다.\\

시나리오:\\
피해자: 영은\\
범행 장소: 자동차\\
살해 도구: 고기 손질용 도끼(미트클리버)\\
용의자: 예성\\
용의자의 역할: 목수\\
범행 동기: 병적인 집착\\

현재 트리: \\
예성 | 추론된 결론 \\
> 예성이는 범행 수단이 있다. | 추론된 사실 \\
> > 예성이는 평소에 고기를 손질할 때 도끼를 사용한다.  | 추론된 사실 \\
> > > 예성은 육류 손질을 위해 고기 손질용 도끼를 사용한다.  | 추론된 사실 \\
> > > > 예성은 정기적으로 육류를 요리한다.   | 스토리에서 추출된 사실 \\
> > > > 예성은 고기 손질용 도끼를 소유하고 있다.   | 스토리에서 추출된 사실 \\
> > > > 고기를 요리하는 사람들은 종종 특정한 도구를 사용하여 고기를 손질한다.   | 상식 \\
> > > 예성은 주방에서 고기 요리를 자주 한다.  | 추론된 사실 \\
> > > > 예성은 고기 손질용 도끼를 소유하고 있다.  | 스토리에서 추출된 사실 \\
> > > > 예성은 정기적으로 육류를 요리한다.  | 스토리에서 추출된 사실 \\
> > > > 고기를 요리하는 사람들은 종종 특정한 도구를 사용하여 고기를 손질한다.  | 상식 \\
> > > 고기를 자주 요리하는 사람들은 적절한 도구를 사용하여 고기를 손질하려고 한다.  | 상식 \\
> > 예성이는 목수로서 다양한 도구를 소유하고 있다.  | 추론된 사실 \\
> > > 예성은 다양한 목공 도구를 가지고 있다.  | 추론된 사실 \\
> > > > 예성은 작업을 위해 정기적으로 도구를 점검하고 유지 보수한다.  | 스토리에서 추출된 사실 \\
> > > > 예성은 목수로 일하며 자주 작업 공간을 정돈한다.  | 스토리에서 추출된 사실 \\
> > > > 목수들은 일반적으로 작업을 원활하게 하기 위해 다양한 도구를 갖추고 있다.  | 상식 \\
> > > 예성은 목공 작업을 주기적으로 한다.  | 추론된 사실 \\
> > > > 예성은 목수로 일하며 자주 작업 공간을 정돈한다.  | 스토리에서 추출된 사실 \\
> > > > 예성은 작업을 위해 정기적으로 도구를 점검하고 유지 보수한다.  | 스토리에서 추출된 사실 \\
> > > > 목공 작업을 하는 사람들은 작업에 필요한 다양한 도구를 소유하고 있는 경향이 있다.  | 상식 \\
> > > 목공 작업을 하는 사람들은 작업에 필요한 다양한 도구를 소유하고 있는 경향이 있다.  | 상식 \\
> > 목수라면 다양한 도구를 쉽게 구할 수 있으며, 고기 손질용 도끼를 소유하고 있다면 이를 범행에 사용할 수 있다.  | 상식 \\
> 예성이는 범행 기회가 있다. | 추론된 사실 \\
> > 영은과 예성은 같은 동네에 거주한다.  | 추론된 사실 \\
> > > 예성은 동네 공원에서 영은과 함께 산책을 즐겼다.  | 추론된 사실 \\
> > > > 예성은 영은이 참여하는 동네 주민 모임에 자주 참석했다.   | 스토리에서 추출된 사실 \\
> > > > 예성은 동네 공원에서 영은과 함께 산책을 즐겼다.   | 스토리에서 추출된 사실 \\
> > > > 동일한 사회 활동과 장소에서 자주 만나는 사람들은 같은 동네에 거주할 가능성이 높다.   | 상식 \\
> > > 영은은 동네의 주민 모임에서 예성을 자주 만났다.  | 추론된 사실 \\
> > > > 예성은 동네 공원에서 영은과 함께 산책을 즐겼다.  | 스토리에서 추출된 사실 \\
> > > > 예성은 영은이 참여하는 동네 주민 모임에 자주 참석했다.  | 스토리에서 추출된 사실 \\
> > > > 동일한 사회 활동과 장소에서 자주 만나는 사람들은 같은 동네에 거주할 가능성이 높다.  | 상식 \\
> > > 동일한 사회 활동과 장소에서 자주 만나는 사람들은 같은 동네에 거주할 가능성이 높다.  | 상식 \\
> > 예성은 최근에 자주 영은의 차를 수리해주었다.  | 추론된 사실 \\
> > > 예성은 기계에 익숙한 목수로서 차량 수리에 능숙하다.  | 추론된 사실 \\
> > > 예성은 영은이 차량 문제로 자주 도움을 요청했다고 말했다.  | 추론된 사실 \\
> > > 이웃이 차량 문제로 도움을 요청하여 자주 수리를 해주면, 그 차량에 접근할 기회를 가지게 된다.  | 상식 \\
> > 같은 동네에 살면서 자주 차를 수리해줬다면, 자연스럽게 차에 접근할 기회를 가질 수 있다.  | 상식 \\
> 예성이는 범행 동기가 있다. | 추론된 사실 \\

완성해야 할 함의 추론:\\
예성\\
> 예성이는 범행 기회가 있다. 왜냐하면, \\
> > 예성은 최근에 자주 영은의 차를 수리해주었다.  왜냐하면, \\
> > > 예성은 기계에 익숙한 목수로서 차량 수리에 능숙하다.  왜냐하면, \\
> > > > 스토리에서 추출된 사실\\
> > > > 스토리에서 추출된 사실\\
> > > > 상식\\

'완성해야 할 함의 추론'은 '현재 트리'의 일부입니다. '완성해야 할 함의 추론'에 대해서만, 정확한 개수의 '스토리에서 추출된 사실'과 '상식'를 생성하세요.\\
\end{hintbox}

%% file: sections/tables/synthesis-prompts/komm-expansion-translated.tex
\begin{hintbox}{Prompts for Narrative Generation - Murder Mysteries (Translated)}{fig:prompt-mm-expand-en}

Your task is to generate a reasoning tree for a story following the example provided. \\
In this tree, each statement must be logically derived from its child nodes. \\
If a statement already exists (e.g., a root node), do not overwrite it.\\

Story type:\\

We are creating a murder mystery. A murder mystery requires a complex web of evidence entangled with a suspect’s \textit{means}, \textit{motive}, and \textit{opportunity}. \\
The story must be written from the detective’s point of view. Evidence collection should follow ordinary investigative methods (e.g., interrogations, overheard conversations, criminal record searches, mail or trash examination, etc.).\\

1. Every statement in the tree must be logically derived from its child nodes.\\
2. All “Fact from Story” and “Commonsense Knowledge” nodes must be directly related to the final inferred reasoning.\\
3. Each root statement must specify its source (“Fact from Story” or “Commonsense Knowledge”).\\
4. “Fact from Story” must be a concrete statement about characters, locations, or objects in the narrative.\\
5. “Commonsense Knowledge” must represent facts most people would agree with and must not directly reference specific people or events in the story.\\
6. “Commonsense Knowledge” acts as a rule combining sibling nodes to derive higher-level conclusions.\\
7. The generated tree must match the structure of the “Reasoning to be Completed.”\\
8. For each “Fact from Story” and “Commonsense Knowledge” requested in the “Reasoning to be Completed,” generate exactly one node of each type.\\
9. Do not generate any additional “Fact from Story” or “Commonsense Knowledge” nodes not requested in the “Reasoning to be Completed.”\\

Having a \textit{means} implies access to the murder weapon.\\
Having a \textit{motive} implies a reason to kill the victim.\\
Having an \textit{opportunity} implies being present at the crime scene.\\

Even if a parent statement appears unrealistic or implausible, you must preserve it and construct the reasoning tree accordingly.\\

See the examples below.\\

Scenario:\\
Victim: Yuri\\
Crime Scene: House\\
Murder Weapon: Gun\\
Suspect: Minsu\\
Suspect’s Role: Brother\\
Motive: Financial gain\\

Current tree: \\
Minsu is the murderer. | Deduced Root Conclusion \\
> Minsu has a means. | Deduced Fact \\
> > Minsu practiced shooting. | Fact from Story \\
> > Minsu owns a gun. | Fact from Story \\
> > If someone owns a gun and has practiced using it, they have the ability to commit murder. | Commonsense Knowledge \\
> Minsu has a motive. | Deduced Fact \\
> > Minsu was extremely desperate for money. | Fact from Story \\
> > Minsu was desperate for Yuri’s money. | Fact from Story \\
> > When someone is desperate, they may resort to extreme actions—including murder—to achieve their goal. | Commonsense Knowledge \\
> Minsu has an opportunity. | Fact from Story \\

Reasoning to be completed:\\
Minsu has an opportunity.\\
> Fact from Story\\
> Commonsense Knowledge\\

Output:\\
Minsu has an opportunity.\\
> Minsu had access to Yuri’s house. | Fact from Story\\
> Having access to someone’s house provides an opportunity to commit murder. | Commonsense Knowledge\\

Another example:\\
Scenario:\\
Victim: Chulsoo\\
Crime Scene: Racetrack\\
Murder Weapon: Shovel\\
Suspect: Jimin\\
Suspect’s Role: Running partner\\
Motive: To prevent harm to another person\\

Current tree:\\
Jimin is the murderer. | Deduced Root Conclusion \\
> Jimin has a means. | Deduced Fact \\
> > Jimin is a farmer. | Fact from Story \\
> > Farmers typically use gardening tools such as shovels. | Commonsense Knowledge \\
> Jimin has a motive. | Fact from Story \\
> Jimin has an opportunity. | Fact from Story \\

Reasoning to be completed:\\
Jimin has a motive.\\
> Fact from Story\\
> Fact from Story\\
> Commonsense Knowledge\\

Output:\\
Jimin has a motive.\\
> Jimin deeply loves Minsu. | Fact from Story\\
> Chulsoo threatened Minsu. | Fact from Story\\
> Deep, passionate love can drive a person to extreme actions—even murder—when a loved one is threatened. | Commonsense Knowledge\\

Another example:\\
Scenario:\\
Victim: Junho\\
Crime Scene: Park bench\\
Murder Weapon: Heroin overdose\\
Suspect: Minchul\\
Suspect’s Role: Drug user\\
Motive: Public humiliation\\

Current tree:\\
Minchul is the murderer. | Deduced Root Conclusion\\
> Minchul has a means. | Fact from Story\\
> Minchul has a motive. | Fact from Story\\
> Minchul has an opportunity. | Fact from Story\\

Reasoning to be completed:\\
Minchul has a means.\\
> Fact from Story\\
> Fact from Story\\
> Commonsense Knowledge\\

Output:\\
Minchul has a means.\\
> Minchul had access to heroin. | Fact from Story\\
> Minchul knew the lethal dosage for overdose. | Fact from Story\\
> If someone has access to heroin and knows the lethal dosage, they can intentionally administer a fatal dose to commit murder. | Commonsense Knowledge\\

Now it’s your turn.\\

Scenario:\\
Victim: Yeong-eun\\
Crime Scene: Car\\
Murder Weapon: Meat cleaver\\
Suspect: Yesung\\
Suspect’s Role: Carpenter\\
Motive: Pathological obsession\\

Current tree: \\
Yesung | Deduced Root Conclusion \\
> Yesung has a means. | Deduced Fact \\
> > Yesung often uses a meat cleaver when preparing meat. | Deduced Fact \\
> > > Yesung uses a cleaver for meat preparation. | Deduced Fact \\
> > > > Yesung regularly cooks meat. | Fact from Story \\
> > > > Yesung owns a meat cleaver. | Fact from Story \\
> > > > People who cook meat often use specific tools for cutting it. | Commonsense Knowledge \\
> > > Yesung frequently cooks meat in the kitchen. | Deduced Fact \\
> > > > Yesung owns a meat cleaver. | Fact from Story \\
> > > > Yesung regularly cooks meat. | Fact from Story \\
> > > > People who cook meat often use specific tools for preparation. | Commonsense Knowledge \\
> > > People who frequently cook meat tend to use appropriate tools for cutting it. | Commonsense Knowledge \\
> > Yesung, as a carpenter, owns various tools. | Deduced Fact \\
> > > Yesung possesses various woodworking tools. | Deduced Fact \\
> > > > Yesung regularly inspects and maintains his tools for work. | Fact from Story \\
> > > > Yesung, who works as a carpenter, often organizes his workspace. | Fact from Story \\
> > > > Carpenters generally possess various tools to perform their work efficiently. | Commonsense Knowledge \\
> > > Yesung regularly engages in woodworking. | Deduced Fact \\
> > > > Yesung, who works as a carpenter, often organizes his workspace. | Fact from Story \\
> > > > Yesung regularly inspects and maintains his tools for work. | Fact from Story \\
> > > > People who do woodworking typically own a variety of necessary tools. | Commonsense Knowledge \\
> > > People who do woodworking generally possess the necessary tools for their craft. | Commonsense Knowledge \\
> > Being a carpenter allows easy access to various tools, and owning a cleaver means he could use it in the crime. | Commonsense Knowledge \\
> Yesung has an opportunity. | Deduced Fact \\
> > Yeong-eun and Yesung live in the same neighborhood. | Deduced Fact \\
> > > Yesung enjoyed walking with Yeong-eun in the neighborhood park. | Deduced Fact \\
> > > > Yesung frequently attended community meetings that Yeong-eun also joined. | Fact from Story \\
> > > > Yesung enjoyed walking with Yeong-eun in the neighborhood park. | Fact from Story \\
> > > > People who frequently meet in the same social activities and places are likely to live in the same neighborhood. | Commonsense Knowledge \\
> > > Yeong-eun often met Yesung at local gatherings. | Deduced Fact \\
> > > > Yesung enjoyed walking with Yeong-eun in the neighborhood park. | Fact from Story \\
> > > > Yesung frequently attended community meetings that Yeong-eun also joined. | Fact from Story \\
> > > > People who frequently meet through social activities and shared spaces tend to live nearby. | Commonsense Knowledge \\
> > > People who meet often in similar places and contexts usually live close to each other. | Commonsense Knowledge \\
> > Yesung recently repaired Yeong-eun’s car multiple times. | Deduced Fact \\
> > > As a mechanically skilled carpenter, Yesung is good at vehicle repair. | Deduced Fact \\
> > > Yeong-eun frequently asked for Yesung’s help with car issues. | Deduced Fact \\
> > > When a neighbor frequently helps repair someone’s car, they naturally gain access to it. | Commonsense Knowledge \\
> > Living nearby and frequently repairing the car naturally gives access to it. | Commonsense Knowledge \\
> Yesung has a motive. | Deduced Fact \\

Reasoning to be completed:\\
Yesung\\
> Yesung has an opportunity because\\
> > Yesung recently repaired Yeong-eun’s car multiple times, because\\
> > > Yesung is a mechanically skilled carpenter proficient at vehicle repair, because\\
> > > > Fact from Story\\
> > > > Fact from Story\\
> > > > Commonsense Knowledge\\

The “Reasoning to be Completed” is part of the “Current Tree.” For this section only, generate the exact number of “Fact from Story” and “Commonsense Knowledge” nodes requested.\\
\end{hintbox}

%% file: sections/tables/synthesis-prompts/komm-generation.tex
\begin{hintbox}{Prompts for Narrative Generation - Murder Mysteries}{fig:prompt-mm-gen-ko}

우리는 살인 미스터리를 만들고 있습니다. \\
살인 미스터리에는 용의자와 관련된 복잡한 증거들이 필요합니다. \\
이야기는 형사의 시점에서 쓰여야 합니다. \\
형사는 통상적인 수사 방식을 통해 증거를 수집합니다(심문, 대화 엿듣기, 범죄 기록 조사, 우편물·쓰레기 뒤지기 등).\\

당신에게 사실 목록을 제공할 것이고, 당신은 이 목록에 있는 사실들을 모두 이야기에 포함시켜야 합니다. \\
절대 유도된 사실(추론된 사실)이나 결론을 언급해서는 안 됩니다. 이야기의 흐름은 이 사실들을 충실히 따라가야 합니다.\\

다음은 살인 미스터리에서 용의자와 관련된 한 장(chapter)을 작성해야 합니다. \\
이 장에서는 살인 사건이나 피해자에 관한 직접적인 언급을 최소화하고, \\
형사가 용의자에게 접근하여 수사를 진행하는 과정을 그립니다. \\
아래 규칙을 유의하세요:\\

1. 장의 내용만 작성하고, 제목이나 번호를 달지 마세요. 다른 장과 연속적으로 이어질 수 있도록 간단한 서술만 하세요.\\
2. 용의자가 '범행 수단을 가졌다', '범행 동기를 가졌다', '범행 기회가 있다'라고 절대 말하지 마세요.\\
3. 용의자에게 범행 수단, 범행 동기, 범행 기회가 있다고 암시조차 하지 마세요.\\
4. 유도된 사실(추론된 사실)을 직접적으로 언급하거나 결론짓지 마세요.\\
5. 용의자를 살인자로 단정 지어서 말하지 마세요. 그건 독자가 추리해야 할 부분입니다.\\
6. 이야기의 시점은 형사가 현장 조사를 하거나 용의자를 심문하는 등, 일반적으로 합리적인 수사 방식을 통해 단서를 얻는 상황으로 유지하세요.\\
7. 이야기 속에서 '이 사실은 의심스러워' 같은 멘트를 하지 마세요. 독자가 추론하게 남겨두세요.\\
8. 이야기 속에서 형사의 이름은 '김철수'로 통일하세요.\\

주어진 사실(fact) 목록에서 각 사실을 이야기 속에 반드시 등장시키고, \\
최대 10개의 대사를 포함해 작성하세요.\\

다음은 예시입니다:\\

용의자와 범죄 정보\\
피해자: 유진\\
범행 장소: 인적이 드문 숲\\
살해 도구: 칼\\
용의자: 지훈\\
용의자의 역할: 학교 운동장 관리인\\
범행 동기: 종교적 희생\\

당신은 김철수 형사입니다. \\
반드시 포함해야 할 사실들: \\
- 어떤 목격자는 스파게티 얼굴과 초록색 귀를 가진 누군가를 보았다 \\
- 지훈이는 지역 학교의 잔디 관리인이다 \\
- 지훈이는 근처 주민들에게 페인트칠, 잔디 손질 등 서비스를 제공한다 \\
- 지훈이는 주변 집 중 한 곳을 초록색으로 페인트칠했다 \\
- 지훈이는 어렸을 때 큰 화상을 입었다 \\
- 지훈이의 가족은 여러 세대 동안 이 지역에 살았다 \\
- 이 지역에서는 예전부터 종교적 극단주의자들이 있었고, 모두 신비로운 종교 의식을 치렀다 \\
- 지훈이와 그의 직계 가족들은 모두 잡다한 기술을 가지고 있었다 \\
- 지훈이는 조상과 전통을 존중하는 신념을 갖고 있다 \\
- 유진이는 근처에 있는 새로운 교회에 참여할 생각을 적어두었다 \\
- 유진이의 친구는 유진이가 새로 사귀게 된 불량스러운 무리와 어울리는 것을 걱정했다 \\

출력 예시: \\

김철수 형사는 범행 현장 사진을 훑어보면서 담배를 길게 한 모금 빨았다. 그는 끔찍한 장면들에 어느 정도 익숙해져 있었지만, 유진이의 사건에는 이상하게 마음이 쏠렸다... \\

(중략: 예시 이야기) \\

--- \\

이제 당신 차례입니다. 이야기 외에는 아무것도 출력하지 마세요. \\

용의자와 범죄 정보 \\
피해자: 영은 \\
범행 장소: 자동차 \\
살해 도구: 고기 손질용 도끼(미트클리버) \\
용의자: 예성 \\
용의자의 역할: 목수 \\
범행 동기: 병적인 집착 \\

당신은 김철수 형사입니다. \\

반드시 포함해야 할 사실들: \\
- 예성은 영은이 차량 문제로 자주 도움을 요청했다고 말했다.  \\
- 예성은 영은이 새로운 스포츠 클럽에 가입한 후 다른 사람들과 함께 시간을 보내는 것을 좋아하지 않았다.  \\
- 예성은 영은이 가는 모든 곳을 따라 다녔다.  \\
- 예성은 영은이 참여하는 동네 주민 모임에 자주 참석했다.  \\
- 예성은 동네 공원에서 영은과 함께 산책을 즐겼다.  \\
- 예성은 매일 저녁마다 영은과 함께 시간을 보내려 했다.  \\
- 예성은 영은의 사회적 미디어 활동을 지속적으로 체크했다.  \\
- 예성은 목수로 일하며 자주 작업 공간을 정돈한다.  \\
- 영은이 다른 약속이 있을 때마다 예성은 심한 불안을 느꼈다.  \\
- 예성은 이웃들에게 친절한 목수로 알려져 있다.   \\
- 예성은 목수로서 손재주가 뛰어나다.  \\
- 예성은 정기적으로 육류를 요리한다.  \\
- 영은은 새로운 스포츠 클럽에 가입했다.  \\
- 예성은 영은과의 개인적인 시간을 중요시했고, 그녀가 점점 더 많은 사회적 활동에 빠져드는 것을 불편해했다.  \\
- 예성은 정기적으로 육류를 요리한다.   \\
- 그 스포츠 클럽은 많은 사회적인 활동과 모임을 제공한다.  \\
- 영은의 차량은 최근 잦은 고장을 겪었다.   \\
- 예성은 영은이 참여하는 동네 주민 모임에 자주 참석했다.   \\
- 예성은 동네 공원에서 영은과 함께 산책을 즐겼다.   \\
- 예성은 고기 손질용 도끼를 소유하고 있다.   \\
- 예성은 고기 손질용 도끼를 소유하고 있다.  \\
- 예성은 작업을 위해 정기적으로 도구를 점검하고 유지 보수한다.   \\
\end{hintbox}

%% file: sections/tables/synthesis-prompts/komm-generation-translated.tex
\begin{hintbox}{Prompts for Narrative Generation - Murder Mysteries (Translated)}{fig:prompt-mm-gen-en}

We are creating a murder mystery. \\
A murder mystery requires a web of complex evidence related to the suspects. \\
The story must be written from the perspective of a detective. \\
The detective gathers evidence through standard investigative methods (interrogation, eavesdropping, checking criminal records, examining mail or trash, etc.).\\

You will be provided with a list of facts, and you must include all of these facts in the story. \\
You must not mention any inferred (deduced) facts or conclusions. The flow of the story must faithfully follow only the given facts.\\

You are to write a chapter in the murder mystery related to the suspect. \\
This chapter should minimize direct references to the murder or the victim, \\
and instead depict the detective approaching and investigating the suspect. \\
Please follow the rules below:\\

1. Write only the chapter content—do not add titles or numbers. Keep it simple so that it can connect naturally with other chapters.\\
2. Never say that the suspect “has a means,” “has a motive,” or “has an opportunity” to commit the crime.\\
3. Do not even imply that the suspect has a means, motive, or opportunity.\\
4. Do not explicitly mention or conclude any inferred (deduced) facts.\\
5. Do not state or imply that the suspect is the murderer. That is for the reader to infer.\\
6. The story’s perspective should stay consistent with reasonable investigative methods—detective Kim Cheolsu examining the scene, interrogating the suspect, etc.\\
7. Do not include remarks like “This is suspicious.” Leave that for the reader to deduce.\\
8. The detective’s name must always be “Kim Cheolsu.”\\

Each fact from the given list must appear in the story, \\
and you may include up to 10 dialogue lines.\\

Example:\\

Suspect and Crime Information\\
Victim: Yujin\\
Crime Scene: Secluded forest\\
Murder Weapon: Knife\\
Suspect: Jihoon\\
Suspect’s Role: School field manager\\
Motive: Religious sacrifice\\

You are Detective Kim Cheolsu. \\
Facts that must be included: \\
- A witness saw someone with a spaghetti face and green ears.\\
- Jihoon is a grass caretaker at the local school.\\
- Jihoon provides services like painting and lawn care to nearby residents.\\
- Jihoon painted one of the neighborhood houses green.\\
- Jihoon suffered severe burns as a child.\\
- Jihoon’s family has lived in this area for generations.\\
- The region has a long history of religious extremists who practiced mysterious rituals.\\
- Jihoon and his direct family members are all skilled in various crafts.\\
- Jihoon holds strong beliefs about respecting his ancestors and traditions.\\
- Yujin had written about joining a new church nearby.\\
- Yujin’s friend was worried about her hanging out with a new, troublesome crowd.\\

Example Output:\\

Detective Kim Cheolsu took a long drag on his cigarette as he looked over the photos from the crime scene. He had seen his share of gruesome cases, but something about Yujin’s case drew his attention...\\

(omitted: example narrative)\\

---\\

Now it’s your turn. Output only the story—nothing else.\\

Suspect and Crime Information\\
Victim: Yeongeun\\
Crime Scene: Car\\
Murder Weapon: Meat cleaver\\
Suspect: Yeseong\\
Suspect’s Role: Carpenter\\
Motive: Pathological obsession\\

You are Detective Kim Cheolsu.\\

Facts that must be included: \\
- Yeseong said that Yeongeun often asked him for help with car troubles.\\
- Yeseong disliked that Yeongeun enjoyed spending time with others after joining a new sports club.\\
- Yeseong followed Yeongeun everywhere she went.\\
- Yeseong frequently attended the neighborhood meetings that Yeongeun joined.\\
- Yeseong enjoyed walking with Yeongeun in the local park.\\
- Yeseong tried to spend every evening with Yeongeun.\\
- Yeseong constantly checked Yeongeun’s social media activities.\\
- Yeseong works as a carpenter and often organizes his workspace.\\
- Whenever Yeongeun had other plans, Yeseong felt intense anxiety.\\
- Yeseong is known in the neighborhood as a kind carpenter.\\
- Yeseong is highly skilled at his craft as a carpenter.\\
- Yeseong regularly cooks meat.\\
- Yeongeun joined a new sports club.\\
- Yeseong valued his personal time with Yeongeun and felt uneasy about her increasing social activities.\\
- Yeseong regularly cooks meat.\\
- The sports club hosts many social gatherings and events.\\
- Yeongeun’s car has recently been breaking down frequently.\\
- Yeseong often attended the same neighborhood meetings as Yeongeun.\\
- Yeseong enjoyed walking with Yeongeun in the local park.\\
- Yeseong owns a meat cleaver.\\
- Yeseong owns a meat cleaver.\\
- Yeseong routinely inspects and maintains his tools for work.\\
\end{hintbox}

%% file: sections/tables/synthesis-prompts/koop-expansion.tex
\begin{hintbox}{Prompts for Reasoning Tree Expansion - Object Placements}{fig:prompt-op-expand-ko}
우리는 사람들이 많은 아이템들을 여러 번 옮기는 이야기를 만들고 있습니다. 이 이야기의 목표는 흥미롭고 독창적이면서도, 물건이 어디에 있는지를 명확하게 추적할 수 있도록 하여, 나중에 독자나 언어 모델에게 세계 상태나 누가 무엇을 알고 있는지에 대해 질문할 수 있도록 하는 것입니다.\\

이 이야기를 만들기 위해, 우리는 이야기 전개와 세계 상태의 변화를 개요화한 트리 구조를 만들었습니다. 당신의 역할은 누군가가 어떤 사건을 보았는지 또는 보지 못했는지를 증명하는 포함 추론 트리를 완성하는 것입니다.\\

논리 트리는 중간 노드들이 자식 노드들에 의해 포함되는 트리 구조입니다. 이들은 사실들의 집합에 대해 자연어로 된 추론 증명을 만듭니다.\\

이 트리를 완성하기 위해서는 하나의 함의 추론을 완성해야 합니다. 함의 추론을 완성하는 것은 논리 추론 트리의 한 서브트리를 채우는 것과 유사합니다. 이 단계를 채우기 위해서는 단계의 구조를 따라야 합니다.\\

"스토리에서 추출된 사실"는 우리가 이야기를 쓸 때 명시적으로 서술될 사실들입니다.\\
"상식"는 대부분의 사람들이 사실로 받아들이는, 명시적으로 말할 필요가 없는 사실들입니다.\\
"복합 사실"는 이야기에서의 더 단순한 사실들에 의해 함의 사실들입니다 (이들은 나중에 재귀 호출을 통해 채워질 것입니다!)\\

이 단계의 모든 사실은 결합되어 루트 상위 사실을 포함해야 합니다.\\

현재의 구조 트리와 모순되는 사실은 포함할 수 없습니다.\\

다른 사람들에 대한 사실은 포함하지 마세요. 무언가가 움직이는 것을 보고 있는지 여부에 대한 그 사람에 대한 사실에만 집중하세요.\\

'완성해야 할 함의 추론'과 정확히 일치하게 작성하세요. 스토리에서 추출된 사실와 상식의 개수도 동일하게 작성하고, 순서도 그대로 따라야 합니다.\\

누군가가 어떤 것을 보았거나 보지 못했다고 명시적으로 말하지 마세요. 당신의 역할은 이를 암시하는 사실들을 제공하는 것입니다. 예를 들어, 지수가 철수가 무언가를 옮기는 것을 보았다고 암시하고 싶다면, "지수는 집안일을 하면서 철수를 보고 있었다" 그리고 "누군가를 보고 있다면, 그 사람이 무엇을 하고 있는지를 보는 것이다"라고 말할 수 있습니다. 누군가가 본 사실을 직접적으로 말하지 않고, 보는 행위의 물리적 조건을 설명하는 방식입니다.\\

아이템이 움직였다는 사실을 언급하거나 이미 등장한 아이템을 재사용하지 마세요. 예를 들어, "철수가 자신의 아이폰을 보지 못했다"라는 사실이 있고, 당신이 "유림이는 사과가 움직이는 것을 보지 못했다"를 증명하고 있다면, 철수의 아이폰을 다시 사용하지 마세요. 우리 프로그램은 아이템이 어디에 있는지를 엄격하게 제어하고 있으며, 우리가 고려하지 않은 아이템 배치를 도입하는 것을 원하지 않습니다.\\

각 사실은 추론을 위해 결정적이어야 합니다. 일부러 세부사항을 생략하여 다른 사실들이 이를 보완하도록 하세요. 하나의 사실이 빠지면 결론이 도출되지 않아야 합니다. 같은 사실을 반복해서 사용하지 마세요.\\

"그"나 "그녀" 같은 대명사 대신 항상 사람의 이름을 사용하세요. 이름을 알고 있다면 반드시 이름을 사용하세요.\\

한 번에 하나의 추론만 수행하세요. 당신의 추론은 반드시 "완성해야 할 함의 추론" 템플릿과 정확히 일치해야 하며, 우리가 나중에 이를 파싱할 수 있어야 합니다.\\

예제를 확인하세요.\\

시나리오:\\
민수와 예림이는 노래방에 있다.\\

현재 트리:\\
민수와 예림이는 노래방에 있다. | 추론된 결론\\
> 오프닝 장면 | 추론된 사실\\
> > 민수는 무대에 있는 마이크를 보았다. | 스토리에서 추출된 사실\\
> > 예림이는 무대에 있는 마이크를 보았다. | 스토리에서 추출된 사실\\
> > 민수는 바에 있는 맥주를 보았다. | 스토리에서 추출된 사실\\
> > 예림이는 바에 있는 맥주를 보았다. | 스토리에서 추출된 사실\\
> 민수는 맥주를 테이블로 옮긴다. | 추론된 사실\\
> > 에림이는 맥주가 테이블로 옮겨지는 것을 보지 못했다. | 추론된 사실\\
> > > 예림이는 테이블에서 반대쪽을 보고 있었다. | 추론된 사실\\
> > > > 예림이는 다른 손님과 대화 중이었다. | 스토리에서 추출된 사실\\
> > > > 다른 손님은 테이블 쪽을 보고 있었다. | 스토리에서 추출된 사실\\
> > > > 보통 사람들은 서로를 바라보며 대화하기 때문에, 한 사람이 한 방향을 보고 있으면 다른 사람은 반대 방향을 보고 있게 된다. | 상식\\
> > > 만약 누군가 다른 무언가에서 반대 방향을 보고 있다면, 그 무언가 주변에서 일어나는 일을 볼 수 없다. | 상식\\
> 예림이는 마이크를 테이블로 옮긴다 | 추론된 사실\\
> > 예림이는 마이크를 옮길 때 테이블에 있는 맥주를 보았다. | 스토리에서 추출된 사실\\
> > 민수는 마이크가 테이블로 옮겨지는 것을 보았다. | 추론된 사실\\
> > > 민수는 테이블에서 맥주를 마시고 있었다. | 상식\\
> > > 어떤 일이 사람 바로 옆에서 일어나면, 그 사람은 보통 그 상황을 볼 수 있다. | 스토리에서 추출된 사실\\
> 예림이는 맥주를 쓰레기통으로 옮긴다. | 추론된 사실\\
> > 민수는 맥주가 쓰레기통으로 옮겨지는 것을 보지 못했다. | 스토리에서 추출된 사실\\

완성해야 할 함의 추론:\\
민수와 예림이는 노래방에 있다.\\
> 예림이는 맥주를 쓰레기통으로 옮긴다. 왜냐하면, 이 야야기에 따르면,\\
> > 민수는 맥주가 쓰레기통으로 옮겨지는 것을 보지 못했다. 왜냐하면, 이 야야기에 따르면,\\
> > > 스토리에서 추출된 사실\\
> > > 스토리에서 추출된 사실\\
> > > 상식\\

출력:\\
민수와 예림이는 노래방에 있다.\\
> 예림이는 맥주를 쓰레기통으로 옮긴다. 왜냐하면, 이 야야기에 따르면,\\
> > 민수는 맥주가 쓰레기통으로 옮겨지는 것을 보지 못했다. 왜냐하면, 이 야야기에 따르면,\\
> > > 예림이는 민수를 속여서 "저쪽"을 보게 했다. | 스토리에서 추출된 사실\\
> > > 예림이는 민수에게 쓰레기통의 반대 방향을 가리켰다. | 스토리에서 추출된 사실\\
> > > 누군가를 속여서 다른 곳을 보게 만들면, 반대쪽에서 일어나는 일을 볼 수 없다. | 상식\\

이제 당신의 차례입니다.\\

시나리오:\\

이야기 개요:  \\
영수는 다음 주 제출해야 할 중요한 프로젝트를 진행 중입니다. 실험 결과와 필요한 모든 정보를 USB에 저장해 두었으며, 이를 잃어버리면 큰 문제가 발생할 것입니다. 지민은 실험실의 질서를 유지하면서 영수가 프로젝트에 집중할 수 있도록 돕습니다. 동시에 은혁은 데이터의 안전과 보안을 보장합니다.\\

인물 1:\\
이름: 영수\\
역할: 연구원\\
동기: 영수는 중요한 프로젝트를 제시간에 제출하기 위해 모든 자료를 USB에 저장하고 안전하게 보관하고자 합니다.\\

인물 2:\\
이름: 지민\\
역할: 연구 실험 보조원\\
동기: 지민은 영수가 효율적으로 프로젝트를 수행하도록 돕고, 실험실을 깔끔하게 유지하는 것에 자부심을 느낍니다.\\

인물 3:\\
이름: 은혁\\
역할: IT 기술 지원\\
동기: 은혁은 연구소의 모든 데이터 저장 장치가 안전하게 보관되도록 책임을 집니다.\\

추론을 만들 때에는 인물들의 역할을 존중하고, 그들이 이야기에 따라 \\
어떤 사건을 보거나 보지 못하도록 하는 동기를 부여해 주세요. 또한 이러한 동기나 이야기는 여러 "이동"에 걸쳐 일관되게 이어지도록 노력해 주세요.\\
누군가가 어떤 것을 보았거나 보지 못했다고 명시적으로 말하지 마세요. 당신의 역할은 이를 암시하는 사실들을 제공하는 것입니다. 예를 들어, 지수가 철수가 무언가를 옮기는 것을 보았다고 암시하고 싶다면, "지수는 집안일을 하면서 철수를 보고 있었다" 그리고 "누군가를 보고 있다면, 그 사람이 무엇을 하고 있는지를 보는 것이다"라고 말할 수 있습니다. 누군가가 본 사실을 직접적으로 말하지 않고, 보는 행위의 물리적 조건을 설명하는 방식입니다.\\
물건이 움직였다는 사실을 언급하거나 이미 등장한 물건을 재사용하지 마세요. 예를 들어, "철수가 자신의 아이폰을 보지 못했다"라는 사실이 있고, 당신이 "유림이는 사과가 움직이는 것을 보지 못했다"를 증명하고 있다면, 철수의 아이폰을 다시 사용하지 마세요. 우리 프로그램은 아이템이 어디에 있는지를 엄격하게 제어하고 있으며, 우리가 고려하지 않은 아이템 배치를 도입하면 안 됩니다.\\

현재 트리:\\
영수는 중요한 프로젝트의 자료를 USB에 저장해 가지고 있습니다. USB를 잃어버리면 프로젝트 제출이 불가능해집니다. | 추론된 결론\\
> 이야기 시작 | 추론된 사실\\
> > 영수가 노트북을 실험실 책상에서 보았다. | 스토리에서 추출된 사실\\
> > 영수가 USB를 실험실 책상에서 보았다. | 스토리에서 추출된 사실\\
> > 은혁이 노트북을 실험실 책상에서 보았다. | 스토리에서 추출된 사실\\
> > 은혁이 USB를 실험실 책상에서 보았다. | 스토리에서 추출된 사실\\
> > 지민이 노트북을 실험실 책상에서 보았다. | 스토리에서 추출된 사실\\
> > 지민이 USB를 실험실 책상에서 보았다. | 스토리에서 추출된 사실\\
> 지민이 노트북을 창고 선반으로 옮긴다. 왜냐하면, 지민은 실험이 끝난 후 실험실의 공간을 확보하기 위해 노트북을 정리합니다. | 추론된 사실\\
> > 영수는 노트북을 창고 선반으로 이동하는 것을 보았다. | 추론된 사실\\
> > 은혁이는 노트북을 창고 선반으로 이동하는 것을 보았다. | 추론된 사실\\
> 영수가 USB를 작업대 서랍으로 옮긴다. 왜냐하면, 영수는 중요한 자료를 안전한 장소에 보관하기 위해 가까운 서랍에 넣어 두기로 결정합니다. | 추론된 사실\\
> > 은혁이는 USB를 작업대 서랍으로 이동하는 것을 보았다. | 추론된 사실\\
> > 지민이는 USB를 작업대 서랍으로 이동하는 것을 보지 않았다. | 추론된 사실\\
> 은혁이 USB를 보안 금고로 옮긴다. 왜냐하면, 은혁은 자료의 보안을 강화하기 위해 USB를 보안 금고에 안전하게 보관합니다. | 추론된 사실\\
> > 영수는 USB를 보안 금고로 이동하는 것을 보았다. | 추론된 사실\\
> > 지민이는 USB를 보안 금고로 이동하는 것을 보지 않았다. | 추론된 사실\\

완성해야 할 함의 추론:\\
영수는 중요한 프로젝트의 자료를 USB에 저장해 가지고 있습니다. USB를 잃어버리면 프로젝트 제출이 불가능해집니다.\\
> 지민이 노트북을 창고 선반으로 옮긴다. 왜냐하면, 지민은 실험이 끝난 후 실험실의 공간을 확보하기 위해 노트북을 정리합니다. 왜냐하면, 이 야야기에 따르면,\\
> > 영수는 노트북을 창고 선반으로 이동하는 것을 보았다. 왜냐하면, 이 야야기에 따르면,\\
> > > 스토리에서 추출된 사실\\
> > > 스토리에서 추출된 사실\\
> > > 상식\\

'완성해야 할 함의 추론'은 '현재 트리'의 일부입니다. '완성해야 할 함의 추론'에 대해서만, 정확한 개수의 '스토리에서 추출된 사실'과 '상식'를 생성하세요.\\
\end{hintbox}

%% file: sections/tables/synthesis-prompts/koop-expansion-translated.tex
\begin{hintbox}{Prompts for Reasoning Tree Expansion - Object Placements (Translated)}{fig:prompt-op-expand-en}
We are creating stories in which people move many items multiple times. The goal of these stories is to be engaging and coherent, while allowing readers or language models to track the location of each object clearly so that we can later query them about world states or beliefs.\\

To build such stories, we construct a tree structure outlining both the story progression and changes in the world state. Your task is to complete an inclusion reasoning tree that explains why a person did or did not observe a particular event.\\

Each reasoning tree is a hierarchical structure where intermediate nodes are logically entailed by their child nodes. They form a natural language proof over sets of facts.\\

To complete a reasoning step, you must fill in one implication reasoning subtree that connects back to the higher-level statement. Follow the given structure exactly.\\

“Fact from story” refers to explicitly stated facts that appear in the narrative.\\
“Commonsense” refers to generally accepted truths that do not need to be explicitly stated.\\
“Composite fact” refers to facts implied by simpler facts within the story (these will be recursively expanded later).\\

All facts in this reasoning step must jointly entail the higher-level fact.\\

Do not include facts that contradict the existing tree.\\

Do not include facts about other people except when necessary to explain whether someone could or could not see an event. Focus only on physical or observational conditions.\\

Write your reasoning to exactly match the “Implication Reasoning to Complete” section. The number and order of “Fact from story” and “Commonsense” entries must be identical to the template.\\

Do not state directly that someone saw or did not see something. Instead, provide supporting facts that imply this. For example, to imply that Jisoo saw Chulsoo moving something, you might write “Jisoo was watching Chulsoo while doing household chores” and “When someone is watching another person, they usually see what that person is doing.” Avoid direct observation statements; instead describe physical conditions that make observation possible.\\

Do not mention that items moved, nor reuse already introduced items. For example, if the fact “Chulsoo did not see his iPhone” is present and you are reasoning about “Yurim did not see the apple move,” do not reuse Chulsoo’s iPhone. The system tracks all object states strictly, and new untracked items should not be introduced.\\

Each fact should be decisive for inference. Omit some details intentionally so that other facts complete the reasoning chain. The conclusion must fail if any fact is missing. Avoid repeating the same fact.\\

Always use proper names instead of pronouns.\\

Perform exactly one reasoning step at a time. Your reasoning must strictly conform to the “Implication Reasoning to Complete” template so that it can be parsed programmatically.\\

See the example below.\\

Scenario:\\
Minsu and Yerim are at a karaoke bar.\\

Current tree:\\
Minsu and Yerim are at a karaoke bar. | Deduced Root Conclusion\\
> Opening scene | Deduced Fact\\
> > Minsu saw the microphone on the stage. | Fact from story\\
> > Yerim saw the microphone on the stage. | Fact from story\\
> > Minsu saw the beer at the bar. | Fact from story\\
> > Yerim saw the beer at the bar. | Fact from story\\
> Minsu moves the beer to the table. | Deduced Fact\\
> > Yerim did not see the beer being moved to the table. | Deduced Fact\\
> > > Yerim was facing away from the table. | Deduced Fact\\
> > > > Yerim was talking to another customer. | Fact from story\\
> > > > The other customer was facing the table. | Fact from story\\
> > > > People usually face each other when talking, so one looks opposite the other. | Commonsense\\
> > > If someone is facing away from something, they cannot see what happens around it. | Commonsense\\
> Yerim moves the microphone to the table. | Deduced Fact\\
> > Yerim saw the beer on the table while moving the microphone. | Fact from story\\
> > Minsu saw the microphone being moved to the table. | Deduced Fact\\
> > > Minsu was drinking beer at the table. | Fact from story\\
> > > When something happens right next to someone, they usually see it. | Commonsense\\
> Yerim moves the beer to the trash can. | Deduced Fact\\
> > Minsu did not see the beer being moved to the trash can. | Fact from story\\

Implication reasoning to complete:\\
Minsu and Yerim are at a karaoke bar.\\
> Yerim moves the beer to the trash can because, according to the story,\\
> > Minsu did not see the beer being moved to the trash can because, according to the story,\\
> > > Fact from story\\
> > > Fact from story\\
> > > Commonsense\\

Output:\\
Minsu and Yerim are at a karaoke bar.\\
> Yerim moves the beer to the trash can because, according to the story,\\
> > Minsu did not see the beer being moved to the trash can because, according to the story,\\
> > > Yerim tricked Minsu into looking “over there.” | Fact from story\\
> > > Yerim pointed Minsu toward the opposite direction of the trash can. | Fact from story\\
> > > When someone is deceived into looking elsewhere, they cannot see what happens behind them. | Commonsense\\

Now it’s your turn.\\

Scenario description:  \\
Youngsoo is working on an important project due next week. All experiment results and required data are stored on a USB drive, and losing it would be catastrophic. Jimin maintains order in the lab to help Youngsoo focus, while Eunhyuk ensures data safety and security.\\

When building your reasoning, respect each character’s role and motivation within the story. Ensure that these motivations remain consistent across multiple “moves.”\\
Do not directly state that someone saw or did not see something—use physical or situational implications instead.\\
Do not mention new or reused objects beyond those already in the tree.\\

Current tree:\\
Youngsoo stores his important project data on a USB drive. Losing the USB would make submission impossible. | Deduced Root Conclusion\\
> Story begins | Deduced Fact\\
> > Youngsoo saw the laptop on the lab desk. | Fact from story\\
> > Youngsoo saw the USB on the lab desk. | Fact from story\\
> > Eunhyuk saw the laptop on the lab desk. | Fact from story\\
> > Eunhyuk saw the USB on the lab desk. | Fact from story\\
> > Jimin saw the laptop on the lab desk. | Fact from story\\
> > Jimin saw the USB on the lab desk. | Fact from story\\
> Jimin moves the laptop to the storage shelf because she wants to clear space after finishing the experiment. | Deduced Fact\\
> > Youngsoo saw the laptop being moved to the storage shelf. | Deduced Fact\\
> > Eunhyuk saw the laptop being moved to the storage shelf. | Deduced Fact\\
> Youngsoo moves the USB to the workbench drawer because he wants to store it safely nearby. | Deduced Fact\\
> > Eunhyuk saw the USB being moved to the workbench drawer. | Deduced Fact\\
> > Jimin did not see the USB being moved to the workbench drawer. | Deduced Fact\\
> Eunhyuk moves the USB to the secure safe because he wants to strengthen data protection. | Deduced Fact\\
> > Youngsoo saw the USB being moved to the secure safe. | Deduced Fact\\
> > Jimin did not see the USB being moved to the secure safe. | Deduced Fact\\

Implication reasoning to complete:\\
Youngsoo stores his important project data on a USB drive. Losing the USB would make submission impossible.\\
> Jimin moves the laptop to the storage shelf because she wants to clear space after finishing the experiment. Because, according to the story,\\
> > Youngsoo saw the laptop being moved to the storage shelf because, according to the story,\\
> > > Fact from story\\
> > > Fact from story\\
> > > Commonsense\\

The “Implication reasoning to complete” is a subset of the “Current tree.” Generate exactly the specified number of “Fact from story” and “Commonsense” statements for it.\\
\end{hintbox}

%% file: sections/tables/synthesis-prompts/koop-generation.tex
\begin{hintbox}{Prompts for Narrative Generation - Object Placements}{fig:prompt-op-gen-ko}
[[PROMPT]]:\\
우리가 작성한 이야기를 계속 이어서 작성해주세요. 다음에 일어날 이 사건에 대해 짧은 설명을 써 주세요. 오직 이동에 대해서만 작성하고, 다른 추가 정보는 포함하지 마세요. \\

절대 "누군가가 뭔가를 보지 못했다"거나, 누군가가 어디에 있는지 추론할 수 있는 능력이 있음을 시사하지 마세요. "몰래" 혹은 "알 수 없게"와 같은 표현도 하지 마세요!
아래는 예시입니다.\\

한두 문장만 작성하세요. 매우 짧은 설명이어야 합니다.\\

이야기 개요: 용준이는 자신이 받을 자격이 있다고 생각했던 직업을 상우가 부정행위를 통해 차지했다고 생각하여 화가 났습니다! 그래서 용준이는 상우의 소지품을 버리기 시작했습니다.\\

인물:\\
이름: 용준\\
역할: 최근 졸업했고 아파트를 공동으로 사용 중인 사람\\
동기: 부정행위로 룸메이트가 자신의 취업 자리를 빼앗았다고 생각해 매우 화가 나 있다.\\

이벤트:\\
- 용준이가 차 열쇠를 쓰레기통으로 옮긴다. 이유: 용준이는 상우에게 화가 나 그의 열쇠를 버리고 싶었다.\\
- 용준이가 차 열쇠를 옮기다가 쓰레기통에 있는 아이폰을 발견했다.\\

출력 예시: 분노에 찬 동작으로 열쇠가 깡통 쓰레기통에 부딪히며 소리를 냈다. 곧이어 예상치 못한 *퍽* 소리가 났고... 잠시 호기심이 화를 억누르자, 용준이는 쓰레기통 안에서 아이폰을 발견했다.\\

또 다른 예시:\\

이야기 개요: 민수는 방금 새 아파트로 이사했지만, 이전 세입자가 집을 엉망으로 만들어 놓고 떠났다! 집주인은 아무런 조치도 취하지 않아 민수가 직접 청소를 해야 한다.\\

인물:\\
이름: 민수\\
역할: 새롭지만 지저분한 아파트에 막 이사 온 사람\\
동기: 이전 세입자가 남긴 어질러진 상태를 혼자 치워야 하며, 관리 사무소는 전혀 도움이 되지 않는다.\\

이벤트:\\
- 민수가 국수를 팬트리로 옮긴다. \\

이유: 민수는 아파트를 깨끗하게 만들고 싶었고, 국수를 제자리에 두는 것이 마지막 단계였기 때문이다!\\

출력 예시: 민수는 기쁨이 가득한 표정으로 국수를 팬트리에 넣었다. 끝이 없을 것 같던 쓰레기와 잡동사니 치우기는 드디어 마무리되어, 아파트가 비로소 깔끔해졌다!\\

당신의 차례입니다. \\

이야기 개요: 이야기 개요:  \\
영수는 다음 주 제출해야 할 중요한 프로젝트를 진행 중입니다. 실험 결과와 필요한 모든 정보를 USB에 저장해 두었으며, 이를 잃어버리면 큰 문제가 발생할 것입니다. 지민은 실험실의 질서를 유지하면서 영수가 프로젝트에 집중할 수 있도록 돕습니다. 동시에 은혁은 데이터의 안전과 보안을 보장합니다.\\

인물:\\
이름: 지민\\
역할: 연구 실험 보조원\\
동기: 지민은 영수가 효율적으로 프로젝트를 수행하도록 돕고, 실험실을 깔끔하게 유지하는 것에 자부심을 느낍니다.\\

이벤트:\\
- 지민이 노트북을 창고 선반으로 옮긴다. 왜냐하면, 지민은 실험이 끝난 후 실험실의 공간을 확보하기 위해 노트북을 정리합니다.\\

출력:\\
영수는 실험실 책상 위에 놓인 노트북과 USB를 다시 한 번 확인하며 안도의 한숨을 내쉬었다. 주말이면 제출해야 할 중요한 프로젝트가 머릿속을 가득 채우고 있었지만, 모든 실험 결과와 자료가 든 USB가 안전하게 보관되어 있다는 사실에 조금이나마 마음이 놓인 것이다. 지민은 실험실 안을 부지런히 정리하며 영수가 프로젝트에 오롯이 집중할 수 있도록 돕고 있었다. 그는 노트북과 USB가 실험실 책상에 놓여 있다는 것을 정확히 알고 있었고, 깔끔한 환경에서 영수의 연구 효율을 높이고자 했다. 은혁 역시 연구실의 데이터 보안을 책임지는 사람답게 이 두 물건을 주시하고 있었다. 그는 USB 속 민감한 자료가 외부로 새어나가지 않도록 주의 깊게 점검하며, 노트북이 안전하다는 사실도 재차 확인했다. 사실 이곳에 있는 세 사람 모두 노트북과 USB가 실험실 책상 위에 놓여 있다는 것을 알고 있었으며, 그 어느 때보다 신중하게 프로젝트를 준비하고 보안을 지키려 했다.\\

[[MODEL RESPONSE]]:\\
지민은 노트북을 창고 선반으로 옮겼다.\\

[[PROMPT]]:\\
지금까지 작성된 이야기를 이어서, 아래에 제시된 관찰 사실들에 대해 작성하세요.\\
단, 사실들에 대해서만 쓰고 새로운 정보를 추가하지 마세요.\\
"누군가 보았다", "알아차리지 못했다"라는 표현을 사용하지 말고,\\
누군가가 무언가를 본 사실이 유일한 관찰 사실일 때만 "누군가가 X를 보았다"라고 쓰세요.\\

이야기의 분위기를 잡는 데 사용할 수 있는 추가 정보가 있을 수 있지만, \\
항상 관찰 사실을 이야기의 주요 가이드로 삼아야 합니다.\\

아래 이야기의 핵심 물건들은 절대 언급하지 마세요:\\
- USB\\
- 노트북\\

이 단락 이후에 다른 사건이 있을 예정이므로, 사실에만 충실하여 이 단락을 갑작스럽게 마무리하세요. 일반적인 진술은 하지 마세요. 마지막 문장은 제시된 사실에 관한 내용만 담고, 온전한 문장이어야 합니다.\\

이야기는 현재의 가장 최근 상황에서 벌어지고 있다고 가정하세요.\\
다시 말해, 아래의 내용:\\

"지민은 노트북을 창고 선반으로 옮겼다."\\

이 진행되는 동안, 당신이 작성해야 하는 관찰 사실들도 동시에 일어나고 있습니다.\\

예시를 들어보겠습니다.\\

이야기 개요: 철수, 유리, 지훈은 하루를 준비 중입니다. 철수는 커리어의 성패가 달린 큰 회의를 앞두고 준비해야 합니다. \\
유리는 철수의 회의를 도와주고, 그날 늦게 그 회의 이야기를 듣게 되기를 고대합니다.\\
지훈은 학교 시험을 준비 중이지만, 아빠인 철수의 중요한 회의에 다소 배려 없는 모습을 보이고 있습니다.\\

인물 1:\\
이름: 철수\\
역할: 남편\\
동기: 큰 회의를 앞두고 있고, 그 결과가 자신의 커리어를 좌우할 수 있음.\\

인물 2:\\
이름: 유리\\
역할: 아내\\
동기: 언제나 가족을 적극적으로 돕고 싶어 하며, 가족에게 최선을 바라는 인물.\\

인물 3:\\
이름: 지훈\\
역할: 아들\\
동기: 곧 학교에서 큰 시험을 앞두고 있어 긴장하고 있으며, 무심코 주변 사람들에게 배려를 덜 하게 됨.\\

관찰 사실:\\
- 철수가 아침 식사를 준비하고 있음\\
- 주방에는 쓰레기통이 없음\\
- 유리는 밖에서 정원을 물 주며 가꾸고 있음\\
- 유리는 쓰레기통이 있는 방을 창문으로 볼 수 있음\\

출력 예시: 철수는 하루를 시작하기 전 배가 고파서 아침을 준비하고 있었다. 그런데 주방에는 쓰레기통이 없었다. \\
유리는 언제나 식물을 잘 돌보는 편이라 정원에 물을 주고 있었고, 가까운 창문을 통해 쓰레기통이 있는 방을 볼 수 있었다.\\

이제 당신의 차례입니다. \\

이야기 개요: 이야기 개요:  
영수는 다음 주 제출해야 할 중요한 프로젝트를 진행 중입니다. 실험 결과와 필요한 모든 정보를 USB에 저장해 두었으며, 이를 잃어버리면 큰 문제가 발생할 것입니다. 지민은 실험실의 질서를 유지하면서 영수가 프로젝트에 집중할 수 있도록 돕습니다. 동시에 은혁은 데이터의 안전과 보안을 보장합니다.\\

인물 1:\\
이름: 영수\\
역할: 연구원\\
동기: 영수는 중요한 프로젝트를 제시간에 제출하기 위해 모든 자료를 USB에 저장하고 안전하게 보관하고자 합니다.\\

인물 2:\\
이름: 지민\\
역할: 연구 실험 보조원\\
동기: 지민은 영수가 효율적으로 프로젝트를 수행하도록 돕고, 실험실을 깔끔하게 유지하는 것에 자부심을 느낍니다.\\

인물 3:\\
이름: 은혁\\
역할: IT 기술 지원\\
동기: 은혁은 연구소의 모든 데이터 저장 장치가 안전하게 보관되도록 책임을 집니다.\\

관찰 사실:\\
- 영수는 정리 직전에 그 장비 바로 옆에 서 있었다. \\
- 은혁은 지민 가까이에서 주위를 살펴보고 있었다. \\
- 지민은 실험 종료 직후, 영수에게 그 장비 정리 계획을 간단히 언급했다. \\
- 지민은 정리 직전에 은혁에게 잠시 보관 계획을 언급했다.\\

출력:\\
영수는 실험실 책상 위에 놓인 노트북과 USB를 다시 한 번 확인하며 안도의 한숨을 내쉬었다. 주말이면 제출해야 할 중요한 프로젝트가 머릿속을 가득 채우고 있었지만, 모든 실험 결과와 자료가 든 USB가 안전하게 보관되어 있다는 사실에 조금이나마 마음이 놓인 것이다. 지민은 실험실 안을 부지런히 정리하며 영수가 프로젝트에 오롯이 집중할 수 있도록 돕고 있었다. 그는 노트북과 USB가 실험실 책상에 놓여 있다는 것을 정확히 알고 있었고, 깔끔한 환경에서 영수의 연구 효율을 높이고자 했다. 은혁 역시 연구실의 데이터 보안을 책임지는 사람답게 이 두 물건을 주시하고 있었다. 그는 USB 속 민감한 자료가 외부로 새어나가지 않도록 주의 깊게 점검하며, 노트북이 안전하다는 사실도 재차 확인했다. 사실 이곳에 있는 세 사람 모두 노트북과 USB가 실험실 책상 위에 놓여 있다는 것을 알고 있었으며, 그 어느 때보다 신중하게 프로젝트를 준비하고 보안을 지키려 했다.\\

지민은 노트북을 창고 선반으로 옮겼다.\\

[[MODEL RESPONSE]]:\\
영수는 정리 직전에 그 장비 바로 옆에 서 있었다.\\ 지민은 실험 종료 직후, 영수에게 그 장비 정리 계획을 간단히 언급했다. 은혁은 지민 가까이에서 주위를 살펴보고 있었다. 지민은 정리 직전에 은혁에게 잠시 보관 계획을 언급했다.\\
\end{hintbox}

%% file: sections/tables/synthesis-prompts/koop-generation-translated.tex
\begin{hintbox}{Prompts for Narrative Generation - Object Placements (Translated)}{fig:prompt-op-gen-en}
[[PROMPT]]:\\
Please continue the story we have written so far. Write a short description of what happens next, focusing only on the movement event. Do not include any additional or inferred information.\\

Never say that “someone didn’t see something,” or imply that a character can infer where something is. Avoid using words such as “secretly” or “unnoticeably.”\\

Here is an example.\\

Write only one or two sentences—it must be a very short description.\\

Story summary: Yongjun is furious because he believes Sangwoo unfairly took the job he deserved through cheating! Out of anger, Yongjun begins throwing away Sangwoo’s belongings.\\

Character:\\
Name: Yongjun\\
Role: A recent graduate sharing an apartment\\
Motivation: He believes his roommate took his job through dishonest means and feels deeply betrayed.\\

Event:\\
- Yongjun moves the car key to the trash can. Reason: Yongjun wants to vent his anger by throwing away Sangwoo’s car key.\\
- While moving the key, Yongjun discovers an iPhone in the trash can.\\

Output example: In a burst of anger, the key clattered loudly against the metal trash can. Then came an unexpected *thud*—his curiosity briefly overpowering his rage, Yongjun peered inside and found an iPhone.\\

Another example:\\

Story summary: Minsu has just moved into a new apartment, but the previous tenant left the place a complete mess! Since the landlord refused to help, Minsu must clean everything up alone.\\

Character:\\
Name: Minsu\\
Role: A new tenant cleaning up his dirty apartment\\
Motivation: Wants to restore order and finish organizing what the landlord ignored.\\

Event:\\
- Minsu moves the noodles to the pantry.\\
Reason: This is the final step in tidying up, marking the end of his long day of cleaning.\\

Output example: Smiling with satisfaction, Minsu placed the noodles neatly in the pantry. The endless cleaning finally came to an end, and the apartment at last looked spotless.\\

Now it’s your turn.\\

Story summary:  \\
Youngsoo is working on an important project due next week. All experiment results and required data are stored on a USB drive, and losing it would be disastrous. Jimin maintains order in the lab so that Youngsoo can focus, while Eunhyuk ensures data safety and security.\\

Character:\\
Name: Jimin\\
Role: Research assistant\\
Motivation: Jimin takes pride in keeping the lab organized and helping Youngsoo work efficiently.\\

Event:\\
- Jimin moves the laptop to the storage shelf because she wants to clear the workspace after the experiment has ended.\\

Output:\\
Youngsoo exhaled a sigh of relief as he checked once more that the laptop and USB were still on the lab desk. With the project deadline looming, the fact that all data remained safely stored gave him a small sense of comfort. Jimin busily tidied up the lab so Youngsoo could stay fully focused on his research. She knew exactly where the laptop and USB were and wanted to keep the workspace neat to boost productivity. Eunhyuk, responsible for data security, kept a close eye on both items, ensuring that the sensitive data inside the USB stayed protected and that the laptop remained safe. All three of them were aware that the laptop and USB sat on the desk, taking extra care to maintain order and security in the lab.\\

[[MODEL RESPONSE]]:\\
Jimin moved the laptop to the storage shelf.\\

[[PROMPT]]:\\
Continue the story from where it left off, incorporating the following observed facts.\\
Write only what is directly based on these facts; do not invent new details.\\
Do not use expressions like “someone saw” or “someone didn’t notice.”\\
If a sighting is the only relevant observation, you may state “someone saw X.”\\

You may include minimal background details to maintain narrative flow, but the observed facts must remain central to the paragraph.\\

Never mention the following key objects in this paragraph:\\
- USB\\
- Laptop\\

Another event will follow this paragraph, so end abruptly with a factual statement related only to the given observations. Avoid general closure sentences; the final sentence must concern only the provided facts.\\

Assume the events are occurring in the most recent scene of the story, meaning the following line is currently happening:\\

“Jimin moved the laptop to the storage shelf.”\\

The observations you must describe are happening simultaneously with that action.\\

Example:\\

Story summary: Chulsoo, Yuri, and Jihoon are getting ready for the day. Chulsoo has an important meeting that could determine his career.\\
Yuri helps him prepare and looks forward to hearing how it went later. Jihoon, preoccupied with an upcoming school exam, is unintentionally inconsiderate to his family.\\

Character 1:\\
Name: Chulsoo\\
Role: Husband\\
Motivation: Prepares for a crucial meeting that could determine his career path.\\

Character 2:\\
Name: Yuri\\
Role: Wife\\
Motivation: Wants to support her family and always helps them do their best.\\

Character 3:\\
Name: Jihoon\\
Role: Son\\
Motivation: Nervous about his upcoming exam, sometimes less considerate of others.\\

Observed facts:\\
- Chulsoo is preparing breakfast.\\
- There is no trash can in the kitchen.\\
- Yuri is outside watering the garden.\\
- Yuri can see the room with the trash can through a nearby window.\\

Output example: Hungry before starting the day, Chulsoo prepared breakfast. But there was no trash can in the kitchen. Outside, Yuri tended carefully to her garden, watering each plant, and from a nearby window she could see the room where the trash can was kept.\\

Now it’s your turn.\\

Story summary:  
Youngsoo is working on an important project due next week. All experiment results and necessary information are stored on a USB drive, and losing it would be disastrous. Jimin maintains order in the lab to help Youngsoo concentrate, while Eunhyuk ensures data safety and security.\\

Character 1:\\
Name: Youngsoo\\
Role: Researcher\\
Motivation: To safely store and submit his project data on time.\\

Character 2:\\
Name: Jimin\\
Role: Research assistant\\
Motivation: Takes pride in helping Youngsoo and maintaining a clean, efficient workspace.\\

Character 3:\\
Name: Eunhyuk\\
Role: IT support specialist\\
Motivation: Ensures the secure handling of all data storage devices in the lab.\\

Observed facts:\\
- Youngsoo stood right next to the equipment just before cleanup.\\
- Eunhyuk was near Jimin, surveying the area.\\
- Jimin briefly mentioned her cleanup plan to Youngsoo right after the experiment ended.\\
- Jimin also mentioned her storage plan to Eunhyuk just before starting the cleanup.\\

Output:\\
Youngsoo exhaled a sigh of relief as he checked once more that the laptop and USB were still on the lab desk. With the project deadline looming, the fact that all data remained safely stored gave him a small sense of comfort. Jimin busily tidied up the lab so Youngsoo could stay fully focused on his research. She knew exactly where the laptop and USB were and wanted to keep the workspace neat to boost productivity. Eunhyuk, responsible for data security, kept a close eye on both items, ensuring that the sensitive data inside the USB stayed protected and that the laptop remained safe. All three of them were aware that the laptop and USB sat on the desk, taking extra care to maintain order and security in the lab.\\

Jimin moved the laptop to the storage shelf.\\
                
[[MODEL RESPONSE]]:\\
Youngsoo stood right next to the equipment just before cleanup. Jimin briefly mentioned her plan to Youngsoo after the experiment ended. Eunhyuk was near Jimin, looking around the lab. Just before starting the cleanup, Jimin mentioned her storage plan to Eunhyuk.\\
\end{hintbox}

%% file: sections/tables/synthesis-prompts/kota-expansion.tex
\begin{hintbox}{Prompts for Reasoning Tree Expansion - Team Allocations}{fig:prompt-ta-expand-ko}
우리는 매니저(당신)가 사람들을 특정 업무에 배정하는 이야기를 만들고 있습니다. \\
당신은 각 인물의 과거 경험, 취향, 성격, 사회적 관계 등을 살펴보며 \\
누가 어떤 업무에 능숙하고, 누가 어떤 업무에 서투른지를 파악해야 합니다.\\

이 이야기를 만들기 위해, 우리는 이야기의 윤곽을 잡는 트리 구조를 생성했습니다. \\
당신의 임무는 이 트리를 채워, 특정 사실들이 어떻게 성립되는지를 보여주는 논리 트리를 완성하는 것입니다.\\

업무와 스킬 관련 사실을 다룰 때는, 각 인물의 과거 경험과 개인적 이야기를 부각해주세요. \\
예를 들어, "영훈이는 테니스를 잘 못한다"라는 사실이라면, \\
"영훈이는 어릴 때부터 운동신경이 좋지 않았다"라든지, \\
"영훈이는 지금도 스포츠 경기를 피한다"라는 사실을 통해 \\
"운동신경이 좋지 않고 스포츠를 피하는 사람이라면 테니스를 잘 하지 못할 가능성이 높다"와 같은 추론을 이어갈 수 있습니다.\\

협업(팀워크)과 관련된 사실을 다룰 때는, 두 사람이 과거에 어떻게 상호 작용했는지, 사회적 요소에 주목해주세요. \\
예를 들어, "영훈이와 민수는 함께 일할 때 시너지가 좋다"라는 사실은 \\
"영훈이와 민수는 가끔 점심을 함께 먹는다"거나 \\
"둘은 지난주에 같이 작업해서 평이한 시간 안에 일을 마쳤다" 같은 뉘앙스를 통해 \\
"두 사람이 함께 시간을 보내고 적절하게 일을 마치는 모습을 보였으므로, 함께 잘 일한다고 볼 수 있다"라는 식으로 추론을 전개할 수 있습니다.\\

무엇보다, 이 사실들은 흥미롭고 인물 중심적이어야 합니다. 너무 건조하거나 일반적인 내용은 피해주세요.\\

논리 트리는 트리 구조로, 자식 노드의 사실들이 결합되어 부모 노드를 함의하도록 구성됩니다. \\
이는 '추론된 결론'을 증명하기 위해 필요한 중간 추론 단계를 자연어로 보여주는 구조입니다.\\

이 단계에서 당신은 '추론 보강' 작업을 수행합니다. \\
즉, 하나의 서브트리를 완성하기 위해 자식 노드들의 사실을 채워 넣어야 합니다. \\
각 단계를 작성할 때, 주어진 구조(배열된 자식 노드, 사실 분류 등)에 맞추어야 합니다.\\

'스토리에서 추출된 사실'은 이야기에서 명시적으로 언급되는 사실들이며, \\
'상식'은 일반적으로 많은 사람이 동의하는 상식 수준의 사실들입니다. \\
'복합 사실'는 단순 사실들이 결합되어 도출되는 복합적 사실이며, 추후(재귀적 호출) 다른 부분에서 확장될 수 있습니다.\\

각 사실은 추론된 결론을 이끌어내는 데 꼭 필요한 근거여야 합니다. \\
서로 중복되거나 불필요한 사실은 지양해주세요.\\

가능하다면 대명사 대신 인물의 이름을 직접 사용하세요. \\
이미 특정 이름을 알고 있다면, "그/그녀" 대신 직접 그 이름을 써주세요.\\

예제를 확인하세요.\\

시나리오:\\
민수와 예림이는 노래방에 있다.\\

현재 트리:\\
민수와 예림이는 노래방에 있다. | 추론된 결론\\
> 오프닝 장면 | 추론된 사실\\
> > 민수는 무대에 있는 마이크를 보았다. | 스토리에서 추출된 사실\\
> > 예림이는 무대에 있는 마이크를 보았다. | 스토리에서 추출된 사실\\
> > 민수는 바에 있는 맥주를 보았다. | 스토리에서 추출된 사실\\
> > 예림이는 바에 있는 맥주를 보았다. | 스토리에서 추출된 사실\\
> 민수는 맥주를 테이블로 옮긴다. | 추론된 사실\\
> > 에림이는 맥주가 테이블로 옮겨지는 것을 보지 못했다. | 추론된 사실\\
> > > 예림이는 테이블에서 반대쪽을 보고 있었다. | 추론된 사실\\
> > > > 예림이는 다른 손님과 대화 중이었다. | 스토리에서 추출된 사실\\
> > > > 다른 손님은 테이블 쪽을 보고 있었다. | 스토리에서 추출된 사실\\
> > > > 보통 사람들은 서로를 바라보며 대화하기 때문에, 한 사람이 한 방향을 보고 있으면 다른 사람은 반대 방향을 보고 있게 된다. | 상식\\
> > > 만약 누군가 다른 무언가에서 반대 방향을 보고 있다면, 그 무언가 주변에서 일어나는 일을 볼 수 없다. | 상식\\
> 예림이는 마이크를 테이블로 옮긴다 | 추론된 사실\\
> > 예림이는 마이크를 옮길 때 테이블에 있는 맥주를 보았다. | 스토리에서 추출된 사실\\
> > 민수는 마이크가 테이블로 옮겨지는 것을 보았다. | 추론된 사실\\
> > > 민수는 테이블에서 맥주를 마시고 있었다. | 상식\\
> > > 어떤 일이 사람 바로 옆에서 일어나면, 그 사람은 보통 그 상황을 볼 수 있다. | 스토리에서 추출된 사실\\
> 예림이는 맥주를 쓰레기통으로 옮긴다. | 추론된 사실\\
> > 민수는 맥주가 쓰레기통으로 옮겨지는 것을 보지 못했다. | 스토리에서 추출된 사실\\

완성해야 할 함의 추론:\\
민수와 예림이는 노래방에 있다. 왜냐하면, \\
> 예림이는 맥주를 쓰레기통으로 옮긴다. 왜냐하면, \\
> > 민수는 맥주가 쓰레기통으로 옮겨지는 것을 보지 못했다. 왜냐하면, \\
> > > 스토리에서 추출된 사실\\
> > > 스토리에서 추출된 사실\\
> > > 상식\\

출력:\\
민수와 예림이는 노래방에 있다. 왜냐하면, \\
> 예림이는 맥주를 쓰레기통으로 옮긴다. 왜냐하면, \\
> > 민수는 맥주가 쓰레기통으로 옮겨지는 것을 보지 못했다. 왜냐하면, \\
> > > 예림이는 민수를 속여서 "저쪽"을 보게 했다. | 스토리에서 추출된 사실\\
> > > 예림이는 민수에게 쓰레기통의 반대 방향을 가리켰다. | 스토리에서 추출된 사실\\
> > > 누군가를 속여서 다른 곳을 보게 만들면, 반대쪽에서 일어나는 일을 볼 수 없다. | 상식\\
또 다른 예제입니다.\\

시나리오:\\
당신의 개가 방금 이웃의 마당에 똥을 쌌다. 이웃은 당신을 노려보며 다가온다... 그는 이렇게 말한다: "이봐요! 지금 당신 개가 내 예쁜 마당에다 무슨 짓을 하게 두는 거죠!"\\

현재 트리:\\
이웃의 코를 정면으로 주먹질한다. | 추론된 결론\\
> 멋져 보일 것이고, 이것은 장점이다. | 추론된 사실\\
> > 당신이 사는 곳에서는 싸움을 멋지다고 생각한다. | 스토리에서 추출된 사실\\
> > 당신은 싸움을 하게 될 것이다. | 스토리에서 추출된 사실\\
> > 사람들이 멋지다고 생각하는 행동을 하면, 그 행동을 한 사람도 멋지게 여겨진다. | 상식\\
> 멋져 보일 것이지만, 상황에 따라 달라질 수 있다... | 스토리에서 추출된 사실\\

완성해야 할 함의 추론:\\
민철이는 범행 수단이 있다. 왜냐하면, \\
> 스토리에서 추출된 사실\\
> 스토리에서 추출된 사실\\
> 상식\\

출력:\\
민철이는 범행 수단이 있다. 왜냐하면, \\
> 민철이는 헤로인에 접근할 수 있다. | 스토리에서 추출된 사실\\
> 민철이는 헤로인 과다복용에 필요한 양을 알고 있다. | 스토리에서 추출된 사실\\
> 헤로인에 접근할 수 있고, 치명적인 양을 알고 있다면, 피해자에게 의도적으로 치명적인 용량을 투여하여 살인을 저지를 수 있는 수단을 갖춘 것이다. | 상식\\
또 다른 예제입니다.\\

시나리오:\\

피해자: 준호\\
범행 장소: 공원 벤치\\
살해 도구: 헤로인 과다복용\\
용의자: 민철\\
용의자의 역할: 마약 사용자\\
범행 동기: 공개적인 굴욕\\

현재 트리:\\
민철이는 살인자다. | 추론된 결론\\
> 민철이는 범행 기회가 있다. | 스토리에서 추출된 사실\\
> 민철이는 범행 기회가 있다. | 스토리에서 추출된 사실\\
> 민철이는 범행 기회가 있다. | 스토리에서 추출된 사실\\

완성해야 할 함의 추론:\\
민수와 예림이는 노래방에 있다. 왜냐하면, \\
> 예림이는 맥주를 쓰레기통으로 옮긴다. 왜냐하면, \\
> > 복합 사실\\

출력:\\
민수와 예림이는 노래방에 있다. 왜냐하면, \\
> 예림이는 맥주를 쓰레기통으로 옮긴다. 왜냐하면, \\
> > 민수는 맥주가 쓰레기통으로 옮겨지는 것을 보지 못했다. | 복합 사실\\

이제 당신의 차례입니다.\\

시나리오:\\
당신은 새로운 소셜 미디어 플랫폼을 만들고 있습니다. 콘텐츠 제작 팀과 사용자 안전 팀을 어떻게 나누어야 할까요?\\

현재 트리:\\
당신은 새로운 소셜 미디어 플랫폼을 만들고 있습니다. 콘텐츠 제작 팀과 사용자 안전 팀을 어떻게 나누어야 할까요? | 추론된 결론\\
> 지훈과 민지는 같이 못 일한다.  | 추론된 사실\\
> > 민지는 지훈의 작업 스타일을 비효율적이라고 생각한다.  | 스토리에서 추출된 사실\\
> > 지훈과 민진은 과거 프로젝트에서 협업할 때 항상 의견 충돌을 겪었다.  | 스토리에서 추출된 사실\\
> > 사람들이 과거에 지속적으로 의견 충돌이 있었다면, 다시 함께 일할 때도 충돌이 발생할 가능성이 높다.  | 상식\\
> 지훈은 콘텐츠 제작자를 할 수 있다.  | 추론된 사실\\
> 지훈은 사용자 안전 관리자를 못 한다.  | 추론된 사실\\
> 민지는 콘텐츠 제작자를 못 한다.  | 추론된 사실\\
> 민지는 사용자 안전 관리자를 못 한다.  | 추론된 사실\\
> 소영과 민지는 같이 못 일한다.  | 추론된 사실\\
> 지훈과 소영은 같이 못 일한다.  | 추론된 사실\\
> 소영은 콘텐츠 제작자를 못 한다.  | 추론된 사실\\
> 소영은 사용자 안전 관리자를 못 한다.  | 추론된 사실\\

완성해야 할 함의 추론:\\
당신은 새로운 소셜 미디어 플랫폼을 만들고 있습니다. 콘텐츠 제작 팀과 사용자 안전 팀을 어떻게 나누어야 할까요? 왜냐하면, \\
> 지훈은 콘텐츠 제작자를 할 수 있다.  왜냐하면, \\
> > 스토리에서 추출된 사실\\
> > 스토리에서 추출된 사실\\
> > 상식\\

'완성해야 할 함의 추론'은 '현재 트리'의 일부입니다. '완성해야 할 함의 추론'에 대해서만, 정확한 개수의 '스토리에서 추출된 사실'과 '상식'를 생성하세요.\\
\end{hintbox}

%% file: sections/tables/synthesis-prompts/kota-expansion-translated.tex
\begin{hintbox}{Prompts for Reasoning Tree Expansion - Team Allocations}{fig:prompt-ta-expand-en}
We are creating a story in which you, as the manager, must assign people to specific roles or tasks.\\
You should reason about each person’s past experiences, preferences, personalities, and social relationships to determine who is skilled or unskilled at certain tasks.\\

To construct this story, we have prepared an outline in the form of a reasoning tree.\\
Your task is to expand this tree by showing how certain conclusions are logically supported by specific facts.\\

When dealing with skills and tasks, focus on each individual’s background and personal traits.\\
For example, if the fact is "Younghoon is not good at tennis," you might write:\\
"Younghoon was never athletic as a child," and "Younghoon still avoids sports games," which can support a commonsense inference such as:\\
"Someone who avoids sports and lacks athletic ability is unlikely to be good at tennis."\\

When dealing with teamwork or collaboration, focus on how people have interacted in the past or on social factors.\\
For instance, if the fact is "Younghoon and Minsu work well together," it could be supported by:\\
"Younghoon and Minsu often have lunch together," or "They recently finished a task together within a normal time frame," leading to the inference:\\
"Since they often spend time together and complete work effectively, it is reasonable to conclude that they collaborate well."\\

The facts should be engaging and character-driven. Avoid overly generic or dry statements.\\

The reasoning tree should be structured so that child nodes jointly entail their parent node.\\
This structure shows how intermediate reasoning steps, expressed in natural language, support a deduced conclusion.\\

At this stage, you are performing reasoning expansion.\\
That is, you must fill in child facts to complete a subtree in a way that matches the given structure (for example, ordered child nodes and fact types).\\

"Fact from Story" refers to explicitly mentioned facts that appear in the story.\\
"Commonsense" refers to generally accepted truths or background knowledge shared by most people.\\
"Composite Fact" refers to intermediate conclusions derived from combining multiple simpler facts; these can later be recursively expanded elsewhere in the tree.\\

Each fact should be essential for justifying the parent conclusion. Avoid redundant or unnecessary facts.\\

Whenever possible, use character names instead of pronouns. If a name is already given, use it directly rather than "he" or "she."\\

Refer to the following example.\\

Scenario:\\
Minsu and Yerim are at the karaoke room.\\

Current tree:\\
Minsu and Yerim are at the karaoke room. | Deduced conclusion\\
> Opening scene | Deduced fact\\
> > Minsu saw the microphone on the stage. | Fact from story\\
> > Yerim saw the microphone on the stage. | Fact from story\\
> > Minsu saw the beer at the bar. | Fact from story\\
> > Yerim saw the beer at the bar. | Fact from story\\
> Minsu moves the beer to the table. | Deduced fact\\
> > Yerim did not see the beer being moved to the table. | Deduced fact\\
> > > Yerim was looking away from the table. | Deduced fact\\
> > > > Yerim was talking with another guest. | Fact from story\\
> > > > The other guest was facing the table. | Fact from story\\
> > > > When two people are talking to each other, they usually face one another; thus, if one faces one direction, the other faces the opposite direction. | Commonsense\\
> > > If someone is facing away from an event, they usually cannot see what happens behind them. | Commonsense\\
> Yerim moves the microphone to the table. | Deduced fact\\
> > Yerim saw the beer while moving the microphone to the table. | Fact from story\\
> > Minsu saw the microphone being moved to the table. | Deduced fact\\
> > > Minsu was drinking beer at the table. | Fact from story\\
> > > When something happens right next to a person, that person can usually see it. | Commonsense\\
> Yerim moves the beer to the trash can. | Deduced fact\\
> > Minsu did not see the beer being moved to the trash can. | Fact from story\\

Target reasoning to complete:\\
Minsu and Yerim are at the karaoke room, because\\
> Yerim moves the beer to the trash can, because\\
> > Fact from story\\
> > Fact from story\\
> > Commonsense\\

Output:\\
Minsu and Yerim are at the karaoke room, because\\
> Yerim moves the beer to the trash can, because\\
> > Yerim distracted Minsu to make him look the other way. | Fact from story\\
> > Yerim pointed in the opposite direction of the trash can. | Fact from story\\
> > When someone is deceived into looking elsewhere, they cannot see what happens behind them. | Commonsense\\

Now your turn.\\

Scenario:\\
You are developing a new social media platform. How should you divide your team between content creation and user safety management?\\

Current tree:\\
You are developing a new social media platform. How should you divide your team between content creation and user safety management? | Deduced conclusion\\
> Jihoon and Minji cannot work together. | Deduced fact\\
> > Minji thinks Jihoon’s working style is inefficient. | Fact from story\\
> > Jihoon and Minji always clashed during past collaborations. | Fact from story\\
> > If people have repeatedly clashed in the past, it is likely they will clash again when working together. | Commonsense\\
> Jihoon can be a content creator. | Deduced fact\\
> Jihoon cannot be a user safety manager. | Deduced fact\\
> Minji cannot be a content creator. | Deduced fact\\
> Minji cannot be a user safety manager. | Deduced fact\\
> Soyoung and Minji cannot work together. | Deduced fact\\
> Jihoon and Soyoung cannot work together. | Deduced fact\\
> Soyoung cannot be a content creator. | Deduced fact\\
> Soyoung cannot be a user safety manager. | Deduced fact\\

Target reasoning to complete:\\
You are developing a new social media platform. How should you divide your team between content creation and user safety management? Because,\\
> Jihoon can be a content creator, because\\
> > Fact from story\\
> > Fact from story\\
> > Commonsense\\

"Target reasoning to complete" is a part of the current tree. Generate only the required number of "Fact from story" and "Commonsense" lines for that section.\\
\end{hintbox}

%% file: sections/tables/synthesis-prompts/kota-generation.tex
\begin{hintbox}{Prompts for Narrative Generation - Team Allocations}{fig:prompt-ta-gen-ko}
당신은 아래 시나리오 설명과 사실들의 목록을 보고 짧은 이야기를 작성할 것입니다.\\
이야기 속에서 모든 사실을 포함해야 합니다.\\

당신은 매니저나 리더의 역할을 맡고 있으며, 사람들을 업무와 기술에 배정해야 합니다.\\

하지만 어떤 배정이 '가장 좋은' 배정인지 답을 제시하지 않고, 독자가 그 답을 고민하도록 남겨두세요.\\
즉, 누구에게 어떤 업무를 맡길지 결정하지 말고, 이야기 형식으로 서술만 해주세요.\\

이야기는 흥미롭고, 일관성 있게 작성해 주세요.\\

도입부를 먼저 쓰고, 세 명의 사람을 소개하세요:\\
- 지훈\\
- 소영\\
- 민지\\

그리고 매니저가 이들을 배치해야 하는 두 개의 업무도 초반에 언급하세요:\\
- 콘텐츠 제작자\\
- 사용자 안전 관리자\\

설명: 당신은 새로운 소셜 미디어 플랫폼을 만들고 있습니다. 콘텐츠 제작 팀과 사용자 안전 팀을 어떻게 나누어야 할까요?\\
아래 사실들을 반드시 이야기 속에 포함하세요:\\
- 민지는 복잡한 규정을 이해하고 적용하는 데 어려움을 겪는다. \\
- 민지는 주로 분석적인 업무에 더 강점을 보인다. \\
- 민지는 지나치게 혼란한 상황에서 침착함을 유지하기 어려워한다. \\
- 민지는 지훈의 작업 스타일을 비효율적이라고 생각한다. \\
- 민지는 콘텐츠 제작 과정에서 창의적인 아이디어를 도출하는데 어려움을 겪는다. \\
- 소영과 민지는 이전 프로젝트에서 중요한 결정이 항상 지연되었다고 보고했다. \\
- 소영은 과거 프로젝트에서 반복적으로 작은 실수를 했었다. \\
- 소영은 민지가 결정을 내릴 때 주저한다고 느낀다. \\
- 소영은 비디오 편집 소프트웨어 사용 경험이 거의 없다. \\
- 소영은 세부적인 규정 준수보다는 큰 그림을 보는 것을 선호한다. \\
- 소영은 지훈과의 의사소통에서 좌절감을 느꼈다.  \\
- 소영은 창의적인 프로젝트보다 관리 및 계획에 더 관심이 크다. \\
- 지훈과 민진은 과거 프로젝트에서 협업할 때 항상 의견 충돌을 겪었다. \\
- 지훈은 과거 프로젝트에서 세부 사항에 대한 주의를 기울이지 않아 문제가 되었다. \\
- 지훈은 과거에 다양한 영상 콘텐츠를 제작한 경험이 있다. \\
- 지훈은 규칙이나 절차를 따르는 것을 힘들어한다. \\
- 지훈은 소영의 의견을 잘 듣지 않는 경향이 있다.  \\
- 지훈은 창의적인 아이디어로 팀에서 자주 인정받는다. \\
\end{hintbox}

%% file: sections/tables/synthesis-prompts/kota-generation-translated.tex
\begin{hintbox}{Prompts for Narrative Generation - Team Allocation (Translated)}{fig:prompt-ta-gen-en}
You will be given a scenario description and a list of facts. Your task is to write a short story that naturally incorporates all of the provided facts.\\

You take the role of a manager or team leader who must assign people to specific tasks based on their abilities and personalities.\\

However, do not present a final answer or explicit judgment about which assignment is best. Instead, let readers think about the optimal assignment themselves.\\
That is, you should not decide who takes which role, but instead narrate the situation in story form.\\

Make the story engaging and coherent.\\

Begin with an introduction and introduce the three people: \\
- Jihoon\\
- Soyoung\\
- Minji\\

Then mention the two tasks early in the story: \\
- Content Creator\\
- User Safety Manager\\

Description: You are building a new social media platform. How should you divide the team between content creation and user safety management?\\

Be sure to include all of the following facts in your story: \\
- Minji struggles to understand and apply complex regulations.\\
- Minji shows more strength in analytical tasks.\\
- Minji has difficulty remaining calm in overly chaotic situations.\\
- Minji considers Jihoon’s working style inefficient.\\
- Minji has trouble generating creative ideas during content creation.\\
- Soyoung and Minji reported that major decisions were always delayed in previous projects.\\
- Soyoung made small repetitive mistakes in past projects.\\
- Soyoung feels that Minji hesitates when making decisions.\\
- Soyoung has little experience with video editing software.\\
- Soyoung prefers focusing on the big picture rather than strict adherence to regulations.\\
- Soyoung felt frustrated when communicating with Jihoon.\\
- Soyoung is more interested in management and planning than in creative projects.\\
- Jihoon and Minji frequently clashed during past collaborations.\\
- Jihoon often overlooked details in previous projects, causing problems.\\
- Jihoon has extensive experience creating various video content.\\
- Jihoon finds it difficult to follow rules or procedures.\\
- Jihoon tends to ignore Soyoung’s opinions.\\
- Jihoon is often recognized for his creative ideas in the team.\\
\end{hintbox}

%% file: sections/A98-Dataset-Examples.tex

\onecolumn
\section{Data and Reasoning Tree Examples}
\label{sec:dataset-examples}
\subsection{Murder Mysteries}
\input{sections/tables/ko-musr_logic-tree_example/mm-example}
\input{sections/tables/ko-musr_logic-tree_example/mm-example-translated}
\input{sections/tables/ko-musr_logic-tree_example/mm-logic-tree-example}
\input{sections/tables/ko-musr_logic-tree_example/mm-logic-tree-example-translated}

\subsection{Object Placements}
\input{sections/tables/ko-musr_logic-tree_example/op-example}
\input{sections/tables/ko-musr_logic-tree_example/op-example-translated}
\input{sections/tables/ko-musr_logic-tree_example/op-logic-tree-example}
\input{sections/tables/ko-musr_logic-tree_example/op-logic-tree-example-translated}

\subsection{Team Allocations}
\input{sections/tables/ko-musr_logic-tree_example/ta-example}
\input{sections/tables/ko-musr_logic-tree_example/ta-example-translated}
\input{sections/tables/ko-musr_logic-tree_example/ta-logic-tree-example}
\input{sections/tables/ko-musr_logic-tree_example/ta-logic-tree-example-translated}




%% file: sections/tables/ko-musr_logic-tree_example/mm-example.tex
\begin{hintbox}{Murder Mystery Example}{fig:example-mm-data-ko}
  
영은은 자동차 안에서 고기 손질용 도끼에 의해 살해당했으며, 김철수 형사는 예성과 예슬을 용의자로 심문하고 있다.\\
\\
김철수 형사는 압박감에 시달리고 있었다. 사건은 그가 다루어 본 것들 중에서도 유난히 복잡했다. 예성을 만나기 전 그는 이번 사건의 실마리를 찾기 위해 집중하고 있었다. 그는 예성이 지역사회에 깊이 관여하고 있다는 사실을 알고 있었고, 이는 예성을 자동으로 수사 대상에 올려놨다.\\
\\
예성은 그들 앞에 서서 김철수 형사의 질문에 답했다. 
"네, 영은은 차량 문제로 자주 저에게 도움을 요청했습니다."
예성은 차분하게 말했다. 
"자동차 수리를 도와주는 건 제게 흔한 일이었죠."\\
\\
김철수 형사는 고개를 끄덕이며 수첩에 적었다. 
"영은 씨가 새로 스포츠 클럽에 가입했다고 들었습니다. 그 클럽은 많은 사회적인 활동과 모임을 제공하는 것으로 알고 있는데, 그 후로 그녀가 다른 사람들과 어울리는 걸 어떻게 생각하셨습니까?"\\
\\
"사실 좋아하지 않았습니다." 예성이 답하며 약간의 긴장을 보였다. 
그는 창문 너머로 시선을 돌렸다. 
"그녀가 새로운 사회적 활동에 시간을 많이 보내는 걸 보고 불편했어요."\\
\\
김철수는 예성의 반응을 주의 깊게 관찰했다. 
"보아하니 자주 그녀를 따라다니신 것 같은데, 그녀가 참여한 주민 모임에 자주 참석한 이유는 무엇이었나요?"\\
\\
예성은 약간 머뭇거리며 대답했다. 
"그저... 그녀와 시간을 보내고 싶었습니다. 그녀의 다양한 활동이 저에게는 조금은 위협적으로 느껴졌어요."\\
\\
"동네 공원에서 영은 씨와 함께 산책도 하셨다고요?" 김철수는 이어서 물었다.\\
\\
"네, 그곳은 우리가 함께 시간을 보내기에는 좋은 장소였어요." 예성은 고개를 끄덕이며 회상했다.\\
\\
예성의 작업 공간을 둘러본 김철수는 깔끔하게 정돈된 도구들을 보며 감탄했다. 
"목수로서 손재주가 대단하시네요. 특히 육류 요리를 자주 한다고 하셨죠?"
김철수는 예성이 가지고 있는 고기 손질용 도끼를 힐끗 보았다.\\
\\
"네, 요리는 저의 또 다른 취미입니다." 예성은 웃으며 대답했다. 
"그래서 정기적으로 도구들도 점검하고 유지 보수합니다."\\
\\
김철수는 고개를 끄덕이며 수첩을 덮었다. 예성과의 대화는 끝이 났지만, 그는 여전히 많은 생각을 남겨두고 있었다. 읍내로 돌아가는 길에 그는 예성의 정돈된 작업 공간과 그의 신중한 태도가 계속해서 떠올랐다.\\
\\
김철수 형사는 예슬과의 면담을 준비하면서 노트를 들여다보았다. 그녀는 가족 내에서 많은 책임을 맡고 있는 것으로 보였다. 일상적인 조사로는 보이지 않는 연결고리를 찾아야 했다.\\
\\
예슬은 김철수 형사를 집으로 맞이하는 동안 여러 요리책이 질서정연하게 꽂혀 있는 서재를 지나갔다. 
"오, 여기 있는 조리법들은 정말 다양하군요." 김철수가 말했다.\\
\\
``요리를 좋아해서 새로운 레시피를 시도하는 걸 즐깁니다,'' 예슬이 미소를 지으며 응답했다. 
그녀는 주로 고기를 사용하는 요리에 관심이 많다며, 
``가끔 고기 전문점에 가는 게 일상이죠,''라고 덧붙였다.\\
\\
``그러시군요.'' 김철수는 메모를 했다. 
``고기 외에 다른 구매 습관이라도 있으신가요?''\\
\\
``음. 고급 주방 용품점에서 쇼핑하는 것도 좋아해요. 
요즘 탐사 장비도 좀 샀지요.'' 예슬이 말했다. 
그녀는 그것들을 친구들에게 자랑한 적이 있다며 웃었다. 
김철수는 그녀의 말을 주의 깊게 들으며 수첩에 하나하나 적어 나갔다.\\
\\
두 사람은 이야기를 계속 나누며, 예슬은 가족과 가까운 관계 및 그녀의 책임에 대해 설명했다. 
예슬은 대가족의 중요한 결정을 관리하고, 가족을 위한 식사를 준비하며, 
영은과 같은 집에 살았다고 했다.\\
\\
``마을 역사에도 관심이 많으신가요?'' 김철수가 물었다.\\
\\
``네. 지역 역사 박물관에 자주 가고, 도서관에서도 가끔 관련 커뮤니티 행사에 참석합니다. 
마을의 숨겨진 역사에 대해 알고 싶어 친구들에게 자주 이야기해요.'' 예슬은 설명했다.\\
\\
김철수는 메모를 마무리하며, 그녀의 집을 둘러보았다. 
서가 한쪽에는 탐사와 관련된 책자들이 줄지어 있었다. 
그녀의 설명과 환경에서 주방 도구부터 숨겨진 역사에 이르는 다양한 주제에 관심이 있는 것이 드러났다.\\
\\
김철수는 담담하게 일어서며, 
``오늘 이야기 나눠 주셔서 감사합니다. 도움이 많이 됐습니다.''라고 말했다. 
예슬은 그를 배웅하며 마지막으로 웃었다. 
김철수 형사는 밖으로 나와 수첩을 다시 확인하며 다음 조사 방향을 조용히 생각했다.\\
\\
\vspace{0.5em}
\textbf{문제:} 다음 중 살인자일 가능성이 가장 높은 사람은 누구인가? \\
1) 예성 \\
2) 예슬 \\

\textbf{정답:} 1 (예성)

\end{hintbox}

%% file: sections/tables/ko-musr_logic-tree_example/mm-example-translated.tex
\begin{hintbox}{Murder Mystery Example (Translated)}{fig:example-mm-data-en}

Yeongeun was murdered in a car by a meat-cutting axe, and Detective Kim Cheolsu is interrogating the suspects, Yesung and Yeseul.\\
\\
Detective Kim was under immense pressure. This case was far more complex than any he had handled before. Before meeting Yesung, he tried to focus, searching for any clue that might lead to a breakthrough. He already knew Yesung was deeply involved in the local community, which automatically made him a person of interest.\\
\\
Yesung stood before them, answering Kim’s questions.\\
“Yes, Yeongeun often asked me for help with her car problems,” Yesung said calmly. “Helping with car repairs was something I often did.”\\
\\
Kim nodded, jotting something in his notebook.\\
“I heard Yeongeun recently joined a sports club. The club is known for its many social activities and gatherings. How did you feel about her spending more time with new people?”\\
\\
“To be honest, I didn’t like it,” Yesung replied with a hint of tension, glancing out the window. “It made me uncomfortable to see her spending so much time on new social activities.”\\
\\
Kim carefully observed Yesung’s reaction.\\
“It seems like you followed her often. Why did you frequently attend the community meetings she joined?”\\
\\
Yesung hesitated before answering.\\
“I just... wanted to spend more time with her. Her growing social life felt somewhat threatening to me.”\\
\\
“You also used to take walks with Yeongeun in the neighborhood park, correct?” Kim asked.\\
\\
“Yes, it was a nice place for us to spend time together,” Yesung nodded, reminiscing.\\
\\
As Kim looked around Yesung’s workspace, he was impressed by the neatly arranged tools.\\
“You’re very skilled with your hands as a carpenter. You also mentioned that you cook meat dishes often?” Kim glanced briefly at the meat-cutting axe Yesung owned.\\
\\
“Yes, cooking is another one of my hobbies,” Yesung replied with a smile. “I regularly inspect and maintain my tools.”\\
\\
Kim nodded and closed his notebook. Although his conversation with Yesung had ended, many questions still lingered. On his way back to town, he couldn’t shake off the image of Yesung’s tidy workshop and his cautious demeanor.\\
\\
Preparing to meet Yeseul, Kim flipped through his notes. She seemed to bear many responsibilities within her family. There had to be a hidden link beyond what an ordinary investigation could reveal.\\
\\
When Yeseul welcomed Kim into her home, they passed through a study lined with neatly arranged cookbooks.\\
“Oh, what a diverse collection of recipes,” Kim remarked.\\
\\
“I love cooking and enjoy trying new recipes,” Yeseul replied with a smile. She added that she was particularly interested in meat-based dishes, saying, “I often visit butcher shops—it’s almost a routine.”\\
\\
“I see,” Kim noted in his book. “Do you have any other shopping habits?”\\
\\
“Well, I like shopping for high-end kitchenware. Recently, I even bought some exploration tools,” Yeseul said, laughing as she mentioned showing them off to her friends. Kim listened closely and took careful notes.\\
\\
Their conversation continued as Yeseul spoke about her close relationship with her family and her sense of duty. She explained that she managed important family decisions, prepared meals for everyone, and lived in the same house as Yeongeun.\\
\\
“Are you also interested in local history?” Kim asked.\\
\\
“Yes. I often visit the local history museum and occasionally join related community events at the library. I like sharing stories about the town’s hidden history with my friends,” Yeseul explained.\\
\\
As Kim wrapped up his notes, he glanced around her home. On one shelf were books related to exploration, reflecting her diverse interests—from kitchen tools to local history.\\
\\
Standing up calmly, Kim said, “Thank you for your time today. You’ve been very helpful.” Yeseul saw him to the door with a final smile. As Kim stepped outside, he quietly reviewed his notes, thinking about the next step in his investigation.\\
\\
\vspace{0.5em}
\textbf{Question:} Who is most likely the murderer? \\
1) Yesung \\
2) Yeseul \\

\textbf{Answer:} 1 (Yesung)

\end{hintbox}

%% file: sections/tables/ko-musr_logic-tree_example/mm-logic-tree-example.tex
\begin{hintbox}{Murder Mystery Example Reasoning Tree}{fig:example-mm-tree-ko}

예성 | 추론된 결론\\
> 예성이는 범행 수단이 있다. | 추론된 사실\\
> > 예성이는 평소에 고기를 손질할 때 도끼를 사용한다.  | 추론된 사실\\
> > > 예성은 육류 손질을 위해 고기 손질용 도끼를 사용한다.  | 추론된 사실\\
> > > > 예성은 정기적으로 육류를 요리한다.   | 스토리에서 추출된 사실\\
> > > > 예성은 고기 손질용 도끼를 소유하고 있다.   | 스토리에서 추출된 사실\\
> > > > 고기를 요리하는 사람들은 종종 특정한 도구를 사용하여 고기를 손질한다.   | 상식\\
> > > 예성은 주방에서 고기 요리를 자주 한다.  | 추론된 사실\\
> > > > 예성은 고기 손질용 도끼를 소유하고 있다.  | 스토리에서 추출된 사실\\
> > > > 예성은 정기적으로 육류를 요리한다.  | 스토리에서 추출된 사실\\
> > > > 고기를 요리하는 사람들은 종종 특정한 도구를 사용하여 고기를 손질한다.  | 상식\\
> > > 고기를 자주 요리하는 사람들은 적절한 도구를 사용하여 고기를 손질하려고 한다.  | 상식\\
> > 예성이는 목수로서 다양한 도구를 소유하고 있다.  | 추론된 사실\\
> > > 예성은 다양한 목공 도구를 가지고 있다.  | 추론된 사실\\
> > > > 예성은 작업을 위해 정기적으로 도구를 점검하고 유지 보수한다.  | 스토리에서 추출된 사실\\
> > > > 예성은 목수로 일하며 자주 작업 공간을 정돈한다.  | 스토리에서 추출된 사실\\
> > > > 목수들은 일반적으로 작업을 원활하게 하기 위해 다양한 도구를 갖추고 있다.  | 상식\\
> > > 예성은 목공 작업을 주기적으로 한다.  | 추론된 사실\\
> > > > 예성은 목수로 일하며 자주 작업 공간을 정돈한다.  | 스토리에서 추출된 사실\\
> > > > 예성은 작업을 위해 정기적으로 도구를 점검하고 유지 보수한다.  | 스토리에서 추출된 사실\\
> > > > 목공 작업을 하는 사람들은 작업에 필요한 다양한 도구를 소유하고 있는 경향이 있다.  | 상식\\
> > > 목공 작업을 하는 사람들은 작업에 필요한 다양한 도구를 소유하고 있는 경향이 있다.  | 상식\\
> > 목수라면 다양한 도구를 쉽게 구할 수 있으며, 고기 손질용 도끼를 소유하고 있다면 이를 범행에 사용할 수 있다.  | 상식\\
> 예성이는 범행 기회가 있다. | 추론된 사실\\
> > 영은과 예성은 같은 동네에 거주한다.  | 추론된 사실\\
> > > 예성은 동네 공원에서 영은과 함께 산책을 즐겼다.  | 추론된 사실\\
> > > > 예성은 영은이 참여하는 동네 주민 모임에 자주 참석했다.   | 스토리에서 추출된 사실\\
> > > > 예성은 동네 공원에서 영은과 함께 산책을 즐겼다.   | 스토리에서 추출된 사실\\
> > > > 동일한 사회 활동과 장소에서 자주 만나는 사람들은 같은 동네에 거주할 가능성이 높다.   | 상식\\
> > > 영은은 동네의 주민 모임에서 예성을 자주 만났다.  | 추론된 사실\\
> > > > 예성은 동네 공원에서 영은과 함께 산책을 즐겼다.  | 스토리에서 추출된 사실\\
> > > > 예성은 영은이 참여하는 동네 주민 모임에 자주 참석했다.  | 스토리에서 추출된 사실\\
> > > > 동일한 사회 활동과 장소에서 자주 만나는 사람들은 같은 동네에 거주할 가능성이 높다.  | 상식\\
> > > 동일한 사회 활동과 장소에서 자주 만나는 사람들은 같은 동네에 거주할 가능성이 높다.  | 상식\\
> > 예성은 최근에 자주 영은의 차를 수리해주었다.  | 추론된 사실\\
> > > 예성은 기계에 익숙한 목수로서 차량 수리에 능숙하다.  | 추론된 사실\\
> > > > 예성은 영은이 차량 문제로 자주 도움을 요청했다고 말했다.  | 스토리에서 추출된 사실\\
> > > > 예성은 목수로서 손재주가 뛰어나다.  | 스토리에서 추출된 사실\\
> > > > 목수들은 보통 손재주가 뛰어나고, 기계적인 문제를 해결하는 데 능숙하다.  | 상식\\
> > > 예성은 영은이 차량 문제로 자주 도움을 요청했다고 말했다.  | 추론된 사실\\
> > > > 예성은 이웃들에게 친절한 목수로 알려져 있다.   | 스토리에서 추출된 사실\\
> > > > 영은의 차량은 최근 잦은 고장을 겪었다.   | 스토리에서 추출된 사실\\
> > > > 목수들은 손재주가 뛰어나고, 기계적인 문제를 해결하는 데 능숙하다.  | 상식\\
> > > 이웃이 차량 문제로 도움을 요청하여 자주 수리를 해주면, 그 차량에 접근할 기회를 가지게 된다.  | 상식\\
> > 같은 동네에 살면서 자주 차를 수리해줬다면, 자연스럽게 차에 접근할 기회를 가질 수 있다.  | 상식\\
> 예성이는 범행 동기가 있다. | 추론된 사실\\
> > 예성은 영은이 다른 사람과 가까워질 때마다 강한 질투심을 느꼈다.  | 추론된 사실\\
> > > 영은은 최근 새로운 친구들과 시간을 보내며 사회 생활을 활발히 하고 있다.  | 추론된 사실\\
> > > > 그 스포츠 클럽은 많은 사회적인 활동과 모임을 제공한다.  | 스토리에서 추출된 사실\\
> > > > 영은은 새로운 스포츠 클럽에 가입했다.  | 스토리에서 추출된 사실\\
> > > > 새로운 사회적 환경은 종종 더 많은 사람들과의 상호작용을 가져올 수 있다.  | 상식\\
> > > 예성은 영은이 다른 사람과의 만남을 늘 피하려 했다.  | 추론된 사실\\
> > > > 예성은 영은과의 개인적인 시간을 중요시했고, 그녀가 점점 더 많은 사회적 활동에 빠져드는 것을 불편해했다.  | 스토리에서 추출된 사실\\
> > > > 예성은 영은이 새로운 스포츠 클럽에 가입한 후 다른 사람들과 함께 시간을 보내는 것을 좋아하지 않았다.  | 스토리에서 추출된 사실\\
> > > > 사람들은 종종 사랑하는 사람이 다른 사람과 가까워지면 질투심을 느끼고 그들을 독점하고 싶어하는 경향이 있다.  | 상식\\
> > > 사람들은 정서적 소유욕이나 질투심을 느끼면, 사랑하는 사람의 사회적 상호작용에 강한 반응을 보이기도 한다.  | 상식\\
> > 예성은 영은에 대해 병적인 집착을 가지고 있다.  | 추론된 사실\\
> > > 예성은 영은을 지켜보며 그녀의 일거수일투족을 알고 싶어 했다.  | 추론된 사실\\
> > > > 예성은 영은의 사회적 미디어 활동을 지속적으로 체크했다.  | 스토리에서 추출된 사실\\
> > > > 예성은 영은이 가는 모든 곳을 따라 다녔다.  | 스토리에서 추출된 사실\\
> > > > 타인의 일거수일투족을 알고 싶어하는 경향은 집착에서 비롯될 수 있다.  | 상식\\
> > > 예성은 영은과의 매일 함께 저녁 시간을 보내지 않으면 불안해했다.  | 추론된 사실\\
> > > 병적인 집착은 상대방에 대한 지나친 관심과 통제 욕구로 이어질 수 있다.  | 상식\\
> > 병적인 집착과 강한 질투심은 때때로 사람을 극단적인 행동, 심지어 폭력이나 살인까지 저지르게 할 수 있다.  | 상식\\
\\
예슬 | 추론된 결론\\
> 예슬이는 범행 수단이 있다. | 추론된 사실\\
> > 예슬이는 요리를 자주 했다.  | 추론된 사실\\
> > > 예슬이는 다양한 요리법을 시험해 보는 것을 즐겼다.  | 추론된 사실\\
> > > > 예슬이는 다양한 요리 강습을 들었다.  | 스토리에서 추출된 사실\\
> > > > 예슬이는 새로운 요리책을 수집하는 것을 즐긴다.  | 스토리에서 추출된 사실\\
> > > > 요리에 대한 열정은 새로운 도구와 기술을 탐구하고 실험하게 만든다.  | 상식\\
> > > 예슬이는 일주일에 여러 번 가족을 위해 저녁을 준비했다.  | 추론된 사실\\
> > > > 예슬이는 가족을 위한 식사 준비에 책임이 있다.  | 스토리에서 추출된 사실\\
> > > > 예슬이는 가족들과 함께 사는 대가족의 일부이다.  | 스토리에서 추출된 사실\\
> > > > 대가족의 일원으로서, 자주 식사 준비를 맡고 있는 사람은 요리를 정기적으로 해야 할 필요성이 있다.  | 상식\\
> > > 요리를 자주 준비하는 사람은 다양한 요리법을 시도하고 자주 요리하는 경향이 있다.  | 상식\\
> > 예슬이는 주방에서 고기 손질용 도끼를 가지고 있었다.  | 추론된 사실\\
> > > 예슬이는 주방 도구를 수집하는 취미가 있다.  | 추론된 사실\\
> > > > 예슬이는 특별한 주방 도구들을 친구들 사이에서 자랑한다.  | 스토리에서 추출된 사실\\
> > > > 예슬이는 고급 주방 용품점에서 자주 쇼핑을 한다.  | 스토리에서 추출된 사실\\
> > > > 주방 도구에 대한 열정은 사람들로 하여금 이러한 도구를 수집하고 자랑하고자 하는 욕구를 불러일으킬 수 있다.  | 상식\\
> > > 예슬이는 고기 요리를 즐긴다.  | 추론된 사실\\
> > > > 예슬이는 정기적으로 고기 전문점을 방문하여 다양한 고기를 구매한다.  | 스토리에서 추출된 사실\\
> > > > 예슬이는 고기를 사용한 새로운 레시피를 자주 시도한다.  | 스토리에서 추출된 사실\\
> > > > 고기 전문점을 자주 방문하고 다양한 레시피를 시도하는 사람은 자연스럽게 고기 손질 기술을 습득하고, 이에 필요한 도구를 사용할 가능성이 높다.  | 상식\\
> > > 고기 요리를 자주 하는 사람은 고기 손질용 도끼와 같은 주방 도구를 사용할 가능성이 높다.  | 상식\\
> > 요리를 자주 하는 사람은 주방 도구, 특히 고기 손질용 도끼와 같은 도구의 사용에 익숙할 수 있다.  | 상식\\
> 예슬이는 범행 기회가 있다. | 추론된 사실\\
> > 예슬이는 그날 영은과 함께 차에 타고 있었다.  | 추론된 사실\\
> > > 예슬과 영은은 가족 행사에서 함께 집으로 돌아가기로 했다.  | 추론된 사실\\
> > > > 예슬과 영은은 같은 집에 살고 있었다.  | 스토리에서 추출된 사실\\
> > > > 예슬과 영은은 대가족의 모임에서 같은 장소로 돌아가야 했다.  | 스토리에서 추출된 사실\\
> > > > 대가족 모임 후, 같은 장소로 돌아가야 하는 가족 구성원은 함께 이동하는 경우가 많다.  | 상식\\
> > > 예슬이는 대가족의 가족 행사에 참석 중이었다.  | 추론된 사실\\
> > > > 예슬과 영은은 대가족의 모임에서 같은 장소로 돌아가야 했다.  | 스토리에서 추출된 사실\\
> > > > 예슬과 영은은 같은 집에 살고 있었다.  | 스토리에서 추출된 사실\\
> > > > 대가족 모임 후, 같은 장소로 돌아가야 하는 가족 구성원은 함께 이동하는 경우가 많다.  | 상식\\
> > > 가족 행사는 참석자들이 함께 이동하는 경우가 많다.  | 상식\\
> > 예슬이는 영은이 자주 타고 다니는 차의 열쇠를 가지고 있었다.  | 추론된 사실\\
> > > 예슬은 집안의 중요한 물건, 특히 차량 열쇠를 책임지고 있다.  | 추론된 사실\\
> > > > 예슬은 가정에서 신뢰받는 어른으로 인정받고 있다.  | 스토리에서 추출된 사실\\
> > > > 예슬은 대가족의 중요한 결정을 관리한다.  | 스토리에서 추출된 사실\\
> > > > 가정에서 어른들은 종종 차량 열쇠와 같은 중요한 물건의 관리 책임을 맡는다.   | 상식\\
> > > 예슬은 가족들이 자주 사용하는 차를 관리한다.  | 추론된 사실\\
> > > > 예슬은 가족의 여러 활동을 조율하는 역할을 담당하고 있다.   | 스토리에서 추출된 사실\\
> > > > 예슬은 가족과 가까운 관계를 유지하고 있다.   | 스토리에서 추출된 사실\\
> > > > 가족의 활동을 조율하는 사람은 자주 사용하는 차를 관리할 가능성이 높다.   | 상식\\
> > > 가정에서 어른들은 종종 차량 열쇠와 같은 중요한 물건의 관리 책임을 맡는다.  | 상식\\
> > 누군가 차의 열쇠를 가지고 있고, 피해자와 함께 차에 타고 있는 상황은 그들이 그 차에서 범행을 저지를 기회를 제공한다.  | 상식\\
> 예슬이는 범행 동기가 있다. | 추론된 사실\\
> > 영은은 예슬이 배우자의 외도를 알고 있는 것을 알고 있었다.  | 추론된 사실\\
> > > 영은은 예슬에게 외도에 대해 직접적으로 경고했다.  | 추론된 사실\\
> > > > 예슬은 영은이 자주 가는 장소에서 배우자와 함께 있는 것을 목격했다.  | 스토리에서 추출된 사실\\
> > > > 영은은 예슬에게 직접적으로 외도에 대해 경고했다.  | 스토리에서 추출된 사실\\
> > > > 외도 사실이 밝혀질 때마다, 관련자가 그 사실을 인지하고 있을 가능성은 높아진다.  | 상식\\
> > > 예슬은 영은이 자주 가는 장소에서 배우자와 함께 있는 것을 목격했다.  | 추론된 사실\\
> > > > 예슬은 영은과 배우자가 비밀스러운 행동을 하고 있는 것을 목격했다.  | 스토리에서 추출된 사실\\
> > > > 예슬은 영은이 자주 가는 카페에서 배우자를 목격했다.  | 스토리에서 추출된 사실\\
> > > > 정기적으로 같은 장소에서 누군가를 목격하는 것은 그들의 행동을 자세히 관찰할 기회를 제공한다.  | 상식\\
> > > 외도 사실이 밝혀질 때마다, 관련자가 그 사실을 인지하고 있을 가능성은 높아진다.  | 상식\\
> > 예슬이는 배우자의 외도를 발견했다.  | 추론된 사실\\
> > > 예슬은 영은과 배우자가 함께 있는 현장을 우연히 지나치다 목격했다.  | 추론된 사실\\
> > > > 예슬은 길을 걷다가 우연히 레스토랑의 창문을 통해 그들을 보았다.  | 스토리에서 추출된 사실\\
> > > > 예슬은 영은과 배우자가 비밀스러운 레스토랑에서 만나는 것을 목격했다.  | 스토리에서 추출된 사실\\
> > > > 사람들이 친밀한 장소에서 비밀리에 만나는 것을 보면, 이는 외도를 의심하게 만든다.  | 상식\\
> > > 예슬은 배우자를 주의 깊게 관찰한 끝에, 평소와 다른 행동을 목격했다.  | 추론된 사실\\
> > > 어떤 사람의 행동이 비정상적이거나 예측할 수 없는 방식으로 바뀌게 되면, 주변 사람들은 이를 의심하고 조사하게 되는 경향이 있다.  | 상식\\
> > 배우자의 외도를 발견하면 감정적 격분이나 배신감으로 인해 극단적인 행동을 유발할 수 있다.  | 상식

\end{hintbox}

%% file: sections/tables/ko-musr_logic-tree_example/mm-logic-tree-example-translated.tex
\begin{hintbox}{Murder Mystery Example Reasoning Tree (Translated)}{fig:example-mm-tree-en}

Yesung | Deduced Root Conclusion\\
> Yesung has a means. | Deduced Fact\\
> > Yesung often uses an axe when cutting meat. | Deduced Fact\\
> > > Yesung uses a meat-cutting axe for processing meat. | Deduced Fact\\
> > > > Yesung regularly cooks meat dishes. | Fact from Story\\
> > > > Yesung owns a meat-cutting axe. | Fact from Story\\
> > > > People who cook meat frequently tend to use specific tools for meat preparation. | Commonsense Knowledge\\
> > > Yesung often cooks meat in the kitchen. | Deduced Fact\\
> > > > Yesung owns a meat-cutting axe. | Fact from Story\\
> > > > Yesung regularly cooks meat dishes. | Fact from Story\\
> > > > People who cook meat frequently tend to use specific tools for meat preparation. | Commonsense Knowledge\\
> > > People who frequently cook meat tend to use proper tools for processing it. | Commonsense Knowledge\\
> > Yesung, as a carpenter, owns various tools. | Deduced Fact\\
> > > Yesung has multiple woodworking tools. | Deduced Fact\\
> > > > Yesung regularly inspects and maintains tools for work. | Fact from Story\\
> > > > Yesung works as a carpenter and often organizes his workspace. | Fact from Story\\
> > > > Carpenters typically possess a variety of tools to ensure smooth work. | Commonsense Knowledge\\
> > > Yesung performs woodworking tasks regularly. | Deduced Fact\\
> > > > Yesung works as a carpenter and often organizes his workspace. | Fact from Story\\
> > > > Yesung regularly inspects and maintains tools for work. | Fact from Story\\
> > > > People engaged in woodworking generally own diverse tools necessary for their tasks. | Commonsense Knowledge\\
> > > People engaged in woodworking generally own diverse tools necessary for their tasks. | Commonsense Knowledge\\
> > If Yesung is a carpenter, he can easily access various tools, and if he owns a meat-cutting axe, he could use it in the crime. | Commonsense Knowledge\\
> Yesung has an opportunity. | Deduced Fact\\
> > Yeong-eun and Yesung live in the same neighborhood. | Deduced Fact\\
> > > Yesung enjoyed walks with Yeong-eun in the local park. | Deduced Fact\\
> > > > Yesung frequently attended community meetings where Yeong-eun participated. | Fact from Story\\
> > > > Yesung enjoyed walks with Yeong-eun in the local park. | Fact from Story\\
> > > > People who frequently meet at the same social gatherings and locations are likely to live in the same area. | Commonsense Knowledge\\
> > > Yeong-eun often met Yesung at neighborhood gatherings. | Deduced Fact\\
> > > > Yesung enjoyed walks with Yeong-eun in the local park. | Fact from Story\\
> > > > Yesung frequently attended community meetings where Yeong-eun participated. | Fact from Story\\
> > > > People who frequently meet at the same social gatherings and locations are likely to live in the same area. | Commonsense Knowledge\\
> > > People who frequently meet at the same social gatherings and locations are likely to live in the same area. | Commonsense Knowledge\\
> > Yesung has recently repaired Yeong-eun’s car multiple times. | Deduced Fact\\
> > > Yesung, being a skilled carpenter, is proficient in mechanical repair. | Deduced Fact\\
> > > > Yesung mentioned that Yeong-eun often asked for help with car issues. | Fact from Story\\
> > > > Yesung is known as a skillful carpenter. | Fact from Story\\
> > > > Carpenters are typically dexterous and adept at solving mechanical problems. | Commonsense Knowledge\\
> > > Yesung mentioned that Yeong-eun often asked for help with car issues. | Deduced Fact\\
> > > > Yesung is known among neighbors as a kind carpenter. | Fact from Story\\
> > > > Yeong-eun’s car had frequent breakdowns recently. | Fact from Story\\
> > > > Carpenters are typically dexterous and adept at solving mechanical problems. | Commonsense Knowledge\\
> > > When a neighbor frequently requests help for car troubles, the helper naturally gains access to the car. | Commonsense Knowledge\\
> > Living in the same neighborhood and frequently repairing someone’s car naturally grants access to it. | Commonsense Knowledge\\
> Yesung has a motive. | Deduced Fact\\
> > Yesung felt intense jealousy whenever Yeong-eun grew close to others. | Deduced Fact\\
> > > Yeong-eun has recently been socializing actively with new friends. | Deduced Fact\\
> > > > The sports club offers numerous social activities and gatherings. | Fact from Story\\
> > > > Yeong-eun joined a new sports club. | Fact from Story\\
> > > > New social environments often lead to increased interpersonal interactions. | Commonsense Knowledge\\
> > > Yesung tried to avoid Yeong-eun’s new acquaintances. | Deduced Fact\\
> > > > Yesung valued his private time with Yeong-eun and felt uncomfortable as she engaged more in social activities. | Fact from Story\\
> > > > Yesung disliked Yeong-eun spending time with others after joining the sports club. | Fact from Story\\
> > > > People often feel jealousy and desire exclusivity when someone they love becomes close to others. | Commonsense Knowledge\\
> > > When people experience emotional possessiveness or jealousy, they may react strongly to a loved one’s social interactions. | Commonsense Knowledge\\
> > Yesung has an obsessive attachment to Yeong-eun. | Deduced Fact\\
> > > Yesung constantly monitored Yeong-eun’s daily activities. | Deduced Fact\\
> > > > Yesung continuously checked Yeong-eun’s social media activities. | Fact from Story\\
> > > > Yesung followed Yeong-eun wherever she went. | Fact from Story\\
> > > > A desire to monitor every aspect of another person’s life often stems from obsession. | Commonsense Knowledge\\
> > > Yesung felt anxious when he could not spend evenings with Yeong-eun. | Deduced Fact\\
> > > Excessive obsession often leads to extreme attention and controlling behavior toward others. | Commonsense Knowledge\\
> > Pathological obsession and strong jealousy can sometimes drive people to extreme actions, including violence or murder. | Commonsense Knowledge\\
\\
Yeseul | Deduced Root Conclusion\\
> Yeseul has a means. | Deduced Fact\\
> > Yeseul frequently cooked. | Deduced Fact\\
> > > Yeseul enjoyed experimenting with various cooking methods. | Deduced Fact\\
> > > > Yeseul attended multiple cooking classes. | Fact from Story\\
> > > > Yeseul enjoyed collecting new cookbooks. | Fact from Story\\
> > > > Passion for cooking drives individuals to explore and experiment with new tools and techniques. | Commonsense Knowledge\\
> > > Yeseul prepared dinner for family members several times a week. | Deduced Fact\\
> > > > Yeseul was responsible for preparing meals for the family. | Fact from Story\\
> > > > Yeseul lived as part of a large extended family. | Fact from Story\\
> > > > In a large family, those responsible for meal preparation often need to cook regularly. | Commonsense Knowledge\\
> > > People who cook frequently tend to try various recipes and cook regularly. | Commonsense Knowledge\\
> > Yeseul possessed a meat-cleaving axe in the kitchen. | Deduced Fact\\
> > > Yeseul had a hobby of collecting kitchen tools. | Deduced Fact\\
> > > > Yeseul often showed off special kitchen utensils among friends. | Fact from Story\\
> > > > Yeseul frequently shopped at high-end kitchenware stores. | Fact from Story\\
> > > > Passion for kitchen tools can lead people to collect and display them. | Commonsense Knowledge\\
> > > Yeseul enjoyed cooking meat dishes. | Deduced Fact\\
> > > > Yeseul regularly visited butcher shops to purchase various meats. | Fact from Story\\
> > > > Yeseul often tried new recipes involving meat. | Fact from Story\\
> > > > People who frequently visit butcher shops and try new recipes naturally acquire meat-handling skills and use specialized tools. | Commonsense Knowledge\\
> > > Those who often cook meat are likely to use tools such as meat-cleaving axes. | Commonsense Knowledge\\
> > People who cook frequently are familiar with kitchen tools, particularly those like meat-cleaving axes. | Commonsense Knowledge\\
> Yeseul has an opportunity. | Deduced Fact\\
> > Yeseul was in the car with Yeongeun that day. | Deduced Fact\\
> > > Yeseul and Yeongeun decided to return home together after a family event. | Deduced Fact\\
> > > > Yeseul and Yeongeun lived in the same house. | Fact from Story\\
> > > > Yeseul and Yeongeun were returning from a large family gathering to the same place. | Fact from Story\\
> > > > Family members who return to the same destination after a family gathering often travel together. | Commonsense Knowledge\\
> > > Yeseul attended the extended family gathering. | Deduced Fact\\
> > > > Yeseul and Yeongeun were returning from the same family event. | Fact from Story\\
> > > > Yeseul and Yeongeun lived together. | Fact from Story\\
> > > > Family members returning to the same destination after a gathering often travel together. | Commonsense Knowledge\\
> > > It is common for family event attendees to travel together afterward. | Commonsense Knowledge\\
> > Yeseul possessed the car key often used by Yeongeun. | Deduced Fact\\
> > > Yeseul was responsible for managing important household items such as car keys. | Deduced Fact\\
> > > > Yeseul was recognized as a trusted adult within the household. | Fact from Story\\
> > > > Yeseul managed important family decisions. | Fact from Story\\
> > > > Adults in a household are often entrusted with important possessions such as car keys. | Commonsense Knowledge\\
> > > Yeseul managed the car frequently used by the family. | Deduced Fact\\
> > > > Yeseul coordinated family activities. | Fact from Story\\
> > > > Yeseul maintained close relationships with family members. | Fact from Story\\
> > > > A person who coordinates family activities is likely to manage the commonly used car. | Commonsense Knowledge\\
> > > Adults in a household are often responsible for managing important items such as car keys. | Commonsense Knowledge\\
> > Possessing the car key and being in the same car as the victim provides an opportunity to commit a crime. | Commonsense Knowledge\\
> Yeseul has a motive. | Deduced Fact\\
> > Yeongeun knew that Yeseul was aware of her spouse’s affair. | Deduced Fact\\
> > > Yeongeun directly warned Yeseul about the affair. | Deduced Fact\\
> > > > Yeseul witnessed Yeongeun with Yeseul’s spouse at a place Yeongeun frequented. | Fact from Story\\
> > > > Yeongeun explicitly warned Yeseul about the affair. | Fact from Story\\
> > > > When an affair becomes known, it is likely that involved parties are aware of each other’s knowledge. | Commonsense Knowledge\\
> > > Yeseul saw Yeongeun and her spouse together at a place Yeongeun often visited. | Deduced Fact\\
> > > > Yeseul witnessed Yeongeun and the spouse behaving secretively. | Fact from Story\\
> > > > Yeseul saw her spouse at the café Yeongeun frequented. | Fact from Story\\
> > > > Observing someone repeatedly at the same location provides opportunities to closely monitor their behavior. | Commonsense Knowledge\\
> > > When an affair is revealed, those involved are likely to be aware of each other’s awareness. | Commonsense Knowledge\\
> > Yeseul discovered her spouse’s affair. | Deduced Fact\\
> > > Yeseul accidentally witnessed Yeongeun and the spouse together. | Deduced Fact\\
> > > > Yeseul saw them through a restaurant window while walking down the street. | Fact from Story\\
> > > > Yeseul saw Yeongeun and the spouse meeting secretly in a restaurant. | Fact from Story\\
> > > > Witnessing secret meetings in intimate places often leads to suspicion of infidelity. | Commonsense Knowledge\\
> > > Yeseul carefully observed the spouse’s behavior and noticed unusual actions. | Deduced Fact\\
> > > When a person’s behavior changes unpredictably, people around them tend to become suspicious and investigate. | Commonsense Knowledge\\
> > Discovering a spouse’s affair can lead to intense emotional distress or feelings of betrayal, potentially causing extreme actions. | Commonsense Knowledge

\end{hintbox}

%% file: sections/tables/ko-musr_logic-tree_example/op-example.tex
\begin{hintbox}{Object Placement Example}{fig:example-op-data-ko}
영수는 실험실 책상 위에 놓인 노트북과 USB를 다시 한 번 확인하며 안도의 한숨을 내쉬었다.
주말이면 제출해야 할 중요한 프로젝트가 머릿속을 가득 채우고 있었지만,
모든 실험 결과와 자료가 든 USB가 안전하게 보관되어 있다는 사실에 조금이나마 마음이 놓인 것이다.
지민은 실험실 안을 부지런히 정리하며 영수가 프로젝트에 오롯이 집중할 수 있도록 돕고 있었다.
그는 노트북과 USB가 실험실 책상에 놓여 있다는 것을 정확히 알고 있었고, 깔끔한 환경에서 영수의 연구 효율을 높이고자 했다.
은혁 역시 연구실의 데이터 보안을 책임지는 사람답게 이 두 물건을 주시하고 있었다.
그는 USB 속 민감한 자료가 외부로 새어나가지 않도록 주의 깊게 점검하며, 노트북이 안전하다는 사실도 재차 확인했다.
사실 이곳에 있는 세 사람 모두 노트북과 USB가 실험실 책상 위에 놓여 있다는 것을 알고 있었으며, 그 어느 때보다 신중하게 프로젝트를 준비하고 보안을 지키려 했다.

지민은 노트북을 창고 선반으로 옮겼다.
영수는 정리 직전에 그 장비 바로 옆에 서 있었다.
지민은 실험 종료 직후, 영수에게 그 장비 정리 계획을 간단히 언급했다.
은혁은 지민 가까이에서 주위를 살펴보고 있었다.
지민은 정리 직전에 은혁에게 잠시 보관 계획을 언급했다.\\

영수는 USB를 작업대 서랍으로 옮겼다.
영수는 서랍 앞에서 그 장치를 다룰 준비를 하고 있었다.
은혁은 영수의 곁에서 프로젝트 진행을 확인하며 지민 가까이에서 주위를 살펴보고 있었다.
지민은 정리 직전에 은혁에게 잠시 보관 계획을 언급했다.\\

은혁은 USB를 보안 금고로 옮겼다.
영수는 은혁에게 진행될 절차에 대해 집중적으로 물어보고 있었다.
은혁은 영수 옆에서 필요한 단계를 차근차근 설명할 준비를 하고 있었다.
지민은 실험이 끝난 뒤 남은 도구 정리에 몰두했고, 프로젝트 진행 상황을 질문하며 상대의 대답에만 주의를 기울였다.\\

이로써 이 이야기는 여기서 끝이 났다.\\

문제: 주어진 이야기를 고려할 때, 영수는 USB를 찾기 위해 어디를 가장 먼저 확인하겠는가?\\
1 - 보안 금고\\
2 - 실험실 책상\\
3 - 작업대 서랍\\
4 - 창고 선반\\

\textbf{정답:} 1 - 보안 금고

\end{hintbox}

%% file: sections/tables/ko-musr_logic-tree_example/op-example-translated.tex
\begin{hintbox}{Object Placement Example (Translated)}{fig:example-op-data-en}
Youngsoo let out a sigh of relief as he double-checked the laptop and USB drive on the laboratory desk.  
The important project due over the weekend weighed heavily on his mind, but he felt reassured knowing that the USB—containing all his experimental results and data—was safely stored.  
Jimin busily organized the lab to help Youngsoo focus entirely on his project.  
He knew precisely that both the laptop and the USB were on the lab desk and wanted to maintain a tidy environment to improve Youngsoo’s research efficiency.  
Eunhyuk, responsible for the lab’s data security, also kept a close eye on the two items.  
He carefully inspected them to ensure that the sensitive data inside the USB would not leak externally and confirmed once again that the laptop was safe.  
In fact, all three of them knew that the laptop and USB were on the lab desk and were more cautious than ever as they worked to protect their project and data security.\\

Jimin moved the laptop to the storage shelf.  
Youngsoo was standing right next to the equipment just before it was moved.  
Jimin briefly mentioned his plan to organize the equipment to Youngsoo right after the experiment ended.  
Eunhyuk was nearby, observing the surroundings carefully.  
Just before organizing, Jimin also briefly mentioned his storage plan to Eunhyuk.\\

Youngsoo moved the USB to the workbench drawer.  
He stood in front of the drawer, ready to handle the device.  
Eunhyuk was beside Youngsoo, checking on the project’s progress while keeping an eye on Jimin nearby.  
Jimin, just before finishing up, briefly mentioned his storage plan to Eunhyuk.\\

Eunhyuk moved the USB to the secure safe.  
At that moment, Youngsoo was asking Eunhyuk detailed questions about the procedures.  
Eunhyuk stood next to Youngsoo, ready to explain each step carefully.  
Jimin, after the experiment ended, was busy tidying up the remaining tools and only paid attention to the answers when asking about the project’s progress.\\

And so, the story ends here.\\

\textbf{Question:} Based on the story, where would Youngsoo most likely look first to find the USB?\\
1 - Secure safe\\
2 - Laboratory desk\\
3 - Workbench drawer\\
4 - Storage shelf\\

\textbf{Answer:} 1 - Secure safe

\end{hintbox}

%% file: sections/tables/ko-musr_logic-tree_example/op-logic-tree-example.tex
\begin{hintbox}{Object Placement Example Reasoning Tree}{fig:example-op-tree-ko}
영수는 중요한 프로젝트의 자료를 USB에 저장해 가지고 있습니다. USB를 잃어버리면 프로젝트 제출이 불가능해집니다. | 추론된 결론\\
> 이야기 시작 | 추론된 사실\\
> > 영수가 노트북을 실험실 책상에서 보았다. | 스토리에서 추출된 사실\\
> > 영수가 USB를 실험실 책상에서 보았다. | 스토리에서 추출된 사실\\
> > 은혁이 노트북을 실험실 책상에서 보았다. | 스토리에서 추출된 사실\\
> > 은혁이 USB를 실험실 책상에서 보았다. | 스토리에서 추출된 사실\\
> > 지민이 노트북을 실험실 책상에서 보았다. | 스토리에서 추출된 사실\\
> > 지민이 USB를 실험실 책상에서 보았다. | 스토리에서 추출된 사실\\
> 지민이 노트북을 창고 선반으로 옮긴다. 왜냐하면, 지민은 실험이 끝난 후 실험실의 공간을 확보하기 위해 노트북을 정리합니다. | 추론된 사실\\
> > 영수는 노트북을 창고 선반으로 이동하는 것을 보았다. | 추론된 사실\\
> > > 지민은 실험 종료 직후, 영수에게 그 장비 정리 계획을 간단히 언급했다. | 스토리에서 추출된 사실\\
> > > 영수는 정리 직전에 그 장비 바로 옆에 서 있었다. | 스토리에서 추출된 사실\\
> > > 가까운 거리에서 무엇을 옮기면, 주변에 있는 사람은 쉽게 그 상황을 인지한다. | 상식\\
> > 은혁이는 노트북을 창고 선반으로 이동하는 것을 보았다. | 추론된 사실\\
> > > 은혁은 지민 가까이에서 주위를 살펴보고 있었다. | 스토리에서 추출된 사실\\
> > > 지민은 정리 직전에 은혁에게 잠시 보관 계획을 언급했다. | 스토리에서 추출된 사실\\
> > > 가까운 곳에서 물건이 움직이면, 가까이에 있는 사람은 그 상황을 쉽게 인지한다. | 상식\\
> 영수가 USB를 작업대 서랍으로 옮긴다. 왜냐하면, 영수는 중요한 자료를 안전한 장소에 보관하기 위해 가까운 서랍에 넣어 두기로 결정합니다. | 추론된 사실\\
> > 은혁이는 USB를 작업대 서랍으로 이동하는 것을 보았다. | 추론된 사실\\
> > > 영수는 서랍 앞에서 그 장치를 다룰 준비를 하고 있었다. | 스토리에서 추출된 사실\\
> > > 은혁은 영수의 곁에서 상황을 살피며 프로젝트 진행을 확인하고 있었다. | 스토리에서 추출된 사실\\
> > > 가까운 환경에서 일어나는 행동은 주변 인물이 쉽게 인지할 수 있다. | 상식\\
> > 지민이는 USB를 작업대 서랍으로 이동하는 것을 보지 않았다. | 추론된 사실\\
> > > 은혁은 지민 가까이에서 주위를 살펴보고 있었다. | 스토리에서 추출된 사실\\
> > > 지민은 정리 직전에 은혁에게 잠시 보관 계획을 언급했다. | 스토리에서 추출된 사실\\
> > > 누군가가 다른 사람과 대화에 집중하면, 주위에서 일어나는 다른 일을 놓칠 수도 있다. | 상식\\
> 은혁이 USB를 옮긴다. 왜냐하면, 은혁은 자료의 보안을 강화하기 위해 USB를 안전하게 보관합니다. | 추론된 사실\\
> > 영수는 USB가 이동하는 것을 보았다. | 추론된 사실\\
> > > 은혁은 영수 바로 옆에서 필요한 단계를 차근차근 설명할 준비를 하고 있었다. | 스토리에서 추출된 사실\\
> > > 영수는 은혁에게 진행될 절차에 대해 집중적으로 물어보고 있었다. | 스토리에서 추출된 사실\\
> > > 가까운 자리에서 함께 과정을 논의한다면, 진행 상황을 자연스럽게 지켜보게 된다. | 상식\\
> > 지민이는 USB가 이동하는 것을 보지 않았다. | 추론된 사실\\
> > > 지민은 실험이 끝난 뒤, 남은 도구 정리에 몰두하고 있었다. | 스토리에서 추출된 사실\\
> > > 지민은 영수가 맡은 프로젝트 진행 상황을 질문하며, 상대의 대답에만 주의를 기울였다. | 스토리에서 추출된 사실\\
> > > 누군가가 한 가지 일에 집중하면, 동시에 벌어지는 다른 일을 놓치기 쉽다. | 상식\\
\end{hintbox}

%% file: sections/tables/ko-musr_logic-tree_example/op-logic-tree-example-translated.tex
\begin{hintbox}{Object Placement Example Reasoning Tree (Translated)}{fig:example-op-tree-en}
Yeongsu stores important project data in a USB drive. If the USB is lost, it will be impossible to submit the project. | Deduced Conclusion\\
> Beginning of the story | Deduced Fact\\
> > Yeongsu saw the laptop on the laboratory desk. | Fact From Story\\
> > Yeongsu saw the USB on the laboratory desk. | Fact From Story\\
> > Eunhyeok saw the laptop on the laboratory desk. | Fact From Story\\
> > Eunhyeok saw the USB on the laboratory desk. | Fact From Story\\
> > Jimin saw the laptop on the laboratory desk. | Fact From Story\\
> > Jimin saw the USB on the laboratory desk. | Fact From Story\\
> Jimin moves the laptop to the storage shelf because she wants to clear space in the lab after the experiment. | Deduced Fact\\
> > Yeongsu saw the laptop being moved to the storage shelf. | Deduced Fact\\
> > > Right after the experiment ended, Jimin briefly mentioned her plan to organize the equipment to Yeongsu. | Fact From Story\\
> > > Yeongsu was standing right next to the equipment just before it was moved. | Fact From Story\\
> > > When something is moved at close range, nearby people can easily notice it. | Commonsense Knowledge\\
> > Eunhyeok saw the laptop being moved to the storage shelf. | Deduced Fact\\
> > > Eunhyeok was nearby, keeping an eye on the surroundings. | Fact From Story\\
> > > Jimin briefly mentioned her storage plan to Eunhyeok before organizing. | Fact From Story\\
> > > When an object is moved nearby, people close to it can easily perceive the situation. | Commonsense Knowledge\\
> Yeongsu moves the USB to the workbench drawer because he decides to keep it in a nearby, secure place for safekeeping. | Deduced Fact\\
> > Eunhyeok saw the USB being moved to the workbench drawer. | Deduced Fact\\
> > > Yeongsu was preparing to handle the device in front of the drawer. | Fact From Story\\
> > > Eunhyeok was beside Yeongsu, observing the situation and checking project progress. | Fact From Story\\
> > > Actions occurring in close proximity are easily recognized by nearby people. | Commonsense Knowledge\\
> > Jimin did not see the USB being moved to the workbench drawer. | Deduced Fact\\
> > > Eunhyeok was near Jimin, looking around the area. | Fact From Story\\
> > > Jimin briefly discussed the storage plan with Eunhyeok before cleaning up. | Fact From Story\\
> > > When someone is focused on a conversation, they may miss other events happening around them. | Commonsense Knowledge\\
> Eunhyeok moves the USB because he wants to strengthen data security by storing it safely. | Deduced Fact\\
> > Yeongsu saw the USB being moved. | Deduced Fact\\
> > > Eunhyeok was standing right next to Yeongsu, preparing to explain the necessary steps. | Fact From Story\\
> > > Yeongsu was asking detailed questions about the upcoming procedure. | Fact From Story\\
> > > When two people discuss a process closely together, they naturally observe the ongoing actions. | Commonsense Knowledge\\
> > Jimin did not see the USB being moved. | Deduced Fact\\
> > > After the experiment ended, Jimin was absorbed in tidying up the remaining tools. | Fact From Story\\
> > > Jimin was asking Yeongsu about the project’s progress, focusing solely on his responses. | Fact From Story\\
> > > When someone concentrates on one task, it is easy to miss other things happening simultaneously. | Commonsense Knowledge\\
\end{hintbox}

%% file: sections/tables/ko-musr_logic-tree_example/ta-example.tex
\begin{hintbox}{Team Allocation Example}{fig:example-ta-data-ko}
어느 따뜻한 봄날 아침, 새로운 소셜 미디어 플랫폼을 만들기 위해 모인 팀의 매니저로서 나는 사무실 창 밖을 바라보며 오늘의 과제를 떠올렸다. 
콘텐츠 제작자와 사용자 안전 관리자로 적합한 인물을 찾는 일이었다.
팀에는 지훈, 소영, 민지라는 세 명의 주목할 만한 인물이 있었고, 
그들의 강점과 약점을 잘 알고 있었기에 배치가 고민스러웠다.\\

먼저 지훈은 과거에 다양한 영상 콘텐츠를 제작한 경험이 있는 사람이었다. \\
그는 창의적인 아이디어로 팀에서 자주 인정받았지만, 
규칙이나 절차를 따르는 것을 힘들어했다. 
세부 사항에 대한 주의를 기울이지 않아 문제가 되기도 했고, 
민지와의 협업에서는 종종 의견 충돌을 겪곤 했다. 
게다가 그는 소영의 의견을 잘 듣지 않는 경향이 있었다.\\

소영은 관리 및 계획에 더 큰 관심을 보였으며 
창의적인 프로젝트보다는 큰 그림을 보는 것을 선호했다. 
그러나 그녀는 비디오 편집 소프트웨어 사용 경험이 거의 없었고, 
과거 프로젝트에서 반복적으로 작은 실수를 하기도 했다.
소영은 또한 상세한 규정 준수보다는 포괄적인 접근을 선호하여, 
때때로 민지가 결정을 내릴 때 주저한다고 느꼈다.
소영은 지훈과의 의사소통에서 좌절감을 느끼기도 했다.\\

마지막으로 민지는 분석적인 업무에 강점을 보여왔지만, 
복잡한 규정을 이해하고 적용하는 데 어려움을 겪었다. 
또한, 콘텐츠 제작 과정에서 창의적인 아이디어를 도출하는 데 어려움을 겪었고 
지나치게 혼란한 상황에서 침착함을 유지하기 어려웠다. 
민지는 지훈의 작업 스타일을 비효율적이라고 여겼으며, 
소영과 함께했던 이전 프로젝트에서는 중요한 결정이 항상 지연되었다고 보고했다.\\

이 세 사람을 놓고 생각해보면, 
각자의 고유한 강점과 약점이 조화를 이루도록 적절한 배치를 찾는 것이 중요했다. 
콘텐츠 제작자와 사용자 안전 관리자로 이들을 어떻게 나누어야 할지 고민이 깊어졌다. 
완벽한 대답은 존재하지 않을지도 모른다. 
어느 쪽이든, 그들의 개성과 역량을 최대로 끌어내기 위해서는 잘못된 조합을 피하고, 
이들의 다양한 특성을 잘 연결하는 것이 핵심이었다. 
위대한 플랫폼의 성공은 결국 올바른 사람을 올바른 자리에 배치하는 데 있을 것이기 때문이다.\\

\textbf{문제:} 이 이야기를 바탕으로, 두 가지 업무가 모두 효율적으로 수행되도록 사람을 어떻게 배정하시겠습니까?\\

\textbf{선택지:}\\
1. 콘텐츠 제작자: 지훈, 사용자 안전 관리자: 민지와 소영\\
2. 콘텐츠 제작자: 소영, 사용자 안전 관리자: 민지와 지훈\\
3. 콘텐츠 제작자: 민지, 사용자 안전 관리자: 소영과 지훈\\

\textbf{정답:} 1. 콘텐츠 제작자: 지훈, 사용자 안전 관리자: 민지와 소영

\end{hintbox}

%% file: sections/tables/ko-musr_logic-tree_example/ta-example-translated.tex
\begin{hintbox}{Team Allocation Example (Translated)}{fig:example-ta-data-en}
One warm spring morning, as the manager of a team gathered to build a new social media platform, I gazed out the office window and recalled today’s task: finding suitable members for the roles of content creator and user safety manager.  
The team had three notable members—Jihoon, Soyoung, and Minji—and knowing their strengths and weaknesses well made the decision all the more difficult.\\

First, Jihoon had previous experience creating various video content.\\
He was often recognized for his creative ideas, but struggled to follow rules and procedures.  
His lack of attention to detail had caused issues before, and he often clashed with Minji during collaboration.  
Furthermore, he tended to disregard Soyoung’s opinions.\\

Soyoung, on the other hand, was more interested in management and planning, preferring to see the big picture rather than focusing on creative projects.  
However, she had little experience using video editing software and had made several small mistakes in past projects.  
Soyoung also preferred broad approaches over strict compliance with detailed regulations, which sometimes made Minji hesitant when making decisions.  
She also felt frustrated when communicating with Jihoon.\\

Finally, Minji had shown strength in analytical tasks but struggled to understand and apply complex regulations.  
She also found it difficult to generate creative ideas during the content creation process and had trouble staying calm in overly chaotic situations.  
Minji considered Jihoon’s work style inefficient and reported that important decisions were often delayed in previous projects with Soyoung.\\

Considering these three individuals, it was important to find an arrangement where their unique strengths and weaknesses could complement each other.  
I pondered deeply over how to divide them between the roles of content creator and user safety manager.  
There might not be a perfect answer, but to maximize their potential, it was essential to avoid poor combinations and find a configuration that linked their diverse characteristics effectively.  
After all, the success of a great platform ultimately depends on placing the right people in the right positions.\\

\textbf{Question:} Based on this story, how would you assign the people so that both tasks are performed efficiently?\\

\textbf{Choices:}\\
1. Content Creator: Jihoon, User Safety Managers: Minji and Soyoung\\
2. Content Creator: Soyoung, User Safety Managers: Minji and Jihoon\\
3. Content Creator: Minji, User Safety Managers: Soyoung and Jihoon\\

\textbf{Answer:} 1. Content Creator: Jihoon, User Safety Managers: Minji and Soyoung
\end{hintbox}

%% file: sections/tables/ko-musr_logic-tree_example/ta-logic-tree-example.tex
\begin{hintbox}{Team Allocation Example Reasoning Tree}{fig:example-ta-tree-ko}
당신은 새로운 소셜 미디어 플랫폼을 만들고 있습니다. 콘텐츠 제작 팀과 사용자 안전 팀을 어떻게 나누어야 할까요? | 추론된 결론\\
> 지훈과 민지는 같이 못 일한다. | 추론된 사실\\
> > 민지는 지훈의 작업 스타일을 비효율적이라고 생각한다. | 스토리에서 추출된 사실\\
> > 지훈과 민진은 과거 프로젝트에서 협업할 때 항상 의견 충돌을 겪었다. | 스토리에서 추출된 사실\\
> > 사람들이 과거에 지속적으로 의견 충돌이 있었다면, 다시 함께 일할 때도 충돌이 발생할 가능성이 높다. | 상식\\
> 지훈은 콘텐츠 제작자를 할 수 있다. | 추론된 사실\\
> > 지훈은 창의적인 아이디어로 팀에서 자주 인정받는다. | 스토리에서 추출된 사실\\
> > 지훈은 과거에 다양한 영상 콘텐츠를 제작한 경험이 있다. | 스토리에서 추출된 사실\\
> > 콘텐츠 제작자는 창의성과 제작 경험이 중요한 요소다. | 상식\\
> 지훈은 사용자 안전 관리자를 못 한다. | 추론된 사실\\
> > 지훈은 과거 프로젝트에서 세부 사항에 대한 주의를 기울이지 않아 문제가 되었다. | 스토리에서 추출된 사실\\
> > 지훈은 규칙이나 절차를 따르는 것을 힘들어한다. | 스토리에서 추출된 사실\\
> > 사용자 안전 관리자는 규칙 준수와 세부 사항에 대한 철저한 관리가 필요하다. | 상식\\
> 민지는 콘텐츠 제작자를 못 한다. | 추론된 사실\\
> > 민지는 주로 분석적인 업무에 더 강점을 보인다. | 스토리에서 추출된 사실\\
> > 민지는 콘텐츠 제작 과정에서 창의적인 아이디어를 도출하는데 어려움을 겪는다. | 스토리에서 추출된 사실\\
> > 콘텐츠 제작자는 창의력을 요구하며, 분석적인 성향보다는 창의적인 사고방식을 필요로 한다. | 상식\\
> 민지는 사용자 안전 관리자를 못 한다. | 추론된 사실\\
> > 민지는 복잡한 규정을 이해하고 적용하는 데 어려움을 겪는다. | 스토리에서 추출된 사실\\
> > 민지는 지나치게 혼란한 상황에서 침착함을 유지하기 어려워한다. | 스토리에서 추출된 사실\\
> > 사용자 안전 관리자는 복잡한 상황에서도 냉정함을 유지하고 규정을 정확히 이해하는 것이 중요하다. | 상식\\
> 소영과 민지는 같이 못 일한다. | 추론된 사실\\
> > 소영과 민지는 이전 프로젝트에서 중요한 결정이 항상 지연되었다고 보고했다. | 스토리에서 추출된 사실\\
> > 소영은 민지가 결정을 내릴 때 주저한다고 느낀다. | 스토리에서 추출된 사실\\
> > 사람들이 함께 일할 때 결정 지연이 지속적으로 발생하면, 향후에도 같은 문제가 발생하는 경향이 있다. | 상식\\
> 지훈과 소영은 같이 못 일한다. | 추론된 사실\\
> > 소영은 지훈과의 의사소통에서 좌절감을 느꼈다. | 스토리에서 추출된 사실\\
> > 지훈은 소영의 의견을 잘 듣지 않는 경향이 있다. | 스토리에서 추출된 사실\\
> > 의사소통이 원활하지 않으면, 효율적인 협업이 어렵다. | 상식\\
> 소영은 콘텐츠 제작자를 못 한다. | 추론된 사실\\
> > 소영은 창의적인 프로젝트보다 관리 및 계획에 더 관심이 크다. | 스토리에서 추출된 사실\\
> > 소영은 비디오 편집 소프트웨어 사용 경험이 거의 없다. | 스토리에서 추출된 사실\\
> > 콘텐츠 제작자는 창의적인 기술과 전문 소프트웨어 사용 능력이 필요하다. | 상식\\
> 소영은 사용자 안전 관리자를 못 한다. | 추론된 사실\\
> > 소영은 세부적인 규정 준수보다는 큰 그림을 보는 것을 선호한다. | 스토리에서 추출된 사실\\
> > 소영은 과거 프로젝트에서 반복적으로 작은 실수를 했었다. | 스토리에서 추출된 사실\\
> > 사용자 안전 관리자는 작은 실수도 치명적으로 이어질 수 있어, 세부적인 규정 준수가 필수적이다. | 상식

\end{hintbox}

%% file: sections/tables/ko-musr_logic-tree_example/ta-logic-tree-example-translated.tex
\begin{hintbox}{Team Allocation Example Reasoning Tree (Translated)}{fig:example-ta-tree-en}
You are creating a new social media platform. How should you divide the content creation team and the user safety team? | Deduced Conclusion\\
> Jihoon and Minji cannot work together. | Deduced Fact\\
> > Minji thinks Jihoon’s work style is inefficient. | Fact From Story\\
> > Jihoon and Minji always had conflicts when collaborating on past projects. | Fact From Story\\
> > If people have had continuous conflicts in the past, it is highly likely that conflicts will arise again when they work together. | Commonsense Knowledge\\
> Jihoon can be a content creator. | Deduced Fact\\
> > Jihoon is often recognized by the team for his creative ideas. | Fact From Story\\
> > Jihoon has experience producing various video contents in the past. | Fact From Story\\
> > Creativity and production experience are important qualities for a content creator. | Commonsense Knowledge\\
> Jihoon cannot be a user safety manager. | Deduced Fact\\
> > Jihoon once caused problems in past projects by failing to pay attention to details. | Fact From Story\\
> > Jihoon finds it difficult to follow rules or procedures. | Fact From Story\\
> > A user safety manager must adhere to rules and pay close attention to detail. | Commonsense Knowledge\\
> Minji cannot be a content creator. | Deduced Fact\\
> > Minji is generally stronger in analytical work. | Fact From Story\\
> > Minji struggles to come up with creative ideas during the content creation process. | Fact From Story\\
> > A content creator requires creativity and needs a creative mindset rather than an analytical one. | Commonsense Knowledge\\
> Minji cannot be a user safety manager. | Deduced Fact\\
> > Minji finds it difficult to understand and apply complex regulations. | Fact From Story\\
> > Minji has trouble staying calm in overly chaotic situations. | Fact From Story\\
> > A user safety manager must stay composed in complex situations and understand regulations accurately. | Commonsense Knowledge\\
> Soyeong and Minji cannot work together. | Deduced Fact\\
> > Soyeong and Minji reported that important decisions were always delayed in their previous project. | Fact From Story\\
> > Soyeong feels that Minji hesitates when making decisions. | Fact From Story\\
> > When people consistently experience decision-making delays while working together, similar problems tend to occur in the future. | Commonsense Knowledge\\
> Jihoon and Soyeong cannot work together. | Deduced Fact\\
> > Soyeong felt frustrated when communicating with Jihoon. | Fact From Story\\
> > Jihoon tends not to listen well to Soyeong’s opinions. | Fact From Story\\
> > Ineffective communication makes efficient collaboration difficult. | Commonsense Knowledge\\
> Soyeong cannot be a content creator. | Deduced Fact\\
> > Soyeong is more interested in management and planning than in creative projects. | Fact From Story\\
> > Soyeong has little experience using video editing software. | Fact From Story\\
> > A content creator needs creative technical skills and proficiency with professional software. | Commonsense Knowledge\\
> Soyeong cannot be a user safety manager. | Deduced Fact\\
> > Soyeong prefers to focus on the big picture rather than on detailed rule compliance. | Fact From Story\\
> > Soyeong repeatedly made small mistakes in past projects. | Fact From Story\\
> > In user safety management, even small mistakes can be critical, so strict compliance with detailed regulations is essential. | Commonsense Knowledge
\end{hintbox}

%% file: sections/A99-Large-Tables.tex
\onecolumn
\section{Examples for Human Evaluations}
\label{sec:irb-examples}

\input{sections/tables/icl-examples/icl-komm-original}
\input{sections/tables/icl-examples/icl-komm-translated}
\input{sections/tables/icl-examples/icl-koop-original}
\input{sections/tables/icl-examples/icl-koop-translated}
\input{sections/tables/icl-examples/icl-kota-original}
\input{sections/tables/icl-examples/icl-kota-translated}

\section{Listings}
\label{sec:misc}
\input{sections/tables/annotation-instruction/humaneval-instruction}



\twocolumn

\input{sections/tables/annotation-instruction/instruction_1}
\input{sections/tables/annotation-instruction/irb-consents}

\input{sections/tables/table-zeroshot-direct-nohint-full}
\input{sections/tables/table-3shot-cot-hint-full}
\input{sections/tables/table-slms-full}
\input{sections/tables/table-ablation-num-fewshots-full}
\input{sections/tables/table-ablation-hints-full}
\input{sections/tables/table-ablation-cot-full}
\input{sections/tables/table-ablation-slm-full}

%% file: sections/tables/icl-examples/icl-komm-original.tex
\begin{hintbox}{1 Shot CoT Hint Example for Ko-MuSR - Murder Mysteries}{fig:irb-example-mm-ko}

\textbf{문제}\\

동물원의 한적한 오후, 영은은 전기충격기에 의해 살해되었고, 담당 형사는 재균과 서율 두 용의자를 심문하기 시작했다.\\

비가 부슬부슬 내리는 동물원 주차장에 다다르자, 김 형사는 금속문 소리를 들으며 날카로운 공기 속에서 서율의 존재감을 느꼈다. 그는 늘 그런 식으로 주변을 압도하는 기운을 풍겼다. 김 형사는 그의 사무실로 향하는 길에 중얼거리듯 자신과 보이지 않는 존재에 대해 이야기하던 그의 말이 머릿속을 스쳤다. 그가 불가해한 말들을 내뱉던 때가 한두 번이 아니었기 때문이다.\\

사무실 문을 열고 들어서자, 서율은 눈에 띄게 피곤해 보였다. 그는 김 형사를 보자 비즈니스 연필 꽂이에서 연필을 하나 뽑아 손가락 사이에서 놀리며 자리에서 일어섰다.\\

“김 형사님, 와주셔서 감사해요. 제가 도와드릴 수 있는 게 있을까요?” 서율이 말했다.\\

“서율 씨, 언제나 그랬듯이 동물원 운영에 대해 몇 가지 물어보고 싶습니다.” 김 형사는 점잖게 말했다.\\

김 형사는 간단히 목소리를 가다듬고 다시 질문을 던졌다.\\

“영은 씨와의 관계는 어떤가요? 최근에 자주 같이 계신다는 이야기를 들었습니다.”\\

서율은 잠시 시선을 피하더니 다시 김 형사를 바라보았다.\\

“글쎄요, 우리의 사업적 충돌은 있었습니다. 하지만 그 이상은 아닙니다.”\\

장미의 가시 사이로 피뢰침이 번쩍이는 순간처럼 서율의 눈길이 강하게 빛났다.\\

김 형사는 그의 말을 듣고 잠시 생각에 잠겼다. 모든 것이 명확하게 서지 않았지만, 이러한 모든 이야기가 얽히고설킨 추리의 일부처럼 다가오는 것을 느꼈다. 이 회색 구름 속에 감춰진 진실을 찾기 위해 김 형사는 서율을 계속 지켜봐야겠다는 생각이 들었다. 그런 생각을 하며, 김 형사는 다음으로 움직였다.\\

김 형사가 동물원 입구에 도착했을 때, 그는 이미 많은 관중과 경찰들로 붐비는 광경을 목격했다. 영은의 사건이 발생한 현장은 일반적인 사건 현장과는 사뭇 다른 분위기를 자아내고 있었다. 동물원의 복잡한 구조에도 불구하고, 그는 재빨리 사건 현장으로 이동했다.\\

“아, 형사님. 그냥 시간을 좀 확인하고 있었습니다. 언제 퇴근할 수 있는지요.”\\

재균은 자연스럽게 대답했다.\\

김 형사는 사무실 내부를 살펴보며 중요한 메모가 놓여 있는 책상을 발견했다. 재균이 개인적으로 적어두는 메모들이었다. 그는 메모를 눈으로 훑어보다가, 동물원 내부 구조와 보안 시스템에 대해 자세히 기록한 내용들을 보고 잠시 멈칫했다.\\

“이 메모는 자주 쓰시나요?” 김 형사가 물었다.\\

“네, 일을 매끄럽게 처리하기 위해서요. 동물원의 복잡한 구조 때문에 도움을 받곤 합니다.”\\

재균은 그럴듯하게 설명했다.\\

김 형사는 재균이 매일 퇴근 시간이 다가올 때마다 동료들의 책상을 지나가며 시간을 확인한다는 이야기를 들은 바 있었다. 사람들의 출퇴근 패턴을 유심히 살펴보는 그를 지켜보던 사람들도 많았다.\\

“여기서 일하면서 특별히 신경 쓰는 부분이 있으신가요?” 형사는 계속해서 질문을 던졌다.\\

“업무상 보안 시스템을 잘 알아야 하니까요. 주로 동료들이 주로 어두운 구석에서 대화를 나누는 걸 지켜보곤 합니다.”\\

재균은 담담하게 대답했다. 김 형사는 다음 질문을 물었다.\\

"최근에 영은 씨와의 관계는 어땠나요?"\\

재균은 잠시 고개를 떨군 채 침묵하다가 조용히 입을 열었다.\\

“솔직히... 예전 같지 않았습니다. 처음엔 서로 도우며 잘 지냈지만, 요즘 들어 영은 씨가 저를 의심하는 눈초리를 보내곤 했습니다.”\\

“의심이라면 어떤 부분을 말씀하시는 건가요?” 김 형사는 조용하지만 단호한 어조로 되물었다.\\

“전기 장비 점검 기록이나 출입구 로그에 대해 제가 잘못 관리하고 있다고 동물원 관리자에게 보고한 것도 그녀였죠. 덕분에 저는 조사까지 받았고, 그 일 이후로 동료들 사이에서도 신뢰를 잃었습니다. 저는 그저 제 방식대로 효율적으로 일하려 했을 뿐이었는데…”\\

그의 목소리에는 억눌린 분노가 스며 있었다. 김 형사는 조용히 고개를 끄덕이며 속으로 정리했다. 다음으로, 김 형사는 재균이 동물원의 구조를 너무나도 잘 알고 있다는 점을 주목했다. 재균의 그늘에서의 행동들은 그리 단순하지 않다는 것을 암시하는 듯했다. 형사는 재균 주변을 한 바퀴 돌아보며 그가 또 무슨 이야기를 덧붙일지 기다렸다.\\

“혹시 임무에 전기충격기가 관련되는 일은 있었습니까?” 형사가 묻자, 재균은 잠시 침묵했다.\\

“그런 일이 있었다면, 모두 업무 차원이었을 것입니다.”\\

재균은 말을 얼버무렸다.\\

형사는 더 깊은 이야기가 나올 것을 기대하며, 재균의 답변에 귀를 기울였다. 동물원의 사건과 관련된 비밀의 실마리가, 이제 조금씩 드러날 준비를 하고 있었다.\\

다음 중 살인자일 가능성이 가장 높은 사람은 누구인가요?\\

다음 중 하나를 선택하세요:\\
1 - 서율\\
2 - 재균\\

\textbf{힌트}\\

반드시 하나의 선택지만 고르셔야 합니다. 선택하기 전에, 단계별로 당신의 추론 과정을 설명하세요. 살인자는 반드시 세 가지 조건을 충족해야 합니다:
동기(피해자를 죽일 만한 이유가 있는가), 수단(범행 도구에 접근할 수 있는가), 기회(범행 현장에 접근할 수 있었는가)\\

무고한 용의자는 이 세 가지 중 두 가지까지만 입증될 수 있으며, 의심스러운 행동을 보일 수는 있어도 세 가지 조건을 모두 갖추지는 않습니다. 만약 두 용의자 모두 수단, 동기, 기회를 가진 것처럼 보인다면, 그 중에서도 가장 명확하게 이 세 가지가 드러난 인물을 선택해야 합니다. 반대로, 어느 누구도 세 가지를 모두 갖추지 않은 경우라면, 세 가지 조건 중 가장 뚜렷하게 드러난 인물을 선택해야 합니다.\\

단계별로 논리적인 추론을 마친 후, 정답을 고르세요\\

\textbf{예시풀이}\\

두 용의자를 각각 분석하여 어떤 용의자가 범행 동기, 범행 수단, 범행 기회를 모두 가지는지 추론해 보겠습니다.\\

먼저, 서율에 대해 살펴보겠습니다.\\

서율은 피해자인 영은과 사업적 충돌이 있었다고 말합니다. 이는 업무 관계 속에서 갈등이 있었음을 보여주며, 충분한 살인의 동기가 될 수 있습니다. 따라서 서율은 범행 동기가 있습니다.\\

서율은 전기충격기에 접근할 수 있었는지, 다룰 수 있었는지에 대한 직접적인 언급은 없습니다. 그의 업무상 전기충격기를 사용할 일에 대한 정보는 전혀 드러나 있지 않습니다. 따라서 서율은 범행 수단이 없습니다. \\

서율은 동물원 직원으로 사건 당일 동물원 내부에 있었고, 동물원 운영에 관련된 사람이므로 범행 장소에 접근할 수 있는 위치에 있었습니다. 따라서 서율은 범행 기회가 있습니다.\\

다음으로 재균에 대해 살펴보곘습니다.\\

재균은 피해자인 영은에 대해 직접적인 불만과 분노를 드러냅니다. 영은이 재균을 상대로 관리 문제를 관리자에게 보고했다고 하였고, 그로 인해 재균이 조사를 받았으며 동료들에게 신뢰를 잃게 된 계기가 되었습니다. 따라서 재균은 범행 동기가 있습니다.\\

형사가 재균에게 전기충격기 관련 임무가 있었느냐고 물었을 때, 그는 부정하지 않고 업무상 있었을 수도 있다고 얼버무리는 반응을 보였습니다. 또한, 재균은 전기 장비 점검 기록을 관리하고 동물원의 보안 시스템과 구조에 매우 정통한 인물입니다. 이는 그가 업무 차원에서 전기충격기에 접근할 수 있었고, 그 사용법에 대해서도 알고 있었을 가능성을 시사합니다. 따라서 재균은 범행 수단이 있습니다.\\

재균은 동물원의 출입 동선과 시간대에 매우 익숙하며, 사건 발생 시간 전후로 동물원 내에서 활동하고 있었던 정황이 있습니다. 동료들의 퇴근 시간까지 관찰하고 있었고, 동물원 보안 시스템에 익숙하며, 은밀한 장소에서 동료들의 대화를 지켜보는 습관도 언급됩니다. 이는 그가 사건 현장에 몰래 접근할 수 있는 능력과 기회를 충분히 가지고 있었음을 의미합니다. 따라서 재균은 범행 기회가 있습니다.\\

위의 분석을 정리해보면 다음과 같습니다:\\
서율은 범행 동기, 범행 기회가 있지만 범행 수단이 없습니다. \\
재균은 범행 동기, 범행 수단, 범행 기회가 모두 있습니다.\\

따라서, 가장 유력한 살인 용의자는 재균입니다.
\end{hintbox}

%% file: sections/tables/icl-examples/icl-komm-translated.tex
\begin{hintbox}{1 Shot CoT Hint Example for Ko-MuSR - Murder Mysteries (Translated)}{fig:irb-example-mm-en}

\textbf{Question}\\

On a quiet afternoon at the zoo, Yeongeun was killed by an electric shock device, and Detective Kim began interrogating the two suspects, Jaekyun and Seoyul.\\

As he arrived at the zoo’s quiet parking lot under a light drizzle, Detective Kim heard the metallic creak of the gate and sensed Seoyul’s presence in the sharp air. He had always had that way of dominating his surroundings with a certain aura. On his way to Seoyul’s office, Kim recalled the man’s strange words about himself and unseen beings—muttered as if in conversation with something invisible. It wasn’t the first time Seoyul had said such incomprehensible things.\\

When Kim opened the office door, Seoyul looked visibly exhausted. Upon seeing the detective, he stood up, picking a pencil from a business-style holder and twirling it between his fingers.\\

“Detective Kim, thank you for coming. How can I help you?” said Seoyul.\\

“Seoyul, as always, I’d like to ask you a few questions about the zoo’s operations,” Kim replied politely.\\

Kim cleared his throat slightly and continued his questioning.\\

“What was your relationship with Yeongeun? I’ve heard you’ve been spending time together recently.”\\

Seoyul averted his gaze briefly before looking back at the detective.\\

“Well, we did have some business conflicts. But nothing beyond that.”\\

His eyes flashed sharply, like lightning striking between the thorns of a rose.\\

Kim paused to think. Nothing was entirely clear yet, but all these fragments felt like pieces of a tangled mystery. To uncover the truth hidden in these gray clouds, Kim decided to keep a close eye on Seoyul. With that thought, he moved on.\\

When Kim arrived at the zoo entrance, he saw crowds of onlookers and police officers. The crime scene of Yeongeun’s death had an atmosphere unlike any other he had encountered. Despite the zoo’s complex layout, he quickly made his way to the scene.\\

“Ah, Detective. I was just checking the time, wondering when I might get off work,” said Jaekyun casually.\\

Kim scanned the office and noticed an important notebook on the desk—it contained Jaekyun’s personal memos. Skimming through, he paused when he saw detailed notes about the zoo’s internal layout and security systems.\\

“Do you often use these notes?” Kim asked.\\

“Yes, to make my work run smoothly. The zoo’s complex structure can be confusing, so they help me,” Jaekyun replied plausibly.\\

Kim recalled hearing that Jaekyun routinely checked the time near coworkers’ desks as the day ended. Many people had observed his habit of watching others’ comings and goings.\\

“Is there anything in particular you pay attention to while working here?” the detective continued.\\

“I need to know the security systems well for my job. I often notice colleagues chatting in dark corners,” Jaekyun answered calmly.\\

“Recently, how was your relationship with Yeongeun?” Kim asked next.\\

Jaekyun lowered his head for a moment in silence before speaking quietly.\\

“To be honest... it wasn’t like before. We used to help each other, but lately, she’s been looking at me with suspicion.”\\

“What kind of suspicion?” Kim asked in a calm but firm tone.\\

“She reported to the zoo manager that I mishandled maintenance logs and entry records. Because of that, I was investigated and lost trust among my coworkers. I only wanted to do my job efficiently in my own way…”\\

His voice carried restrained anger. Kim nodded quietly, organizing his thoughts. He noted that Jaekyun knew the zoo’s structure remarkably well. His shadowy behavior hinted at more than it seemed. Kim walked around him, waiting for what he might say next.\\

“Was your work ever related to the electric shock device?” Kim asked. Jaekyun hesitated.\\

“If such a thing happened, it would have been purely work-related,” he mumbled.\\

Kim listened carefully, expecting something deeper. The threads of the zoo’s mystery were beginning to unravel.\\

Who among the following is most likely to be the murderer?\\

Choose one of the following:\\
1 - Seoyul\\
2 - Jaekyun\\

\textbf{Hint}\\

You must choose only one option. Before choosing, explain your reasoning step by step. The murderer must satisfy all three conditions: motive (a reason to kill the victim), means (access to the murder weapon), and opportunity (access to the crime scene).\\

An innocent suspect can satisfy at most two of these conditions and may appear suspicious but cannot fulfill all three. If both suspects seem to have motive, means, and opportunity, choose the one for whom these are most clearly established. Conversely, if neither meets all three, select the one who most clearly satisfies the conditions that are evident.\\

After reasoning logically step by step, choose your answer.\\

\textbf{Example Solution}\\

We will analyze each suspect to determine who possesses all three elements—motive, means, and opportunity.\\

First, let’s examine Seoyul.\\

Seoyul admits to having business conflicts with the victim, Yeongeun. This shows there was tension in their professional relationship, which could serve as a motive for murder. Therefore, Seoyul has a motive.\\

There is no direct mention of Seoyul having access to or handling the electric shock device. His job responsibilities do not appear to involve such equipment. Therefore, Seoyul lacks the means.\\

As a zoo employee present on the day of the incident and involved in zoo operations, Seoyul had access to the crime scene. Therefore, he had the opportunity.\\

Next, let’s consider Jaekyun.\\

Jaekyun expresses direct resentment and anger toward the victim. Yeongeun reported him to the zoo manager for mishandling maintenance records, leading to an investigation and a loss of trust among colleagues. Thus, Jaekyun had a motive.\\

When asked whether his duties involved the electric shock device, Jaekyun neither denied nor confirmed it, implying possible involvement. He also manages electrical inspection logs and is highly knowledgeable about the zoo’s security systems and layout. This suggests he could access and operate the electric shock device through his work. Therefore, Jaekyun had the means.\\

Jaekyun is familiar with the zoo’s pathways and schedule patterns, and he was known to observe coworkers’ movements. His awareness of the zoo’s security and ability to move unnoticed indicate that he had access to the scene. Therefore, Jaekyun had the opportunity.\\

To summarize:\\
Seoyul has motive and opportunity but lacks means.\\
Jaekyun has motive, means, and opportunity.\\

Therefore, the most likely murderer is Jaekyun.
\end{hintbox}

%% file: sections/tables/icl-examples/icl-koop-original.tex
\begin{hintbox}{1 Shot CoT Hint Example for Ko-MuSR - Object Placements}{fig:irb-example-op-ko}

\textbf{문제}\\

민재는 떠나기 전 마지막으로 가방을 살펴보았다. 가방 안에 있는 카메라를 손으로 확인하자, 괜히 한숨이 놓이는 느낌이었다. 그 순간 지영은 책장에 놓인 수첩을 힐끗 바라보며 “이것도 꼭 챙겨야지. 놓고 가면 곤란하니까.” 하고 말했다. 옆에서 일정을 체크하던 준호는 메모를 하듯 수첩과 카메라의 위치를 다시 한번 상기했다. 사실 세 사람 모두 카메라는 가방에, 수첩은 책장에 있음을 잘 알고 있었다. 여행을 준비하는 마지막 점검이긴 했지만, 서로의 목표가 같았기에 불필요한 혼란은 없었다. 무엇보다 민재는 소중한 카메라를, 지영은 깔끔한 정리를, 그리고 준호는 철저한 일정 관리를 마음에 두고 무사히 여행길에 오르려 하고 있었다.\\

민재는 기기를 점검하기 위해 카메라를 탁자로 옮겼다. 민재는 탁자에 놓인 기기를 점검하고 있었다. 준호는 방 구석에서 자신이 챙겨야 할 일정을 고민하며 떠날 준비에만 집중하고 있었다. 지영이는 민재에게 점검 관련 의견을 전하며 가까이에서 필요한 도움을 주고 있었다. 준호는 여전히 민재에게 신경 쓰지 않았고, 지영이는 민재가 기기를 다루는 과정을 함께하고 있었다.\\

지영은 탁자에 있던 카메라를 들어 서랍 안에 넣었다. 민재는 친구가 무언가를 다루는 순간에 돕고자 주변 상황에 신경 쓰고 있었다. 민재와 지영은 같은 공간에서 나란히 정리 작업을 하고 있었다. 준호는 지영의 움직임에 집중하며 필요한 정보를 교환하고 있었다. 준호와 지영은 같은 공간에서 가까운 거리를 유지하고 있었다.\\

준호는 책장에 있던 수첩을 가방으로 옮겼다. 민재는 곧 출발한다는 생각에 마음이 급해, 주변 움직임을 살필 여유가 없었다. 떠나기 전 필요한 예약 내역을 다시 확인하느라 시선이 분산되었다. 지영이는 급히 챙겨야 할 준비 사항을 떠올리며 한곳에 집중해 있었고, 일정 확인에 몰두하느라 주변 시야를 살피지 않았다.\\

이 이야기는 여기서 끝이다. 

주어진 이야기를 고려할 때, 민재는 수첩을 찾기 위해 어디를 가장 먼저 확인하겠는가?\\

다음 중 하나만 고르시오:\\
1 – 가방\\
2 – 서랍\\
3 – 책장\\
4 - 탁자\\

\textbf{힌트} \\

이 이야기를 바탕으로, 우리는 특정 인물이 어떤 물건이 어디에 있다고 생각하고 있는지를 파악하려고 한다. 이를 위해서는 이야기 흐름을 따라가며 각 인물이 물건이 어디에 있다고 생각하는지 상태 추적이 필요하다. 물건이 옮겨질 때, 그 장면을 직접 보았거나, 옮겨진 위치를 인지할 수 있는 상태였다면 그들은 물건이 옮겨진 위치를 알고 있다고 본다. 하지만 그 순간을 보지 못했거나, 너무 다른 일에 집중하고 있었다면, 그들은 여전히 해당 물건이 이전 위치에 있다고 믿는다.\\

\textbf{예시풀이}\\

이 문제는 단계적으로 생각해보며 풀어야 합니다. 우리가 알고 싶은 것은 민재가 수첩을 찾기 위해 가장 먼저 어디를 확인할 것인가? 입니다. 이를 위해 민재가 이야기 속에서 수첩의 위치를 어디에서 봤고, 어디에 있다고 믿고 있는지 순서대로 확인해보겠습니다.\\

이야기 초반에는 다음과 같은 문장이 등장합니다: “사실 세 사람 모두 카메라는 가방에, 수첩은 책장에 있음을 잘 알고 있었다.” 이를 통해, 처음에 민재는 수첩이 책장에 있다고 인지하고 있음을 알 수 있습니다.\\

이야기 후반에, 준호가 책장에 있던 수첩을 가방으로 옮깁니다. 하지만, 이어지는 서술에 따르면 민재는 주변 움직임을 살필 여유가 없었다고 합니다. 따라서 민재는 준호가 수첩을 옮기는 장면을 보지 못했으며, 수첩이 옮겨졌다는 사실을 모른다는 점을 알 수 있습니다. 따라서, 민재는 여전히 수첩이 책장에 있다고 믿고 있다고 추측할 수 있습니다.\\

또한, 이야기 속 어느 부분에서도 민재가 서랍이나 가방 혹은 탁자 등 다른 장소로 수첩이 이동된 것을 직접 보거나 눈치채는 장면은 없습니다. 이 모든 정황을 종합해 보면, 민재는 수첩이 여전히 책장에 있다고 믿고 있을 것입니다.\\

따라서, 민재가 수첩을 찾기 위해 가장 먼저 확인할 장소는 책장입니다.

\end{hintbox}

%% file: sections/tables/icl-examples/icl-koop-translated.tex
\begin{hintbox}{1 Shot CoT Hint Example for Ko-MuSR - Object Placements (Translated)}{fig:irb-example-op-en}

\textbf{Question}\\

Before leaving, Minjae took one last look inside his bag. When he confirmed the presence of the camera inside, he felt a sense of relief. At that moment, Jiyeong glanced at the notebook on the bookshelf and said, “We should definitely take this too. It would be troublesome if we left it behind.” Junho, checking his schedule nearby, made a mental note of the notebook and camera’s locations once more. In fact, all three of them knew well that the camera was in the bag and the notebook was on the bookshelf. It was their final check before the trip, but since they shared the same goal, there was no unnecessary confusion. Above all, Minjae cared about his precious camera, Jiyeong cared about neat organization, and Junho, focused on precise scheduling, all prepared to depart smoothly.\\

Minjae moved the camera from the bag to the table to inspect the device. He examined it carefully, making sure everything was working. Junho, in the corner of the room, focused entirely on preparing for the trip and reviewing his itinerary. Jiyeong, standing nearby, assisted Minjae with suggestions and feedback. Junho continued to concentrate on his own preparations, paying no attention to Minjae, while Jiyeong remained beside him, involved in the inspection process.\\

Jiyeong then took the camera from the table and placed it inside the drawer. Minjae, noticing his friend’s actions, stayed attentive to help if needed. The two worked side by side, tidying up the space. Junho, meanwhile, observed Jiyeong’s movement closely, exchanging necessary information. Junho and Jiyeong were in close proximity, sharing the same space.\\

Afterward, Junho moved the notebook from the bookshelf to the bag. Minjae, feeling rushed to leave, didn’t have the composure to notice his surroundings. His attention was scattered as he reviewed his reservation details one last time. Jiyeong, preoccupied with final preparations, was also too focused to notice anything around her.\\

The story ends here.\\

Given this story, where would Minjae most likely check first when looking for the notebook?\\

Choose one of the following:\\
1 – Bag\\
2 – Drawer\\
3 – Bookshelf\\
4 – Table\\

\textbf{Hint}\\

Based on this story, we want to determine where a particular character believes an object to be located. To do this, we must track each character’s belief state by following the story and reasoning about what they have or have not observed. If a character sees an item being moved or can clearly infer its movement, they are aware of the new location. However, if they do not see the movement or are too focused on other activities, they continue to believe that the object remains in its previous location.\\

\textbf{Example Solution}\\

This problem should be solved step by step. The goal is to determine where Minjae would first look for the notebook. To answer this, we need to trace what Minjae saw during the story and where he believes the notebook to be.\\

At the beginning, we see the sentence: “All three of them knew well that the camera was in the bag and the notebook was on the bookshelf.” From this, we know that Minjae initially believes the notebook is on the bookshelf.\\

Later in the story, Junho moves the notebook from the bookshelf to the bag. However, it is stated that Minjae was too preoccupied to notice his surroundings at that time. Therefore, Minjae did not see Junho move the notebook and is unaware that it has been relocated.\\

Nowhere in the story does Minjae observe or realize that the notebook has moved elsewhere. Considering these details, Minjae would still believe that the notebook remains on the bookshelf.\\

Thus, the location that Minjae would most likely check first is the bookshelf.

\end{hintbox}

%% file: sections/tables/icl-examples/icl-kota-original.tex
\begin{hintbox}{1 Shot CoT Hint Example for Ko-MuSR - Team Assignments}{fig:irb-example-ta-ko}

\textbf{문제}\\

헬스케어 스타트업의 새로운 여정을 시작하며, 당신은 서진, 민주, 유환이라는 세 명의 팀원과 함께하게 되었습니다. 이들은 각기 다른 배경과 경험을 가지고 있으며, 회사의 성공을 위해 중요한 두 가지 과제인 '환자 관리 시스템 개발'과 '의료 마케팅 전략'에 기여할 예정입니다.\\

먼저, 서진은 디테일을 중시하며 소프트웨어 개발 분야에서 5년 넘게 경력을 쌓았습니다. 창의적인 문제 해결 능력을 인정받아 여러 차례 수상한 경험이 있는 서진은 최신 환자 관리 시스템 기술에 대한 세미나들도 꾸준히 참석하고 있었습니다. 그러나 과거에 민주와 협업하면서 중요도 우선순위 문제로 충돌을 겪었던 기억이 있습니다. \\

이에 비해, 민주는 반복적인 업무를 더 선호하며, 창의성 발휘에 어려움을 겪었습니다. 복잡한 시스템을 설계하거나 개발한 경험은 없지만 유환과는 협력의 경험이 많은데, 두 사람은 과거 여러 프로젝트에서 좋은 팀워크를 발휘했습니다. 특히, 팀 워크샵에서는 함께 의견을 제안하며 빼어난 호흡을 보여줬습니다.\\

유환은 소프트웨어 개발에 대한 풍부한 경험을 바탕으로 과거 헬스케어 시스템 개발 프로젝트에도 참여한 바 있습니다. 그는 다수의 성공적인 의료 캠페인을 주도했고, 이전 회사에서는 의료 마케팅 팀의 핵심 멤버로 활약하기도 했습니다. 그러나 서진과는 작업 스타일의 차이로 인해 예전 프로젝트에서 의견 충돌이 잦았습니다.\\

이제, 당신의 책상 위에는 두 개의 메모가 놓여 있습니다. 하나는 환자 관리 시스템 개발자로서의 역할을, 다른 하나는 의료 마케팅 전략가 역할을 위한 것입니다. 각자 적합한 인재에게 이 업무를 배정해야 하는데, 여기서 과연 어떤 배정이 '가장 좋은' 배정일까요?\\

당신은 이들의 과거 경험과 서로 간의 협력 관계를 고려하며 최적의 조합을 고민하게 됩니다. 답은 독자의 판단에 달려 있습니다. 이 세 명의 인재 중 누가 어느 역할을 맡는 것이 팀과 회사의 성공을 위한 최상의 선택일까요?\\

이 이야기를 바탕으로, 두 가지 업무가 모두 효율적으로 수행되도록 사람을 어떻게 배정하시겠습니까?\\
다음 중 하나를 선택하세요:\\
1 – 개발자 : 서진, 마케팅 전략가: 민주,유환\\
2 - 개발자: 민주, 마케팅 전략가: 서진,유환\\
3 - 개발자: 유환, 마케팅 전략가: 민주,서진\\

\textbf{힌트}\\

당신은 세 가지 중 하나의 선택지를 골라야 합니다. 스토리를 통해 각 사람이 한 가지 업무에 얼마나 능숙한지를 파악할 수 있습니다. 일반적으로 각 사람은 어떤 작업에 대해 뛰어나거나, 보통이거나, 부족한 실력을 가지고 있습니다. 우리는 가능한 한 각자의 강점을 최대한 발휘할 수 있도록 사람을 적절한 업무에 배치하려고 합니다.\\

또한, 두 사람이 함께 맡아야 하는 업무가 하나 있으며, 이 경우 두 사람의 팀워크 수준(훌륭함, 보통, 나쁨)도 전체 업무 성과에 중요한 영향을 미칩니다.\\

단, 두 사람이 함께 일해야 하는 작업에서 한 사람이 실력이 부족하고, 두 사람의 팀워크도 좋지 않다면, 다른 한 사람이 아무리 뛰어나도 전체 결과에는 도움이 되지 않을 수 있습니다.\\

각기 다른 강점, 약점, 그리고 두 사람 간의 상호작용을 고려하여, 전체 과제가 가장 효율적으로 수행될 수 있도록 팀원들을 적절히 배치해야 합니다. \\

\textbf{예시풀이}\\

이제 단계별로 생각해보며 문제를 해결해봅시다. 먼저 각 인물의 업무별 능력을 살펴보고, 이후에 서로 간의 협업 관계를 고려하여 가장 효율적인 선택지를 찾아보겠습니다.\\

먼저, 서진부터 살펴보겠습니다.\\

서진은 소프트웨어 개발 분야에서 5년 이상의 경력을 보유하고 있으며, 최신 환자 관리 시스템 관련 세미나에도 꾸준히 참석하고 있습니다. 따라서 서진의 환자 관리 시스템 개발 능력을 3점으로 평가합니다.의료 마케팅 분야에서는 직접적인 경험은 없지만, 창의적인 문제 해결 능력으로 수상 경력이 있다고 하였습니다. 따라서 서진의 의료 마케팅 능력을 2점으로 평가합니다.\\

다음은 민주입니다.\\

민주는 반복적인 업무를 선호하고, 복잡한 시스템을 설계하거나 개발한 경험이 없습니다. 따라서 민주의 환자 관리 시스템 개발 능력을 1점으로 평가합니다. 또한, 민주는 반복적인 업무를 선호하고 창의성 발휘에 어려움을 겪었다고 하였습니다. 따라서 민주의 의료 마케팅 능력을 1점으로 평가합니다.\\

마지막으로 유환에 대해 고려해보겠습니다.\\

유환은 헬스케어 시스템 개발 경험이 있고, 소프트웨어 개발 경험도 풍부합니다. 따라서 유환의 환자 관리 시스템 개발 능력을 2점으로 평가합니다. 또한 그는 의료 마케팅 캠페인을 성공적으로 주도한 경험이 있고, 해당 팀의 핵심 멤버였습니다. 따라서 유환의 의료 마케팅 능력을 3점으로 평가합니다.\\

다음으로 인물들 간 관계와 팀워크 수준을 살펴보겠습니다.\\

서진과 민주는 팀워크 수준이 나쁩니다. 서진은 과거 민주와 협업하면서 우선순위 문제로 갈등이 있었기 때문입니다. 따라서 서진과 민주의 팀워크 수준을 1점으로 평가합니다.\\

서진과 유환은 팀워크 수준이 나쁩니다. 유환과 서진은 작업 스타일의 차이로 인해 예전 프로젝트에서 의견 충돌이 잦았기 때문입니다. 따라서 서진과 유환의 팀워크 수준을 1점으로 평가합니다.\\

민주와 유환은 팀워크 수준이 훌륭합니다. 민주와 유환은 과거 여러 프로젝트에서 좋은 팀워크를 보여주었고 팀 워크샵에서는 빼어난 호흡을 보여주었습니다. 따라서 민주와 유환의 팀워크 수준을 3점으로 평가합니다.\\

이제 각 항목을 요약해보면 다음과 같습니다:\\

서진은 환자 관리 시스템 개발 능력이 뛰어나다. (3)\\
서진은 의료 마케팅 능력이 나쁘다. (2)\\
민주는 환자 관리 시스템 개발 능력이 나쁘다. (1)\\
민주는 의료 마케팅 능력이 나쁘다. (1)\\
유환은 환자 관리 시스템 개발 능력이 보통이다. (2)\\
유환은 의료 마케팅 능력이 뛰어나다. (3)\\
서진과 민주는 팀워크 수준이 나쁘다. (1)\\
서진과 유환은 팀워크 수준이 나쁘다. (1)\\
민주와 유환은 팀워크 수준이 훌륭하다. (3)\\

이제 옵션별 점수를 계산해 봅시다.\\

옵션 1:\\

환자 관리 시스템 개발자: 서진 (3)\\
의료 마케팅 전략가: 유환 (3) + 민주 (1)\\
민주와 유환의 협업 점수: 3\\
총합: 3 + 3 + 1 + 3 = 10점\\

옵션 2:\\

환자 관리 시스템 개발자: 민주 (1)\\
의료 마케팅 전략가: 유환 (3) + 서진 (2)\\
유환-서진 협업 점수: 1\\
총합: 1 + 3 + 2 + 1 = 7점\\

옵션 3:\\

환자 관리 시스템 개발자: 유환 (2)\\
의료 마케팅 전략가: 민주 (1) + 서진 (2)\\
민주-서진 협업 점수: 1\\
총합: 2 + 1 + 2 + 1 = 6점\\

따라서, 가장 점수가 높은 옵션 1이 최적의 배정입니다.\\

정답: 1

\end{hintbox}

%% file: sections/tables/icl-examples/icl-kota-translated.tex
\begin{hintbox}{1 Shot CoT Hint Example for Ko-MuSR - Team Assignments (Translated)}{fig:irb-example-ta-en}

\textbf{Question}\\

As you begin a new journey at a healthcare startup, you are joined by three team members: Seojin, Minju, and Yuhwan. Each of them brings a unique background and experience, and they will contribute to two crucial projects for the company's success: developing a patient management system and designing a medical marketing strategy. \\

First, Seojin is detail-oriented and has accumulated over five years of experience in software development. Recognized for her creative problem-solving skills and several awards, she has also consistently attended seminars on the latest patient management system technologies. However, she had previous conflicts with Minju over task prioritization during past collaborations. \\

In contrast, Minju prefers repetitive tasks and struggles with creativity. Although she lacks experience designing or developing complex systems, she has worked well with Yuhwan on several projects, demonstrating excellent teamwork. In particular, during team workshops, they showed great synergy while proposing ideas together.\\

Yuhwan has extensive experience in software development and has participated in healthcare system development projects. He has also led several successful medical campaigns and served as a key member of the medical marketing team at his previous company. However, he often clashed with Seojin in past projects due to differing work styles.\\

Now, on your desk are two notes: one for the patient management system developer role and one for the medical marketing strategist role. You must assign these tasks to the most suitable individuals. Which assignment would be the best choice for the team's and the company's success? \\

Considering their past experiences and teamwork dynamics, how would you allocate the roles to ensure both projects are performed efficiently?  \\

Choose one of the following options:\\
1 – Developer: Seojin, Marketing Strategists: Minju, Yuhwan  \\
2 – Developer: Minju, Marketing Strategists: Seojin, Yuhwan  \\
3 – Developer: Yuhwan, Marketing Strategists: Minju, Seojin  \\

\textbf{Hint}\\

You must choose one of the three options. The story provides information about each person’s strengths for each task. Generally, each person is strong, average, or weak in specific tasks. Our goal is to assign people to roles where they can make the best use of their strengths.\\

Additionally, one of the tasks requires two people to work together, so the quality of their teamwork (excellent, average, poor) also significantly affects overall performance.\\

If two people must collaborate and one lacks skill while their teamwork is poor, the overall performance will be low, regardless of the other person’s ability.\\

Considering individual strengths, weaknesses, and team dynamics, assign roles so that the tasks can be completed as efficiently as possible.\\

\textbf{Example Solution}\\

Let’s solve this step by step. First, we assess each person’s abilities for each task, then consider their teamwork relationships to find the most effective combination.\\

First, let’s look at Seojin.\\

Seojin has over five years of experience in software development and regularly attends seminars on patient management systems. Therefore, we rate Seojin’s ability in patient management system development as 3 points. Although she lacks direct experience in medical marketing, her award-winning creative problem-solving skills merit a score of 2 points in medical marketing.\\

Next is Minju.\\

Minju prefers repetitive tasks and has no experience in designing or developing complex systems. Thus, her patient management system development ability is rated 1 point. She also struggles to show creativity, so her medical marketing ability is also rated 1 point.\\

Finally, Yuhwan.\\

Yuhwan has experience in healthcare system development and extensive software development experience. Thus, we rate his patient management system development ability as 2 points. He also successfully led medical marketing campaigns and was a key team member, so we rate his medical marketing ability as 3 points.\\

Next, let’s review their teamwork levels.\\

Seojin and Minju have poor teamwork. They had conflicts over priorities in the past. Thus, their teamwork score is 1 point.\\

Seojin and Yuhwan also have poor teamwork. They often clashed due to work style differences. Thus, their teamwork score is 1 point.\\

Minju and Yuhwan have excellent teamwork. They worked well together on several projects and demonstrated great chemistry during workshops. Thus, their teamwork score is 3 points.\\

To summarize:\\

Seojin excels at patient management system development (3).\\
Seojin is weak at medical marketing (2).\\
Minju is weak at both patient management (1) and marketing (1).\\
Yuhwan is average at patient management (2) and excellent at marketing (3).\\
Seojin–Minju teamwork: poor (1).\\
Seojin–Yuhwan teamwork: poor (1).\\
Minju–Yuhwan teamwork: excellent (3).\\

Now, let’s calculate the total scores for each option.\\

Option 1:\\
Developer: Seojin (3)\\
Marketing Strategists: Yuhwan (3) + Minju (1)\\
Teamwork (Minju–Yuhwan): 3\\
Total: 3 + 3 + 1 + 3 = 10 points\\

Option 2:\\
Developer: Minju (1)\\
Marketing Strategists: Yuhwan (3) + Seojin (2)\\
Teamwork (Yuhwan–Seojin): 1\\
Total: 1 + 3 + 2 + 1 = 7 points\\

Option 3:\\
Developer: Yuhwan (2)\\
Marketing Strategists: Minju (1) + Seojin (2)\\
Teamwork (Minju–Seojin): 1\\
Total: 2 + 1 + 2 + 1 = 6 points\\

Therefore, the highest total score is achieved with Option 1, which is the optimal assignment.\\

Answer: 1

\end{hintbox}

%% file: sections/tables/annotation-instruction/humaneval-instruction.tex
\begin{hintbox}{Instructions for Human Evaluation Participants(Translated)}{fig:irb-instruction}
This document has been prepared to help participants become familiar with the types of questions in this study and to explain how to solve them.
You will solve a total of three domains in one day, each consisting of 40 questions.
This document provides one example question and its corresponding example solution for each domain.
Please read the example questions and example solutions carefully to understand how to solve the problems in each domain.
You may start solving the problems after you have thoroughly reviewed the examples, and you may refer back to them at any time during the problem-solving session.\\

The questions you will solve are divided into three domains as follows.\\

(1) \textbf{Murder Mysteries:}\\
A story of up to 1,500 characters and two suspects are given. The story includes a detective and the two suspects, and describes the detective’s investigation of them. You are asked to read the story and identify the murderer according to the example solution method.\\

(2) \textbf{Object Placements:}\\
A story of up to 1,000 characters and four candidate locations for an object are given. The story describes how the characters perform their roles and move the object, resulting in changes in its location over time. You are asked to read the story and, following the example solution method, determine the last location where each character saw the object.\\

(3) \textbf{Team Assignments:}\\
A story of up to 1,000 characters and three possible team assignment plans are given. The story describes three individuals, their work abilities, and the level of teamwork among them. You are asked to read the story and, following the example solution method, choose the best team assignment plan among the three options.\\

Example questions and example solutions for each domain are provided on the following pages.
\end{hintbox}

%% file: sections/tables/annotation-instruction/instruction_1.tex
\onecolumn
\begin{table*}[htbp]
\centering
\begin{tabular}{p{0.15\textwidth}|p{0.75\textwidth}}
\toprule
\textbf{First Phase} & 
\tablecell{
\textbf{Objective}: Solve all given questions and submit your answer for each questions. Mark all questions with three types of marks defined below. If you think the data instances require edits or possess issue, please leave it to the \textit{Note} section.}
\begin{itemize}
    \item \textbf{Pass}: The answer can be derived from the narrative and has no issue to be used as a benchmark instance.
    \item \textbf{Minor Issue}: The instance contains small issues (e.g., typos, unclear wording) but it seems to be usable with minor edits.
    \item \textbf{Fail}: The instance has critical problems (e.g., logical errors, ambiguous answers) and cannot be used without significant edits.  
\end{itemize}
\\ \midrule

\textbf{Second Phase} & \tablecell{
\textbf{Objective}: You are given a question where the first reviewer flagged minor issues or incorrectly marked as pass. Refer to the supposed answer and check if this answer can be derived from the narrative. Mark all questions with three types of marks defined below. If you think the data instances require edits or possess issue, please leave it to the \textit{Note} section.}
\begin{itemize}
    \item \textbf{Pass}: The answer can be derived from the narrative and has no issue to be used as a benchmark instance.
    \item \textbf{Minor Issue}: The instance contains small issues (e.g., typos, unclear wording) but it seems to be usable with minor edits.
    \item \textbf{Fail}: The instance has critical problems (e.g., logical errors, ambiguous answers) and cannot be used without significant edits.  
\end{itemize}
\\ \bottomrule

\end{tabular}

\caption{Annotation Instructions for Data Reviewers}
\label{tab:annotation-instruction}
\end{table*}

\twocolumn

%% file: sections/tables/annotation-instruction/irb-consents.tex
\begin{table*}[!htbp]\centering
\resizebox{\textwidth}{!}{ 
\begin{tabular}{l}\toprule
\textbf{Q. What procedures will be followed if the participants take part in the study?}\\
\midrule
\tablecell{
You will solve a total of 120 multiple-choice questions across three domains under the supervision of the assigned researcher.
Once you confirm your willingness to participate, the following process will take place.\\

First, the researcher will provide you with instructions on how to solve the problems, example questions, and sample solutions for those examples.
You will receive one example question per domain.\\

Next, you will solve a total of 120 multiple-choice questions across the three domains and record your answers on the answer sheet for submission.\\

(1) \textbf{Murder Mysteries}\\
You will be given a story (within 1,500 characters) and two suspects. The story involves a detective and two suspects, and describes the detective’s investigation into them. Following the example solution method, you must determine who the murderer is.\\

(2) \textbf{Object Placements}\\
You will be given a story (within 1,000 characters) and four possible choices for the object’s location. The story describes how the characters interact with and move the object over time. Following the example solution method, you must identify where each character last saw the object.\\

(3) \textbf{Team Allocations}\\
You will be given a story (within 1,000 characters) and three team assignment options. The story presents three characters, their work abilities, and their teamwork levels. Following the example solution method, you must select the best team assignment plan among the three options.\\

The researcher will provide you with a study laptop and research materials. To eliminate the influence of AI tools (e.g., ChatGPT) on your problem-solving ability, you are \textbf{not allowed to use AI tools} during the process. The provided laptop will have internet access disabled for security and data protection. 
Although internet use is unnecessary for solving the tasks, you may use your personal device for web searches if desired. At the end of the study, you must return all research materials and the provided laptop.\\
}
\\ \toprule
\textbf{Q. How long will the study participation last?}\\
\midrule
\tablecell{
Your participation will take place on the date agreed upon with the researcher, from 9:00 AM to 6:00 PM.
Although the problem-solving session itself takes about 6 hours, additional time will be required for explanations of the problem-solving methods and related instructions.\\

You will be provided with a one-hour break from 12:00 PM to 1:00 PM.
If you wish to adjust the break time, you may freely do so after consulting with the researcher.
You may also take additional short breaks at your discretion.
Depending on your solving speed, the total duration may increase or decrease.
If you finish solving the problems early, you may end your participation between 4:00 PM and 6:00 PM after consulting with the researcher.
Even if your participation ends early, the compensation for your participation will remain unchanged.
}
\\ \toprule
\textbf{Q. Are there any side effects or risks?} \\ \midrule
\tablecell{
Some problems, such as those in the murder mystery domain, may cause discomfort or unpleasant emotions. In such cases, you may stop solving the problems at any time.
Since you will be solving problems provided as files, there are virtually no other physical or psychological risks associated with participation.
Your problem-solving performance will be evaluated on a 100-point scale, and only the score (not any identifying information) will be made public.

However, you may experience some discomfort due to the content of the questions, the testing environment, or the researcher’s behavior. If this occurs, please notify the researcher immediately.
}
\\ \toprule
\textbf{Q. Will compensation be provided for participating in this study?}\\
\midrule
\tablecell{
As a token of appreciation for your participation, you will receive a payment of 100\$.
However, if you withdraw from the problem-solving session before completion or request that your results be discarded, the compensation unfortunately cannot be provided.
The payment will be given only to participants who complete the entire session and provide results for all 120 questions.
If you request the deletion of your problem-solving results after the payment has been made, you will be required to return the compensation.
} \\
\bottomrule
\end{tabular}}
\caption{Prescribed instructions from IRB-approved participant information sheet.}\label{tab:irb-consent}
\end{table*}

%% file: sections/tables/table-zeroshot-direct-nohint-full.tex
\begin{table*}[!htp]\centering
\resizebox{\textwidth}{!}{ 
\begin{tabular}{lc|ccc|ccc}\toprule
Model &Seed &En-MM &En-OP &En-TA &Ko-MM &Ko-OP &Ko-TA \\\midrule
\multirow{2}{*}{QWEN3 32B} &5678 &66.0 &54.7 &72.4 &70.0 &79.2 &80.8 \\
&9875 &66.8 &50.8 &74.4 &65.2 &80.8 &78.4 \\
\multirow{2}{*}{LLaMA 4 Scout} &5678 &65.2 &55.1 &58.8 &56.0 &67.6 &64.0 \\
&9875 &65.6 &53.1 &62.0 &56.0 &68.0 &64.4 \\
\multirow{2}{*}{Exaone 4.0 32B} &5678 &63.2 &56.2 &60.4 &52.8 &40.4 &64.4 \\
&9875 &59.2 &55.1 &57.2 &52.8 &34.4 &62.0 \\
\multirow{2}{*}{HyperClova X SEED Think 14B} &5678 &51.6 &40.2 &44.4 &46.0 &40.4 &57.6 \\
&9875 &51.6 &37.1 &43.6 &43.2 &42.0 &59.6 \\
\bottomrule
\end{tabular}}
\caption{All Results from Iterations for Table ~\ref{tab:zeroshot-direct-nohint} }\label{tab:zeroshot-direct-nohint-full}
\end{table*}

%% file: sections/tables/table-3shot-cot-hint-full.tex
\begin{table*}[!htp]\centering
\resizebox{\textwidth}{!}{ 
\begin{tabular}{lc|ccc|ccc}\toprule
Model &Seed &En-MM &En-OP &En-TA &Ko-MM &Ko-OP &Ko-TA \\\midrule
\multirow{2}{*}{QWEN3 32B} &5678 &66.0 &54.7 &72.4 &70.0 &79.2 &80.8 \\
&9875 &66.8 &50.8 &74.4 &65.2 &80.8 &78.4 \\
\multirow{2}{*}{ + 1 Shot / Direct / Hint} &5678 &84.4 &75.8 &88.8 &75.6 &86.8 &88.0 \\
&9875 &84.4 &73.8 &89.6 &81.6 &87.2 &87.6 \\
\multirow{2}{*}{ + 1 Shot / CoT / Hint} &5678 &84.0 &73.4 &88.8 &78.8 &79.6 &87.2 \\
&9875 &83.2 &78.1 &90.4 &76.8 &78.8 &88.4 \\
\multirow{2}{*}{ + 3 Shots / Direct / Hint} &5678 &83.2 &77.7 &90.0 &79.2 &88.0 &88.8 \\
&9875 &81.6 &75.8 &90.4 &76.4 &85.6 &88.8 \\
\multirow{2}{*}{ + 3 Shots / CoT / Hint} &5678 &81.6 &75.4 &90.0 &74.8 &81.6 &88.8 \\
&9875 &83.2 &73.0 &87.6 &77.2 &82.0 &89.2 \\ \midrule
\multirow{2}{*}{LLaMA 4 Scout} &5678 &65.2 &55.1 &58.8 &56.0 &67.6 &64.0 \\
&9875 &65.6 &53.1 &62.0 &56.0 &68.0 &64.4 \\
\multirow{2}{*}{ + 1 Shot / Direct / Hint} &5678 &74.0 &61.7 &84.4 &74.4 &79.6 &79.6 \\
&9875 &74.0 &60.5 &87.6 &79.2 &80.0 &82.8 \\
\multirow{2}{*}{ + 1 Shot / CoT / Hint} &5678 &78.8 &69.5 &83.6 &75.6 &81.2 &86.0 \\
&9875 &75.6 &69.9 &85.2 &74.4 &80.4 &85.2 \\
\multirow{2}{*}{ + 3 Shots / Direct / Hint} &5678 &75.2 &68.0 &85.2 &73.2 &80.8 &87.6 \\
&9875 &76.8 &68.4 &88.0 &74.0 &81.6 &86.8 \\
\multirow{2}{*}{ + 3 Shots / CoT / Hint} &5678 &74.8 &68.8 &83.6 &73.6 &82.4 &86.8 \\
&9875 &76.4 &68.4 &84.8 &72.0 &83.6 &89.2 \\ \midrule
\multirow{2}{*}{Exaone 4.0 32B} &5678 &63.2 &56.2 &60.4 &52.8 &40.4 &64.4 \\
&9875 &59.2 &55.1 &57.2 &52.8 &34.4 &62.0 \\
\multirow{2}{*}{ + 1 Shot / Direct / Hint} &5678 &78.0 &78.1 &86.8 &70.4 &61.6 &82.8 \\
&9875 &75.2 &68.8 &84.8 &68.8 &61.6 &77.2 \\
\multirow{2}{*}{ + 1 Shot / CoT / Hint} &5678 &81.2 &73.0 &87.2 &68.0 &72.8 &81.6 \\
&9875 &80.4 &75.0 &86.0 &67.2 &69.2 &77.6 \\
\multirow{2}{*}{ + 3 Shots / Direct / Hint} &5678 &74.0 &73.0 &83.6 &70.4 &67.6 &84.4 \\
&9875 &73.2 &72.3 &83.6 &69.6 &62.0 &84.0 \\
\multirow{2}{*}{ + 3 Shots / CoT / Hint} &5678 &73.2 &73.8 &87.6 &70.8 &71.6 &85.2 \\
&9875 &77.2 &70.3 &84.0 &71.6 &71.6 &86.4 \\ \midrule
\multirow{2}{*}{HyperClova X SEED Think 14B} &5678 &51.6 &40.2 &44.4 &46.0 &40.4 &57.6 \\
&9875 &51.6 &37.1 &43.6 &43.2 &42.0 &59.6 \\
\multirow{2}{*}{ + 1 Shot / Direct / Hint} &5678 &71.2 &43.8 &66.8 &58.4 &44.0 &60.8 \\
&9875 &66.8 &42.2 &67.2 &60.4 &40.0 &55.6 \\
\multirow{2}{*}{ + 1 Shot / CoT / Hint} &5678 &69.2 &51.6 &75.2 &64.8 &38.8 &82.8 \\
&9875 &71.2 &51.2 &75.2 &62.8 &36.4 &86.4 \\
\multirow{2}{*}{ + 3 Shots / Direct / Hint} &5678 &66.8 &66.4 &82.0 &64.0 &51.6 &83.2 \\
&9875 &67.2 &63.3 &85.6 &63.6 &47.6 &84.4 \\
\multirow{2}{*}{ + 3 Shots / CoT / Hint} &5678 &69.2 &68.0 &80.4 &70.4 &48.8 &88.0 \\
&9875 &68.0 &64.5 &81.6 &70.8 &51.2 &87.6 \\
\bottomrule

\end{tabular}}
\caption{All Results from Iterations for Table ~\ref{tab:ablation-cot} }\label{tab:ablation-cot-full}
\end{table*}

%% file: sections/tables/table-slms-full.tex
\begin{table*}[!htp]\centering
\resizebox{\textwidth}{!}{ 
\begin{tabular}{lc|ccc|ccc}\toprule
Model &Seed &En-MM &En-OP &En-TA &Ko-MM &Ko-OP &Ko-TA \\\midrule
\multirow{2}{*}{QWEN3 0.6B} &5678 &59.2 &51.6 &38.8 &56.4 &28.4 &36.8 \\
&9875 &59.2 &52.3 &38.8 &50.8 &28.8 &39.2 \\
\multirow{2}{*}{ + 1 Shot / CoT / Hint} &5678 &58.4 &47.3 &38.8 &50.8 &32.0 &33.6 \\
&9875 &56.0 &46.1 &34.4 &53.2 &22.8 &31.2 \\
\multirow{2}{*}{ + 3 Shots / CoT / Hint} &5678 &57.2 &48.0 &32.8 &50.0 &30.4 &32.4 \\
&9875 &53.6 &45.3 &33.2 &48.0 &30.0 &34.4 \\ \midrule
\multirow{2}{*}{HyperClova X SEED Think 14B} &5678 &55.2 &44.1 &43.6 &48.0 &29.6 &46.0 \\
&9875 &56.0 &45.3 &56.0 &48.0 &28.4 &43.6 \\
\multirow{2}{*}{ + 1 Shot / CoT / Hint} &5678 &65.6 &43.4 &57.2 &58.0 &29.6 &41.6 \\
&9875 &68.0 &44.9 &58.4 &58.0 &30.0 &41.2 \\
\multirow{2}{*}{ + 3 Shots / CoT / Hint} &5678 &63.6 &47.7 &52.8 &58.0 &30.0 &34.4 \\
&9875 &66.0 &41.8 &56.8 &59.2 &33.2 &37.2 \\ \midrule
\multirow{2}{*}{Exaone 4.0 32B} &5678 &63.6 &52.3 &58.0 &48.0 &33.6 &66.0 \\
&9875 &57.6 &52.3 &60.0 &52.0 &34.4 &60.4 \\
\multirow{2}{*}{ + 1 Shot / CoT / Hint} &5678 &71.2 &43.0 &58.8 &56.0 &30.0 &52.4 \\
&9875 &70.4 &41.4 &57.6 &56.0 &31.6 &56.8 \\
\multirow{2}{*}{ + 3 Shots / CoT / Hint} &5678 &66.0 &37.9 &58.4 &62.4 &32.0 &51.6 \\
&9875 &66.0 &41.4 &52.4 &63.6 &34.0 &56.8 \\ \midrule
\multirow{2}{*}{Exaone 3.5 2.4B Instruct} &5678 &52.0 &43.4 &44.8 &51.6 &30.4 &42.0 \\
&9875 &50.8 &39.8 &43.2 &50.4 &29.6 &42.8 \\
\multirow{2}{*}{ + 1 Shot / CoT / Hint} &5678 &60.4 &51.2 &36.4 &57.6 &39.2 &41.2 \\
&9875 &62.4 &44.1 &42.8 &55.2 &41.6 &34.4 \\
\multirow{2}{*}{ + 3 Shots / CoT / Hint} &5678 &62.4 &46.9 &36.4 &55.2 &39.6 &40.4 \\
&9875 &64.4 &50.4 &38.0 &56.8 &42.8 &45.2 \\ \midrule
\multirow{2}{*}{QWEN3 4B} &5678 &64.0 &54.7 &66.0 &58.4 &52.4 &72.8 \\
&9875 &61.6 &56.2 &66.0 &59.6 &49.6 &78.0 \\
\multirow{2}{*}{ + 1 Shot / CoT / Hint} &5678 &76.8 &71.1 &85.6 &68.0 &55.2 &76.4 \\
&9875 &75.6 &67.6 &86.8 &71.6 &56.8 &76.4 \\
\multirow{2}{*}{ + 3 Shots / CoT / Hint} &5678 &72.0 &69.9 &84.4 &70.0 &54.4 &78.8 \\
&9875 &76.0 &71.5 &86.4 &69.6 &56.0 &80.4 \\
\bottomrule

\end{tabular}}
\caption{Full Table for Table ~\ref{tab:slm-main} }\label{tab:slm-main-full}
\end{table*}

%% file: sections/tables/table-ablation-num-fewshots-full.tex
\begin{table*}[!htp]\centering
\resizebox{\textwidth}{!}{ 
\begin{tabular}{lc|ccc|ccc}\toprule
Model &Seed &En-MM &En-OP &En-TA &Ko-MM &Ko-OP &Ko-TA \\\midrule
\multirow{2}{*}{QWEN3 32B} &5678 &66.0 &54.7 &72.4 &70.0 &79.2 &80.8 \\
&9875 &66.8 &50.8 &74.4 &65.2 &80.8 &78.4 \\
\multirow{2}{*}{ + 1 Shot / Direct / No Hint} &5678 &72.8 &63.3 &80.4 &69.2 &82.8 &79.6 \\
&9875 &74.4 &60.9 &81.6 &69.2 &83.2 &81.6 \\
\multirow{2}{*}{ + 3 Shots / Direct / No Hint} &5678 &76.8 &67.2 &75.6 &71.2 &83.6 &82.8 \\
&9875 &75.6 &62.1 &76.4 &72.4 &87.2 &80.8 \\
\multirow{2}{*}{ + 1 Shot / CoT / Hint} &5678 &84.0 &73.4 &88.8 &78.8 &79.6 &87.2 \\
&9875 &83.2 &78.1 &90.4 &76.8 &78.8 &88.4 \\
\multirow{2}{*}{ + 3 Shots / CoT / Hint} &5678 &81.6 &75.4 &90.0 &74.8 &81.6 &88.8 \\
&9875 &83.2 &73.0 &87.6 &77.2 &82.0 &89.2 \\ \midrule
\multirow{2}{*}{LLaMA 4 Scout} &5678 &65.2 &55.1 &58.8 &56.0 &67.6 &64.0 \\
&9875 &65.6 &53.1 &62.0 &56.0 &68.0 &64.4 \\
\multirow{2}{*}{ + 1 Shot / Direct / No Hint} &5678 &68.8 &54.7 &70.4 &74.0 &71.2 &70.4 \\
&9875 &70.4 &58.2 &69.6 &71.6 &74.8 &71.2 \\
\multirow{2}{*}{ + 3 Shots / Direct / No Hint} &5678 &70.8 &59.8 &75.6 &70.4 &76.0 &82.4 \\
&9875 &74.4 &63.3 &80.4 &70.4 &75.2 &82.8 \\
\multirow{2}{*}{ + 1 Shot / CoT / Hint} &5678 &78.8 &69.5 &83.6 &75.6 &81.2 &86.0 \\
&9875 &75.6 &69.9 &85.2 &74.4 &80.4 &85.2 \\
\multirow{2}{*}{ + 3 Shots / CoT / Hint} &5678 &74.8 &68.8 &83.6 &73.6 &82.4 &86.8 \\
&9875 &76.4 &68.4 &84.8 &72.0 &83.6 &89.2 \\ \midrule
\multirow{2}{*}{Exaone 4.0 32B} &5678 &63.2 &56.2 &60.4 &52.8 &40.4 &64.4 \\
&9875 &59.2 &55.1 &57.2 &52.8 &34.4 &62.0 \\
\multirow{2}{*}{ + 1 Shot / Direct / No Hint} &5678 &75.2 &57.4 &61.6 &62.8 &44.4 &65.2 \\
&9875 &74.4 &58.2 &65.2 &62.0 &53.2 &65.6 \\
\multirow{2}{*}{ + 3 Shots / Direct / No Hint} &5678 &75.6 &63.7 &68.4 &68.4 &63.2 &72.4 \\
&9875 &76.4 &61.3 &70.4 &68.8 &57.2 &74.8 \\
\multirow{2}{*}{ + 1 Shot / CoT / Hint} &5678 &81.2 &73.0 &87.2 &68.0 &72.8 &81.6 \\
&9875 &80.4 &75.0 &86.0 &67.2 &69.2 &77.6 \\
\multirow{2}{*}{ + 3 Shots / CoT / Hint} &5678 &73.2 &73.8 &87.6 &70.8 &71.6 &85.2 \\
&9875 &77.2 &70.3 &84.0 &71.6 &71.6 &86.4 \\ \midrule
\multirow{2}{*}{HyperClova X SEED Think 14B} &5678 &51.6 &40.2 &44.4 &46.0 &40.4 &57.6 \\
&9875 &51.6 &37.1 &43.6 &43.2 &42.0 &59.6 \\
\multirow{2}{*}{ + 1 Shot / Direct / No Hint} &5678 &65.6 &42.6 &53.2 &48.8 &37.6 &49.6 \\
&9875 &62.0 &41.4 &54.0 &51.2 &40.4 &54.0 \\
\multirow{2}{*}{ + 3 Shots / Direct / No Hint} &5678 &68.4 &64.5 &78.0 &66.4 &51.2 &86.4 \\
&9875 &67.2 &62.9 &74.0 &64.4 &53.6 &86.0 \\
\multirow{2}{*}{ + 1 Shot / CoT / Hint} &5678 &69.2 &51.6 &75.2 &64.8 &38.8 &82.8 \\
&9875 &71.2 &51.2 &75.2 &62.8 &36.4 &86.4 \\
\multirow{2}{*}{ + 3 Shots / CoT / Hint} &5678 &69.2 &68.0 &80.4 &70.4 &48.8 &88.0 \\
&9875 &68.0 &64.5 &81.6 &70.8 &51.2 &87.6 \\
\bottomrule

\end{tabular}}
\caption{All Results from Iterations for Table ~\ref{tab:ablation-fewshots} }\label{tab:ablation-fewshots-full}
\end{table*}

%% file: sections/tables/table-ablation-hints-full.tex
\begin{table*}[!htp]\centering
\resizebox{\textwidth}{!}{ 
\begin{tabular}{lc|ccc|ccc}\toprule
Model &Seed &En-MM &En-OP &En-TA &Ko-MM &Ko-OP &Ko-TA \\\midrule
\multirow{2}{*}{QWEN3 32B} &5678 &66.0 &54.7 &72.4 &70.0 &79.2 &80.8 \\
&9875 &66.8 &50.8 &74.4 &65.2 &80.8 &78.4 \\
\multirow{2}{*}{ + 1 Shot / Direct / No Hint} &5678 &72.8 &63.3 &80.4 &69.2 &82.8 &79.6 \\
&9875 &74.4 &60.9 &81.6 &69.2 &83.2 &81.6 \\
\multirow{2}{*}{ + 1 Shot / Direct / Hint} &5678 &84.4 &75.8 &88.8 &75.6 &86.8 &88.0 \\
&9875 &84.4 &73.8 &89.6 &81.6 &87.2 &87.6 \\
\multirow{2}{*}{ + 3 Shots / CoT / No Hint} &5678 &75.6 &67.2 &80.0 &71.6 &74.0 &82.8 \\
&9875 &73.2 &71.1 &80.4 &75.2 &76.8 &83.2 \\
\multirow{2}{*}{ + 3 Shots / CoT / Hint} &5678 &81.6 &75.4 &90.0 &74.8 &81.6 &88.8 \\
&9875 &83.2 &73.0 &87.6 &77.2 &82.0 &89.2 \\ \midrule
\multirow{2}{*}{LLaMA 4 Scout} &5678 &65.2 &55.1 &58.8 &56.0 &67.6 &64.0 \\
&9875 &65.6 &53.1 &62.0 &56.0 &68.0 &64.4 \\
\multirow{2}{*}{ + 1 Shot / Direct / No Hint} &5678 &68.8 &54.7 &70.4 &74.0 &71.2 &70.4 \\
&9875 &70.4 &58.2 &69.6 &71.6 &74.8 &71.2 \\
\multirow{2}{*}{ + 1 Shot / Direct / Hint} &5678 &74.0 &61.7 &84.4 &74.4 &79.6 &79.6 \\
&9875 &74.0 &60.5 &87.6 &79.2 &80.0 &82.8 \\
\multirow{2}{*}{ + 3 Shots / CoT / No Hint} &5678 &71.2 &62.5 &77.2 &70.8 &73.6 &84.4 \\
&9875 &72.0 &63.3 &79.2 &72.0 &78.0 &83.2 \\
\multirow{2}{*}{ + 3 Shots / CoT / Hint} &5678 &74.8 &68.8 &83.6 &73.6 &82.4 &86.8 \\
&9875 &76.4 &68.4 &84.8 &72.0 &83.6 &89.2 \\ \midrule
\multirow{2}{*}{Exaone 4.0 32B} &5678 &63.2 &56.2 &60.4 &52.8 &40.4 &64.4 \\
&9875 &59.2 &55.1 &57.2 &52.8 &34.4 &62.0 \\
\multirow{2}{*}{ + 1 Shot / Direct / No Hint} &5678 &75.2 &57.4 &61.6 &62.8 &44.4 &65.2 \\
&9875 &74.4 &58.2 &65.2 &62.0 &53.2 &65.6 \\
\multirow{2}{*}{ + 1 Shot / Direct / Hint} &5678 &78.0 &78.1 &86.8 &70.4 &61.6 &82.8 \\
&9875 &75.2 &68.8 &84.8 &68.8 &61.6 &77.2 \\
\multirow{2}{*}{ + 3 Shots / CoT / No Hint} &5678 &72.0 &60.2 &70.8 &69.2 &64.0 &82.4 \\
&9875 &77.2 &60.2 &69.2 &71.6 &61.6 &72.0 \\
\multirow{2}{*}{ + 3 Shots / CoT / Hint} &5678 &73.2 &73.8 &87.6 &70.8 &71.6 &85.2 \\
&9875 &77.2 &70.3 &84.0 &71.6 &71.6 &86.4 \\ \midrule
\multirow{2}{*}{HyperClova X SEED Think 14B} &5678 &51.6 &40.2 &44.4 &46.0 &40.4 &57.6 \\
&9875 &51.6 &37.1 &43.6 &43.2 &42.0 &59.6 \\
\multirow{2}{*}{ + 1 Shot / Direct / No Hint} &5678 &65.6 &42.6 &53.2 &48.8 &37.6 &49.6 \\
&9875 &62.0 &41.4 &54.0 &51.2 &40.4 &54.0 \\
\multirow{2}{*}{ + 1 Shot / Direct / Hint} &5678 &71.2 &43.8 &66.8 &58.4 &44.0 &60.8 \\
&9875 &66.8 &42.2 &67.2 &60.4 &40.0 &55.6 \\
\multirow{2}{*}{ + 3 Shots / CoT / No Hint} &5678 &67.2 &65.6 &75.2 &63.2 &47.2 &86.4 \\
&9875 &66.8 &61.7 &72.8 &66.0 &46.8 &82.8 \\
\multirow{2}{*}{ + 3 Shots / CoT / Hint} &5678 &69.2 &68.0 &80.4 &70.4 &48.8 &88.0 \\
&9875 &68.0 &64.5 &81.6 &70.8 &51.2 &87.6 \\
\bottomrule

\end{tabular}}
\caption{All Results from Iterations for Table ~\ref{tab:ablation-hint} }\label{tab:ablation-hint-full}
\end{table*}

%% file: sections/tables/table-ablation-cot-full.tex
\begin{table*}[!htp]\centering
\resizebox{\textwidth}{!}{ 
\begin{tabular}{lc|ccc|ccc}\toprule
Model &Seed &En-MM &En-OP &En-TA &Ko-MM &Ko-OP &Ko-TA \\\midrule
\multirow{2}{*}{QWEN3 32B} &5678 &81.6 &75.4 &90.0 &74.8 &81.6 &88.8 \\
&9875 &83.2 &73.0 &87.6 &77.2 &82.0 &89.2 \\
\multirow{2}{*}{LLaMA 4 Scout} &5678 &74.8 &68.8 &83.6 &73.6 &82.4 &86.8 \\
&9875 &76.4 &68.4 &84.8 &72.0 &83.6 &89.2 \\
\multirow{2}{*}{Exaone 4.0 32B} &5678 &73.2 &73.8 &87.6 &70.8 &71.6 &85.2 \\
&9875 &77.2 &70.3 &84.0 &71.6 &71.6 &86.4 \\
\multirow{2}{*}{HyperClova X SEED Think 14B} &5678 &69.2 &68.0 &80.4 &70.4 &48.8 &88.0 \\
&9875 &68.0 &64.5 &81.6 &70.8 &51.2 &87.6 \\

\bottomrule
\end{tabular}}
\caption{All Results from Iterations for Table ~\ref{tab:3shot-cot-hint} }\label{tab:3shot-cot-hint-full}
\end{table*}

%% file: sections/tables/table-ablation-slm-full.tex
\begin{table*}[!htp]\centering
\resizebox{\height}{!}{ 
\begin{tabular}{lc|ccc|ccc}\toprule
Model &Seed &En-MM &En-OP &En-TA &Ko-MM &Ko-OP &Ko-TA \\\midrule
\multirow{2}{*}{QWEN3 0.6B} &5678 &59.2 &51.6 &38.8 &56.4 &28.4 &36.8 \\
&9875 &59.2 &52.3 &38.8 &50.8 &28.8 &39.2 \\
\multirow{2}{*}{ + 1 Shot / Direct / No Hint} &5678 &56.0 &47.3 &32.0 &57.6 &22.0 &32.4 \\
&9875 &54.8 &54.3 &30.8 &49.6 &25.6 &35.6 \\
\multirow{2}{*}{ + 3 Shots / Direct / No Hint} &5678 &52.8 &48.0 &36.4 &58.0 &32.0 &30.4 \\
&9875 &56.0 &49.6 &36.4 &56.0 &25.2 &34.4 \\
\multirow{2}{*}{ + 1 Shot / CoT / Hint} &5678 &58.4 &47.3 &38.8 &50.8 &32.0 &33.6 \\
&9875 &56.0 &46.1 &34.4 &53.2 &22.8 &31.2 \\
\multirow{2}{*}{ + 3 Shots / CoT / Hint} &5678 &57.2 &48.0 &32.8 &50.0 &30.4 &32.4 \\
&9875 &53.6 &45.3 &33.2 &48.0 &30.0 &34.4 \\\midrule
\multirow{2}{*}{Exaone 4.0 1.2B} &5678 &55.2 &44.1 &43.6 &48.0 &29.6 &46.0 \\
&9875 &56.0 &45.3 &56.0 &48.0 &28.4 &43.6 \\
\multirow{2}{*}{ + 1 Shot / Direct / No Hint} &5678 &60.4 &45.7 &47.6 &55.2 &28.0 &42.4 \\
&9875 &60.4 &54.3 &48.0 &55.6 &26.4 &42.0 \\
\multirow{2}{*}{ + 3 Shots / Direct / No Hint} &5678 &62.4 &41.4 &46.4 &46.8 &29.6 &40.0 \\
&9875 &59.6 &44.1 &45.2 &52.0 &24.8 &40.4 \\
\multirow{2}{*}{ + 1 Shot / CoT / Hint} &5678 &65.6 &43.4 &57.2 &58.0 &29.6 &41.6 \\
&9875 &68.0 &44.9 &58.4 &58.0 &30.0 &41.2 \\
\multirow{2}{*}{ + 3 Shots / CoT / Hint} &5678 &63.6 &47.7 &52.8 &58.0 &30.0 &34.4 \\
&9875 &66.0 &41.8 &56.8 &59.2 &33.2 &37.2 \\\midrule
\multirow{2}{*}{QWEN3 1.7B} &5678 &63.6 &52.3 &58.0 &48.0 &33.6 &66.0 \\
&9875 &57.6 &52.3 &60.0 &52.0 &34.4 &60.4 \\
\multirow{2}{*}{ + 1 Shot / Direct / No Hint} &5678 &64.8 &49.6 &53.2 &58.0 &35.2 &56.0 \\
&9875 &63.2 &50.8 &55.6 &52.4 &31.6 &56.0 \\
\multirow{2}{*}{ + 3 Shots / Direct / No Hint} &5678 &60.4 &53.9 &54.0 &59.6 &32.8 &47.2 \\
&9875 &58.8 &52.3 &52.8 &56.4 &36.8 &54.8 \\
\multirow{2}{*}{ + 1 Shot / CoT / Hint} &5678 &71.2 &43.0 &58.8 &56.0 &30.0 &52.4 \\
&9875 &70.4 &41.4 &57.6 &56.0 &31.6 &56.8 \\
\multirow{2}{*}{ + 3 Shots / CoT / Hint} &5678 &66.0 &37.9 &58.4 &62.4 &32.0 &51.6 \\
&9875 &66.0 &41.4 &52.4 &63.6 &34.0 &56.8 \\\midrule
\multirow{2}{*}{Exaone 3.5 2.4B Instruct} &5678 &52.0 &43.4 &44.8 &51.6 &30.4 &42.0 \\
&9875 &50.8 &39.8 &43.2 &50.4 &29.6 &42.8 \\
\multirow{2}{*}{ + 1 Shot / Direct / No Hint} &5678 &58.0 &44.1 &44.0 &56.0 &40.0 &38.4 \\
&9875 &54.4 &46.5 &41.2 &55.6 &40.0 &36.8 \\
\multirow{2}{*}{ + 3 Shots / Direct / No Hint} &5678 &59.6 &44.1 &37.2 &61.6 &41.2 &39.6 \\
&9875 &54.0 &47.3 &39.6 &54.4 &39.2 &39.2 \\
\multirow{2}{*}{ + 1 Shot / CoT / Hint} &5678 &60.4 &51.2 &36.4 &57.6 &39.2 &41.2 \\
&9875 &62.4 &44.1 &42.8 &55.2 &41.6 &34.4 \\
\multirow{2}{*}{ + 3 Shots / CoT / Hint} &5678 &62.4 &46.9 &36.4 &55.2 &39.6 &40.4 \\
&9875 &64.4 &50.4 &38.0 &56.8 &42.8 &45.2 \\\midrule
\multirow{2}{*}{QWEN3 4B} &5678 &64.0 &54.7 &66.0 &58.4 &52.4 &72.8 \\
&9875 &61.6 &56.2 &66.0 &59.6 &49.6 &78.0 \\
\multirow{2}{*}{ + 1 Shot / Direct / No Hint} &5678 &64.4 &52.3 &70.0 &65.2 &50.8 &74.0 \\
&9875 &69.6 &54.3 &68.8 &61.6 &52.8 &74.0 \\
\multirow{2}{*}{ + 3 Shots / Direct / No Hint} &5678 &68.8 &46.9 &68.4 &66.8 &53.2 &71.2 \\
&9875 &67.6 &51.2 &70.8 &66.0 &54.0 &68.4 \\
\multirow{2}{*}{ + 1 Shot / CoT / Hint} &5678 &76.8 &71.1 &85.6 &68.0 &55.2 &76.4 \\
&9875 &75.6 &67.6 &86.8 &71.6 &56.8 &76.4 \\
\multirow{2}{*}{ + 3 Shots / CoT / Hint} &5678 &72.0 &69.9 &84.4 &70.0 &54.4 &78.8 \\
&9875 &76.0 &71.5 &86.4 &69.6 &56.0 &80.4 \\
\bottomrule

\end{tabular}}
\caption{All Results from Iterations for Table ~\ref{tab:slm-main-v3} and Table ~\ref{tab:ablation-slm} }\label{tab:ablation-slm-full}
\end{table*}